\documentclass[review]{elsarticle}

\usepackage{lineno,hyperref}
\modulolinenumbers[5]

\usepackage{graphicx}
\usepackage{tabularx}
\usepackage{booktabs}
\usepackage{multirow}
\usepackage[table,xcdraw]{xcolor}
\usepackage{varwidth}
\usepackage{array}
\usepackage{amssymb}
\usepackage{caption}
\usepackage{makecell}
\usepackage{amsmath}
\usepackage[ruled,vlined]{algorithm2e}

\usepackage[toc,page]{appendix}

\DeclareMathOperator*{\argmax}{arg\,max}

\newcolumntype{C}[1]{>{\centering\let\newline\\\arraybackslash\hspace{0pt}}m{#1}}

\usepackage{pbox}

\journal{Artificial Intelligence Journal (Published)}

\biboptions{authoryear}

\begin{document}

\newcommand{\jose}[1]{\textcolor{green}{#1}}
\newcommand{\red}[1]{\textcolor{red}{#1}}
\newcommand{\daniel}[1]{\textcolor{olive}{#1}}
\newcommand{\blue}[1]{\textcolor{blue}{#1}}

\begin{frontmatter}

\title{LMMS Reloaded: Transformer-based Sense Embeddings for  Disambiguation and Beyond\corref{cor2}}


\author[1]{Daniel Loureiro\corref{cor1}}
\ead{daniel.b.loureiro@inesctec.pt}
\author[1]{Al\'ipio M\'ario Jorge}
\ead{amjorge@fc.up.pt}
\author[2]{Jose Camacho-Collados}
\ead{camachocolladosj@cardiff.ac.uk}

\cortext[cor1]{Corresponding author}
\cortext[cor2]{AIJ Publication: \url{https://doi.org/10.1016/j.artint.2022.103661}}
\address[1]{LIAAD - INESC TEC, Dept. of Computer Science, FCUP, University of Porto, Portugal}
\address[2]{School of Computer Science and Informatics, Cardiff University, United Kingdom}

\begin{abstract}
Distributional semantics based on neural approaches is a cornerstone of Natural Language Processing, with surprising connections to human meaning representation as well.
Recent Transformer-based Language Models have proven capable of producing contextual word representations that reliably convey sense-specific information, simply as a product of self-supervision.
Prior work has shown that these contextual representations can be used to accurately represent large sense inventories as sense embeddings, to the extent that a distance-based solution to Word Sense Disambiguation (WSD) tasks outperforms models trained specifically for the task.
Still, there remains much to understand on how to use these Neural Language Models (NLMs) to produce sense embeddings that can better harness each NLM's meaning representation abilities.
In this work we introduce a more principled approach to leverage information from all layers of NLMs, informed by a probing analysis on 14 NLM variants.
We also emphasize the versatility of these sense embeddings in contrast to task-specific models, applying them on several sense-related tasks, besides WSD, while demonstrating improved performance using our proposed approach over prior work focused on sense embeddings.
Finally, we discuss unexpected findings regarding layer and model performance variations, and potential applications for downstream tasks.
\end{abstract}


\end{frontmatter}


\section{Introduction} \label{sec:introduction}

Lexical ambiguity is prevalent across different languages and plays an important role in improving communication efficiency \citep{PIANTADOSI2012280}.
Word Sense Disambiguation (WSD) is a long-standing challenge in the field of Natural Language Processing (NLP), and Artificial Intelligence more generally, with an extended history of research in computational linguistics \citep{Navigli2009WordSD}.

Interestingly, both computational and psychological accounts of meaning representation have converged on high-dimensional vectors within semantic spaces.

From the computational perspective, there is a rich line of work on learning word embeddings based on statistical regularities from unlabeled corpora, following the well-established Distributional Hypothesis \citep[DH]{harris1954distributional,Firth1957}.
The first type of distributional word representations relied on count-based methods, initially popularized by LSA \citep{deerwester1990indexing}, and later refined with GloVe \citep{pennington-etal-2014-glove}. 
Before GloVe, word embeddings learned with neural networks, first introduced by \cite{10.5555/944919.944966}, gained wide adoption with word2vec \citep{10.5555/2999792.2999959} and, afterwards, culminated with fastText \citep{bojanowski-etal-2017-enriching}.
The development and improvement of word embeddings has been a major contributor to the progress of NLP in the last decade \citep{goldberg2017neural}.

From the psychological perspective, there is also ample behavioural evidence in support of distributional representations of word meaning. Similarly to word embeddings, these representations are related according to the degree of shared features within semantic spaces, which translates into proximity in vector-space \citep{doi:10.1177/1745691619885860,KLEIN2001259}.
Understandably, the nature of the features making up this psychological account of semantic space, among other aspects (e.g., learning method), is not as clear as we find in the computational account.
Nevertheless, contextual co-occurrence is among the most informative factors for meaning representation as well \citep{Mcdonald01testingthe,erk-2016-alligator,Radach_Deubel_Vorstius_Hofmann-(eds.)_2017}.
There are even use cases in neurobiology motivating research into accurate distributional representations of word meaning.
In \cite{pereira2018toward}, word embeddings have proven useful for decoding words and sentences from brain activity, after learning a mapping between corpus-based embeddings (i.e., GloVe and word2vec) and fMRI activation.

The current understanding of how humans perform disambiguation attributes major relevance to sentential context, and other linguistic and paralinguistic cue's (e.g., speaker accent) to a lesser extent \citep{doi:10.1177/1745691619885860,CAI201773}.
However, the previously mentioned computational approaches are not designed for sense-level representation due to the Meaning Conflation Deficiency \citep{CamachoCollados2018FromWT}, as they converge different senses into the same word-level representation. 
Some works have explored variations on the word2vec method for sense-level embeddings \citep{rothe-schutze-2015-autoextend,iacobacci-etal-2015-sensembed,pilehvar-collier-2016-de,mancini-etal-2017-embedding}, but the dynamic word-level interactions composing sentential context were not targeted by those works.

The works of \cite{melamud-etal-2016-context2vec,yuan-etal-2016-semi,peters-etal-2018-deep} were among the first to propose Neural Language Models (NLMs) featuring dynamic word embeddings conditioned on sentential context (i.e., contextual embeddings).
These works showed that NLMs (trained exclusively on language modelling objectives) can produce contextual embeddings for word forms that are sensitive to the word's usage in particular sentences. 
Furthermore, these works also addressed WSD tasks with a simple nearest neighbours solution ($k$-NN) based on proximity between contextual embeddings.
Their results rivalled systems trained specifically for WSD (i.e., with additional modelling objectives), highlighting the accuracy of these contextual embeddings.

However, it was not until the development of Transformer-based NLMs, namely BERT \citep{devlin-etal-2019-bert}, that contextual embeddings from NLMs showed clearly better performance on WSD tasks than previous systems trained specifically for WSD (LMMS, \citealp{loureiro-jorge-2019-language}).

In this earlier work, we explored how to further take advantage of the representational power of NLMs through propagation strategies and encoding sense definitions.
Besides pushing the state-of-the-art of WSD, in \citet{loureiro-jorge-2019-language} we created sense embeddings for every entry in the Princeton WordNet v3.0 (200k word senses, \citealp{Fellbaum2000WordNetA}), so that the semantic space being represented is granular and expansive enough to encompass general knowledge domains for various parts-of-speech of the English language.
With this fully populated semantic space at our disposal we suggested strategies for uncovering biases and world knowledge represented by NLMs.

Since our work on LMMS, others have shown additional performance gains for WSD with fine-tuning or classification approaches that make better usage of sense definitions \citep{huang-etal-2019-glossbert,blevins-zettlemoyer-2020-moving}, semantic relations from external resources \citep{scarlinietal:2020,bevilacqua-navigli-2020-breaking}, or altogether different approaches to WSD \citep{barba-etal-2021-consec}.

However, there are several questions still standing regarding how to leverage NLMs for creating accurate and versatile sense embeddings, beyond optimizing for WSD benchmarks only.
Given that semantic spaces with distributional representations of word meanings feature prominently in both the conventional computational and psychological accounts of word disambiguation, these questions warrant further exploration.

\paragraph{\textbf{Contributions}}
In this extension of LMMS, we broaden our scope to more recent Transformer-based models in addition to BERT \citep{yang2019xlnet,DBLP:journals/corr/abs-1907-11692,Lan2020ALBERT:} (14 model variants in total), verify whether they exhibit similar proficiency at sense representation, and explore how performance variation can be attributed to particular differences in these models.
Striving for a principled approach to sense representation with NLMs, we also introduce a new layer pooling method, inspired by recent findings of layer specialization \citep{coenen-etal-2019-visualizing}, which we show is crucial to effectively use these new NLMs for sense representation.
Most importantly, in this article we provide a general framework for learning sense embeddings with Transformers and perform an extensive evaluation of such sense embeddings from different NLMs on various sense-related tasks, emphasizing the versatility of these representations.

\clearpage

\paragraph{\textbf{Outline}}
This work is organized as follows. We first provide some background information on the main topics of this research: Vector Semantics ($\S$\ref{pre:vec}), Neural Language Modelling ($\S$\ref{pre:nlm}) and Sense Inventories ($\S$\ref{pre:wn}).
Next, we describe related work on Sense Embeddings ($\S$\ref{rel:embs}), WSD ($\S$\ref{rel:wsd}) and Probing NLMs ($\S$\ref{rel:lay}).

The method used to produce this work's sense embeddings is described in Section \ref{met:intro}, covering aspects of the method introduced in \citet{loureiro-jorge-2019-language} (from $\S$\ref{met:embed} to $\S$\ref{met:apply}), as well as our new layer pooling method in Section \ref{met:lay}.

In Section \ref{exp:intro} we describe our experimental setting, providing relevant details about our choice of NLMs ($\S$\ref{exp:models}) and annotated corpora used to learn sense embeddings ($\S$\ref{exp:datasets}).

The layer pooling methodology described in Section \ref{met:lay} requires validating performance under two distinct modes of application.
Consequently, in Section \ref{probing:results} we report on performance variation per layer across NLMs ($\S$\ref{probing:variation}), highlight differences between disambiguation and matching profiles ($\S$\ref{probing:profiles}), and present the rationale for choosing particular profiles for each task ($\S$\ref{probing:choosing}).

In Section \ref{eval:intro}, we tackle several sense-related tasks using our proposed sense embeddings and compare results against the state-of-the-art, namely: WSD ($\S$\ref{eval:wsd}), Uninformed Sense Matching ($\S$\ref{eval:usm}), Word-in-Context ($\S$\ref{eval:wic}), Graded Word Similarity in Context ($\S$\ref{eval:gcs}) and Paired Sense Similarity ($\S$\ref{eval:sid}).

In order to better understand the contributions of this work, Section \ref{analysis:intro} reports on several ablation analyses targeting the following: choice of Sense Profiles ($\S$\ref{analysis:profiles}), impact of unambiguous word annotations ($\S$\ref{analysis:annotations}), merging gloss representations ($\S$\ref{analysis:glosses}), and indirect representation of synsets ($\S$\ref{analysis:indirect}).

We discuss our findings in Section \ref{disc:intro}, regarding representations from intermediate layers of NLMs ($\S$\ref{disc:layers}), irregularities across models and variants ($\S$\ref{disc:meaning}), and potential downstream applications of our sense embeddings focusing on knowledge integration ($\S$\ref{disc:apps}).

Finally, in Section \ref{conclusion} we present our concluding remarks, and provide details about our release of sense embeddings, code and more.

\section{Preliminaries}

This work exploits the interaction between vector-based semantic representations ($\S$\ref{pre:vec}), recent developments on NLMs ($\S$\ref{pre:nlm}), and curated sense inventories ($\S$\ref{pre:wn}).
In this section we provide some background on these topics.

\subsection{Vector Semantics}
\label{pre:vec}

Nearly a century ago, \citet{doi:10.1111/j.1467-968X.1935.tb01254.x} postulated that ``the meaning of a word is always contextual, and no study of meaning apart from context can be taken seriously".
Indeed, after working on formal theories of word meaning definition, \citet{wittgenstein1953philosophical} conceded ``the meaning of a word is its use in a language".
This view of meaning representation became known as the Distributional Hypothesis \citep[DH]{harris1954distributional}, which proposes that words that occur in the same contexts tend to have similar meanings. 
During this period, \citet{osgood1957measurement} further proposed representing the meaning of words as points in multi-dimensional space, with similar words having similar representations, thus being placed closely in this space.
Still, it would take a few more decades of computing advancements to appreciate the implications of the DH.

\paragraph{Early VSMs}
After some early works introducing vector space models (VSMs) for information retrieval (Salton \citeyear{salton1971smart}; \citeyear{salton1975vector}), Deerwester (\citeyear{deerwester1989computer}; \citeyear{deerwester1990indexing}) was the first to use dense vectors to represent word meaning, initially with a method called Latent Semantic Indexing (LSI), and later with Latent Semantic Analysis (LSA).
LSA was based on a word-document weighted frequency matrix from which the first 300-dimensions resulting from Singular-Value Decomposition (SVD) would correspond to word embeddings.
\citet{lund1996producing} introduced another influential method similar to LSA, called Hyperspace Analogue to Language (HAL) which differed from LSA by considering word-word frequencies instead, introducing the notion of a fixed-sized window as context (e.g., the two words to the left and to the right) instead full documents, which would become the standard representation of context.
Following these developments, \citet{landauer1997solution} evaluated the performance of LSA embeddings learned from large corpora on a simple semantic task (synonymy tests) and found that these embeddings performed comparably to school-aged children, when measuring similarity between word pairs as the cosine similarity between their corresponding embeddings (inspired by applications for information retrieval).
Already in this early period, \citet{schutze1992dimensions} and \citet{yarowsky-1995-unsupervised} realized the potential for WSD applications based on the similarity between unsupervised word embeddings.
\citet{10.5555/944919.944937} would later introduce Latent Dirichlet Allocation (LDA) which uses a generative probabilistic approach to generalize and improve on the approach used for LSA, being widely adopted for topic modelling and other applications beyond semantic analysis.

\paragraph{Neural Models}
Having established that corpus-based word embeddings are able to capture semantic knowledge, additional progress followed swiftly.
A milestone in the evolution of word embeddings was the discovery that Neural Language Models (NLMs) implicitly develop word embeddings when training for the task of word prediction \citep{bengio2003neural}.
Shortly after, Collobert \citeyear{collobert-weston-2007-fast}, \citeyear{collobert2008unified}, \citeyear{collobert2011natural} demonstrated that word embeddings could be incorporated into neural architectures for various NLP tasks. 
With word2vec, \citet{mikolov2013distributed} distilled the components of NLMs responsible for learning word embeddings into a lightweight and scalable solution, allowing this neural-based solution to be employed on corpora of unprecedented size (100B tokens).
Nevertheless, count-based solutions would still remain important, particularly GloVe \citep{pennington-etal-2014-glove}, as these methods were also significantly improved.
The next major improvement was the introduction of fastText \citep{bojanowski-etal-2017-enriching}, which was able to represent words absent from training data by leveraging subword information, as well as refining several aspects of word2vec's training method.

\paragraph{Sense Embeddings}

In spite of their success, word2vec, GloVe and fastText conflated different senses of the same word form into the same representation, a shortcoming known as the  Meaning Conflation Deficiency \citep{CamachoCollados2018FromWT}.
While a number of extensions were proposed for the creation of sense-specific representations, such as AutoExtend \citep{rothe-schutze-2015-autoextend}, NASARI \citep{CamachoCollados2016NasariIE}, DeConf \citep{pilehvar-collier-2016-de} or Probabilistic FastText \citep{athiwaratkun-etal-2018-probabilistic}, this issue would require the development of a new generation of NLMs in order to be effectively addressed.

\subsection{Neural Language Modelling}
\label{pre:nlm}

The first major step towards contextual embeddings from NLMs, was the development of context2vec \citep{melamud-etal-2016-context2vec}, a single-layer bidirectional LSTM trained with the objective of maximizing similarity between hidden states and target word embeddings, similarly to word2vec.
\citet{peters-etal-2018-deep} built upon context2vec with ELMo, a deeper bidirectional LSTM trained with language modelling objectives that produce more transferrable representations.
Both context2vec and ELMo emphasized WSD applications, providing the most convincing accounts until then that sense embeddings can be effectively represented as centroids of contextual embeddings, showing 1-NN solutions to WSD tasks that rivalled the performance of task-specific models.

With the introduction of highly-scalable Transformer architectures \citep{vaswani2017attention}, two kinds of very deep NLMs emerged: causal (or left-to-right) models, epitomized by the Generative Pre-trained Transformer \citep[GPT-3]{DBLP:journals/corr/abs-2005-14165}, where the objective is to predict the next word given a past sequence of words; and masked models, where the objective is to predict a masked (i.e., hidden) word given its surrounding words, of which the most prominent example is the Bidirectional Encoder Representations from Transformers \citep[BERT]{devlin-etal-2019-bert}.
The difference in training objectives results in these two varieties of NLMs specializing at different tasks, with causal models excelling at language generation and masked models at language understanding.\footnote{Although recent models like BART \citep{lewis-etal-2020-bart} show progress towards both.}

BERT proved highly successfully at most NLP tasks \citep{rogers-etal-2020-primer}, and motivated the development of numerous derivative models, many of which we also explore in this work.
In spite of this progress, Transformer-based NLMs can still show strong reliance on surface features \citep{mccoy-etal-2019-right} and social biases which are hard to correct \citep{zhou-etal-2021-challenges}.
There are known theoretical limits to how much language understanding can be expected from models trained with language modelling objectives alone \citep{bender-koller-2020-climbing,merrill2021provable}, and it is not clear how far current models are from those limits.

\subsection{Sense Inventories}
\label{pre:wn}

The currently most popular English word sense inventory is the Princeton WordNet \citep{Fellbaum2000WordNetA} (henceforth, WordNet), a large semantic network comprised of general domain concepts curated by experts\footnote{Babelnet \citep{NavigliPonzetto:10}, Wiktionary \citep{Meyer2012WiktionaryAN} and HowNet \citep{Dong2006HownetAT} are popular alternatives covering other languages.}.

The core unit of WordNet is the synset, which represents a cognitive concept.
Each lemma (word or multi-word expression) in WordNet belongs to one or more synsets, and word senses amount to the combination of word forms and synsets (referred as sensekeys).
As a result, the set of words that belong to a synset can be described as synonyms, with some words being ambiguous (belonging to additional synsets) while others not (specific to a synset).
The predominant semantic relation in WordNet, which relates synset pairs, is hypernymy (i.e., Is-A).
Each synset also features a gloss (dictionary definition), part-of-speech (noun, verb, adjective or adverb) and lexname\footnote{Lexnames are also known as supersenses \citep{flekova-gurevych-2016-supersense,pilehvar-etal-2017-towards}.}, which is a syntactic category and logical grouping.
Synsets are formally represented as numerical codes. Following related works, we also represent them using the more readable format $lemma_{POS}^{\#}$, where $lemma$ corresponds to synset's most representative lemma.

As an example, the lemma `mouse' is polysemous belonging to the $mouse_{n}^{1}$ (rodent) and $mouse_{n}^{4}$ (computer mouse) synsets, among others. The most frequent sense for mouse, mouse\%1:05:00:: (sensekey), belongs to the synset $mouse_{n}^{1}$ (02330245n) which has an hypernymy relation with $rodent_{n}^{1}$, lexname `noun.animal', and gloss ``any of numerous small rodents typically [...]".

Following \citet{loureiro-jorge-2019-language}, we use WordNet version 3.0, which contains 117,659 synsets, 206,949 senses, 147,306 lemmas, and 45 lexnames.


\section{Related Work}
\label{rel:intro}

In this section we cover related work on the various well-researched topics that our work intersects, namely Sense Embeddings ($\S$\ref{rel:embs}), WSD ($\S$\ref{rel:wsd}) and Probing NLMs ($\S$\ref{rel:lay}).

\subsection{Sense Embeddings}
\label{rel:embs}

Sense embeddings emerged in NLP due to the so-called meaning conflation deficiency of word embeddings \citep{CamachoCollados2018FromWT}. By merging several meanings into a single representation, the single vector proved insufficient in certain settings \citep{yaghoobzadeh-schutze-2016-intrinsic}, and contradicted common laws in distance metrics, such as the triangle inequality \citep{neelakantan-etal-2014-efficient}. In order to solve this issue, the field of sense vector representation mainly split into two categories: (1) unsupervised, where senses were learned directly from text corpora \citep{reisinger-mooney-2010-multi,huang-etal-2012-improving,vu-parker-2016-k}; (2) or knowledge-based, where senses were linked to a pre-defined sense inventory by exploiting an underlying knowledge resource \citep{rothe-schutze-2015-autoextend,pilehvar-collier-2016-de,mancini-etal-2017-embedding,colla-etal-2020-lesslex}.

In this article, we focus on the latter type of representation, particularly leveraging powerful Transformer-based language models trained on unlabeled text corpora. As such, the final representation is mainly constructed based on the knowledge learned by the language models, and knowledge resources such as WordNet serve to guide the annotation process. The goal of this paper is indeed to construct a task-agnostic sense representation that can be leveraged in semantic and textual applications. This differs from traditional static sense embeddings which, with a few notable exceptions \citep{li-jurafsky-2015-multi,flekova-gurevych-2016-supersense,pilehvar-etal-2017-towards}, were mainly leveraged in intrinsic sense-based tasks only. As we show throughout this paper, general-purpose sense representations learned with the power of Transformers and guided through an underlying lexical resource such as WordNet prove to be robust in a range of text-based semantic tasks, as well as in intrinsic sense-based benchmarks.
%

\clearpage

\subsection{Word Sense Disambiguation}
\label{rel:wsd}

As one of the earliest Artificial Intelligence tasks, WSD has a long history of research.
In this work, our coverage of related work for WSD is focused on recent systems using Transformer-based architectures for two reasons: our own experiments are also focused on Transformer-based systems; the current state-of-the-art for WSD has converged on these systems.
Additionally, we also distinguish between solutions addressing WSD from the nearest neighbors paradigm, using pre-computed sense embeddings, and task-specific solutions fine-tuning Transformer models or training classifiers using their internal representations.

\subsubsection{Nearest Neighbors}

Our prior LMMS work (described throughout this paper) was the first to demonstrate that a nearest neighbors solution based on sense embeddings pooled from internal representations of BERT (i.e., feature extraction) could clearly outperform the state-of-the-art of the time, which still had not adopted Transformer-based models.

SensEmBERT \citep{scarlinietal:2020} followed a similar approach to LMMS, but leveraged BabelNet to reduce dependency on annotated corpora, producing sense embeddings that performed better on WSD, though limited to nouns only.

With ARES, \citet{scarlini-etal-2020-contexts} introduce a method to produce a large number of semi-supervised annotations to dramatically increase the coverage of the sense inventory, and demonstrated that sense embeddings learned from those annotations can perform substantially better on WSD than LMMS.

SensEmBERT and ARES use the same layer pooling method and gloss embeddings as LMMS, although both have employed not only BERT-L, but also its multilingual variant, showing strong performance on languages other than English as well.

In addition to WSD, to our knowledge, the only other task these works have applied their sense embeddings is Word-in-Context \citep[WiC]{pilehvar-camacho-collados-2019-wic}, which we also address in this work.

\subsubsection{Trained Classifiers}

When it comes to using Transformers to train classifiers specific to the WSD task, we encounter a much more diverse set of solutions in comparison to feature extraction approaches.

One of the earliest and most straightforward supervised classifiers for WSD using BERT was the Sense Vocabulary Compression (SVC) of \citet{vial-etal-2019-sense}, which added layers to BERT, topped with a softmax classifier, to be trained targeting a strategically reduced set of admissible candidate senses. 

Following outstanding results on a range of text classification tasks by model fine-tuning, GlossBERT \citep{huang-etal-2019-glossbert} fine-tuned BERT using glosses so that WSD could be framed as a text classification task pairing glosses to words in context.
KnowBERT \citep{peters-etal-2019-knowledge} employs a more sophisticated fine-tuning approach, designed to exploit knowledge bases (WordNet and Wikipedia) as well as glosses.

Straying further from prototypical classifiers,  \citet{blevins-zettlemoyer-2020-moving} (BEM) propose a bi-encoder method which learns to represent senses based on glosses while performing the optimization jointly with the underlying BERT model.
Taking advantage of an ensemble of sense embeddings from LMMS and SensEmBERT, along with additional resources, EWISER  \citep{bevilacqua-navigli-2020-breaking} trains a multifaceted high performance WSD classifier.

Finally, the current state-of-the-art for WSD is ConSeC \citep{barba-etal-2021-consec}, which obtains impressively strong results by framing WSD as an extractive task, similar to extractive question answering, trained through fine-tuning BART \citep{lewis-etal-2020-bart}, a sequence-to-sequence Transformer which outperforms BERT on reading comprehension tasks (while being of comparable size).

In \citet{loureiro-etal-analysis-2021} we extensively compared fine-tuning and feature extraction approaches for the WSD task.
Consistent with prior work, we found that fine-tuning overall outperforms feature extraction.
However, under comparable circumstances, the performance gap is narrow and feature extraction shows improved few-shot performance and less frequency bias.

\clearpage

\subsection{Probing Neural Language Models}
\label{rel:lay}

As NLMs became popular, investigating properties of their internal states, or intermediate representations, also became an important line of research, often referred to as `model probing'.
Probing operates under the assumption that if a relatively simple classifier, based exclusively on representations from NLMs, can perform well at some task, then the required information was already encoded in the representations.
For clarity, we define probes as functions (learned or heuristic) designed to reveal some intrinsic property of NLMs.
In this section we cover probing works focused on lexical semantics and layer-specific variation that inspired our probing analysis.
We distinguish these works by their use of probes trained using representations (learned), and probes directly comparing or analysing unaltered representations (heuristics, such as nearest neighbors).

\subsubsection{Learned Probes}

Among the most influential findings in this line of research was the discovery by \citet{hewitt-manning-2019-structural} that syntactically valid parse trees could be uncovered from linear transformations of word representations obtained from pre-trained ELMo and BERT models.
Motivated by this discovery, \citet{coenen-etal-2019-visualizing} performed additional experiments focused on sense representation, including showing that a nearest neighbors based on BERT representations could outperform the reported WSD state-of-the-art, particularly when following \citet{hewitt-manning-2019-structural}'s methodology to learn a probe tailored to sense representation.
To increase sensitivity to sense-specific information, \citet{coenen-etal-2019-visualizing} used a loss that considered the difference between the average cosine similarity of embeddings of words with the same senses, and embeddings of words with different senses.
Both \citet{hewitt-manning-2019-structural} and \citet{coenen-etal-2019-visualizing} approaches are designed for probing representations obtained from single layers.

With ELMo, \citet{peters-etal-2018-deep} introduced contextualized word representations that are obtained from a linear combination of representations from all layers of the model.
This linear combination uses task-specific weights learned through an optimization process, often referred to in the literature as ``scalar mixing", and produced better results in downstream tasks when compared to representations obtained from individual layers.
On closer inspection, \citet{peters-etal-2018-dissecting} concluded that top layers can be less effective for semantic tasks possibly due to specialization for the language modelling tasks optimized during pre-training.

\citet{tenney-etal-2019-what} proposed an ``edge probing" methodology, using scalar mixing, that allowed for evaluating different syntactic or semantic properties using a common classifier architecture, where probing models are trained to predict graph edges independently.
In \citet{tenney-etal-2019-bert}, edge probing was employed to reveal that BERT implicitly performed different steps of a traditional NLP pipeline, in the expected order as information flows through the model, with lower layers processing local syntax (e.g., Part-of-Speech) and higher layers processing complex semantics of arbitrary distance (e.g., Semantic Roles).
Raising concerns about remaining faithful to the information encoded in the representations, \citet{kuznetsov-gurevych-2020-matter} proposes reducing the expressive power of learned probes while improving edge probing.

\citet{liu-etal-2019-linguistic} ran several probing experiments with simpler probes (i.e., linear classifiers), investigating differences between NLM architectures, namely ELMo, GPT and BERT, while still finding competitive performance with state-of-the-art task-specific models.
They confirm that LSTM-based models (i.e., ELMo) present more task-specific (less transferable) top layers, but Transformers-based models (i.e., BERT) are less predictable and do not exhibit monotonic increase in task-specificity, in line with our own findings.
GPT was found to significantly underperform ELMo and BERT, which \citet{liu-etal-2019-linguistic} attributes to the fact that GPT is trained unidirectionally (left-to-right), while ELMo and BERT are trained bidirectionally.

\subsubsection{Representational Similarity}

Without recourse to learned probes, \citet{ethayarajh-2019-contextual} investigated differences between ELMo, GPT-2 and BERT, relying on experiments based on cosine similarity to learn about the context-specificity of their representations.
\citet{ethayarajh-2019-contextual} found that top layers show highest degree of context-specificity, but all layers of all three models produced highly anisotropic representations, with directions in vector space confined to a narrow cone, concluding that this property is an inherent consequence of the contextualization process.
The anisotropy observed for all contextualized NLMs also supports the hypothesis of \citet{coenen-etal-2019-visualizing} that sense-level information is encoded in a low-dimensional subspace, since contextualization is crucial for sense disambiguation.

\citet{vulic-etal-2020-probing} reached similar conclusions regarding the detrimental contribution of top layers for lexical tasks (e.g., lexical semantic similarity) 
while also finding improved results from averaging different layers, particularly task-specific layer subsets, prompting further research into layer weighting or meta-embedding approaches, and motivating the present work.
Through direct comparison of cosine similarities, \citet{chronis-erk-2020-bishop} reached similar conclusions as \citet{vulic-etal-2020-probing} about the role of top layers for lexical similarity tasks, adding that top layers appear to better approximate relatedness than similarity.

\citet{voita-etal-2019-bottom} probed Transformer-based NLMs from an Information-Bottleneck perspective to learn about differences in information flow across the network according to language modelling pre-training objectives, particularly left-to-right, MLM, and translation.
They find that the MLM objective induces representation of token identity in the lower layers, followed by a more generalized token representation in intermediate layers, and then token identity information gets recreated at top layers.


\citet{mickus-etal-2020-what} specifically verified whether BERT representations comprise a coherent semantic space.
These experiments are explicitly detached from learned probes, as \citet{mickus-etal-2020-what} explains that such methodology interferes with direct assessment of the coherence of the semantic space as produced by NLMs.
Using cluster analyses, they find that BERT indeed appears to represent a coherent semantic space (based only on representations from the final layer), although its Next Sentence Prediction (NSP) modelling objective leads to encoding semantically irrelevant information (sentence position), corrupting similarity relationships and complicating comparisons with other NLMs.

\clearpage

\section{Method}
\label{met:intro}

We propose a principled approach for sense representation based on contextual NLMs trained exclusively with self-supervision.
This approach is an extension of \cite{loureiro-jorge-2019-language}, addressing relevant issues still largely unresolved, particularly the influence of embeddings from the different layers composing NLMs, with the introduction of a novel layer probing methodology.
Moreover, in this work, we reinforce the distinction between sense disambiguation and sense matching by introducing methodological differences specific to each application scenario.

This section starts by explaining the methods used in \cite{loureiro-jorge-2019-language} for learning ($\S$\ref{met:embed}), extending ($\S$\ref{met:ext}) and applying sense embeddings ($\S$\ref{met:apply}).
Afterwards, we introduce our proposed layer probing methodology ($\S$\ref{met:lay}), including how the resulting analysis informs a grounded pooling operation for combining embeddings from all layers of a NLM.

\subsection{Learning Sense Embeddings}
\label{met:embed}

\begin{figure}[htb]
  \centering
  \includegraphics[width=1.0\textwidth]{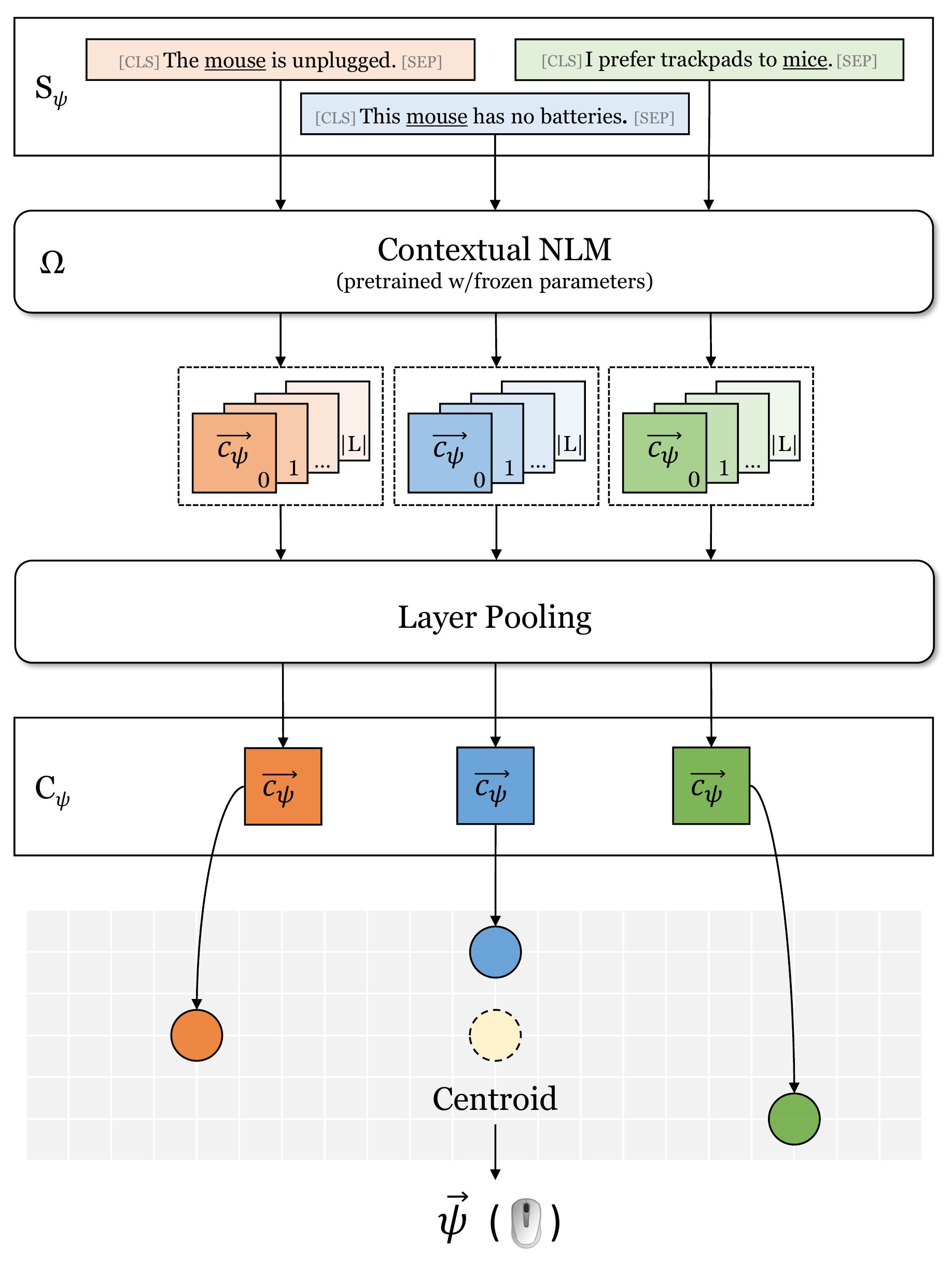}
  \caption{Overview of learning sense embeddings from annotated corpora. Showing how the sense $\psi$ for `computer mouse' is determined from a set for sentences annotated with that sense $S_\psi$ (padded with special tokens as expected by the NLM $\Omega$). After pooling contextual embeddings $C_\psi$ from layers $L$, the sense embedding for $\vec{\psi}$ is computed as the centroid of $C_\psi$.}
  \label{fig:embed}
\end{figure}

The initial process to learn sense embeddings is based on sense-annotated sentences and contextualized embeddings of annotated words or phrases in context.
An overview of the process can be seen at Figure \ref{fig:embed}.

Formally, in order to generate sense embeddings learned in context from natural language, we require a pre-trained contextual NLM $\Omega$ (frozen parameters) and a corpus of sense-annotated sentences $S$.
Every sense $\psi$ is represented from the set of contextual embeddings $\vec{c}_l \in C_\psi$, obtained by employing $\Omega$ on the set of sentences $S_\psi$ annotated with that sense (considering only contextual embeddings specific to tokens annotated with sense $\psi$), using representations at each layer $l \in L$, such that:

\begin{equation}
\vec{\psi} = \frac{1}{|C_\psi|}\sum_{l \in L}^{}\sum_{\vec{c} \in C_\psi}^{}\vec{c}_l\textit{ , where }C_\psi = \Omega(S_\psi)
\end{equation}

The $L$ set of layers typically used for sense representation is the last four $[-1, -2, -3, -4]$ (reversed layer indices), as discussed in Section \ref{rel:embs}.

\clearpage

\clearpage

Contextual NLMs typically operate at the subword-level, so the token-level embeddings $\vec{c}$ produced by $\Omega$ correspond to the average of each token's subword contextual embeddings (depending on the NLM, these may be BPE or WordPiece embeddings).
Similarly, whenever sense-annotations cover a span of several tokens, we also use the average of the corresponding token-level embeddings as the contextual embedding.
Contextual NLMs are pre-trained using special tokens at specific locations, so we also include these tokens in their expected positions (e.g., \texttt{[CLS]} at the start and \texttt{[SEP]} at the end with BERT).

As described in Section \ref{pre:wn}, WordNet can be used to represent senses in two ways: sensekeys and synsets.
Sense-annotated corpora most often use sensekey annotations, so in those cases sensekey embeddings do not require any intermediate mapping.
Synset embeddings can be derived from sensekey annotations in at least two ways, which we differentiate as `direct' and `indirect'.
In the direct approach, each sensekey annotation is converted (mapped) to the corresponding synset, so synset representations are learned from each annotation instance.
In the indirect approach we first learn sensekey-level embeddings, without converting annotations, and afterwards compute synset embeddings as the average of corresponding sensekey embeddings.
The latter approach has been explored in earlier works in sense embeddings \citep{rothe-schutze-2015-autoextend}.
In this work we explore both approaches.

\subsection{Extending Coverage with Additional Resources}
\label{met:ext}

Given that one of the major issues in supervised WSD is the lack of sense annotations \citep{pasini2020knowledge}, not just in their quantity but also in terms of their coverage of the sense inventory, we require solutions to address this in our method.
On this matter, we also follow the methods we first proposed in \cite{loureiro-jorge-2019-language} and later optimized with the introduction of the UWA corpus in \cite{loureiro-camacho-collados-2020-dont}.
The two methods, ontological propagation and gloss representation, are designed to reach full coverage of the sense inventory, and they are complementary  by exploiting different resources, namely semantic relations between senses and glosses (combined with lemmas).

\subsubsection{Ontological Propagation}
\label{met:prop}

In Section \ref{pre:wn} we introduced WordNet and the different elements and relations composing this semantic network.
The ontological propagation method we presented in \cite{loureiro-jorge-2019-language} exploits these relations between senses in WordNet in order to infer embeddings for senses which may not occur in annotated corpora.
It is possible to infer accurate sense embeddings from these relations due to the fine-granularity of WordNet, along with widespread synonymy and hypernymy relations, to the extent that in \cite{loureiro-camacho-collados-2020-dont} we showed that even annotations for unambiguous words can significantly improve the propagation process.

Since available corpora do not provide full-coverage annotations for our sense inventory of interest, by following the process described in Section \ref{met:embed} we are left with a represented senses $\Psi$, and a set of unrepresented senses $\Psi'$.
The propagation process involves three steps, using increasingly abstract relations from WordNet - sets of synonyms (synsets), hypernymy relations, and lexical categories (lexnames or supersenses).
In case we are targeting synset-level representations, then the first step/level should be skipped.

Considering we are provided mappings between sensekeys, synsets, hypernyms and lexnames, we infer $\Psi'$ iteratively following Algorithm \ref{algo:prop}.
After each of these sequential steps, every inferred $\vec{\psi}$  is added to the set of represented senses $\Psi$. This propagation method ensures full-coverage provided that initial sense embeddings $\Psi$ are sufficiently diverse such that falling back on propagating from lexnames (supersenses) is always possible.

\begin{algorithm}
  \SetKwFunction{Insert}{Insert}
  \SetKwFunction{Remove}{Remove}
  \SetKwProg{Propagate}{Propagate}{}{}
  \Propagate{$(\Psi,\Psi',R)$}{
    \ForEach{unrepresented sense $ \psi' \in \Psi'$}{

      $R_{\psi'} \gets \{ \text{all represented } \vec{\psi} \in \Psi \text{ for which } (\psi, \psi') \in R \}$;
      
      \If{$|R_{\psi'}| > 0$}{
        $\vec{\psi'} \gets \text{average of sense embeddings in } R_{\psi'}$\;
        \Insert{$\vec{\psi'}, \Psi$}\tcp*[r]{\small{add to represented}}
        \Remove{$\psi', \Psi'$}\tcp*[r]{\small{remove from unrepresented}}
      }
    }
  \KwRet{$\Psi, \Psi'$}\;
  }
  \BlankLine

  $\Psi, \Psi' \gets$ \Propagate{$(\Psi, \Psi'$, \{all $(\psi, \psi')$ \textup{:} \textup{\texttt{{\small Synset}}}$(\psi)$ \textup{=} \textup{\texttt{\small Synset}}$(\psi')$\}$)$}
  
  $\Psi, \Psi' \gets$ \Propagate{$(\Psi, \Psi'$, \{all $(\psi, \psi')$ \textup{:} \textup{\texttt{{\small Hypernym}}}$(\psi)$ \textup{=} \textup{\texttt{\small Hyper.}}$(\psi')$\}$)$}
  
  $\Psi, \Psi' \gets$ \Propagate{$(\Psi, \Psi'$, \{all $(\psi, \psi')$ \textup{:} \textup{\texttt{{\small Lexname}}}$(\psi)$ \textup{=} \textup{\texttt{\small Lexname}}$(\psi')$\}$)$}

\caption{Propagation method to infer unrepresented senses $\Psi'$, using sense embeddings $\Psi$ learned from annotations, and relations $R$.}
\label{algo:prop}
\end{algorithm}




Since this method is designed to achieve full-representation of the sense inventory based on a subset of senses observed in context, the inferred representations are also of a similar contextual nature.
However, unless the initial set of sense embeddings is nearly complete, and particularly diversified, this propagation method produces some number of identical representations for distinct senses, which is most undesirable for disambiguation applications, and to a lesser extent, sense matching applications as well.

\subsubsection{Leveraging Glosses and Lemmas}
\label{met:gls}

In \cite{loureiro-jorge-2019-language} we introduced a method for representing sense embeddings based on glosses and lemmas.
This method is inspired by a typical baseline approach used in works pertaining to sentence embeddings, and it amounts to simply averaging the contextual embeddings for all tokens present in a sentence.
In our case, we use glosses as sentences, but also introduce lemmas into the gloss' context.
By combining glosses with lemmas, we not only augment the information available to represent senses, but we are also able to generate sense embeddings which are lemma-specific (sensekey-level), instead of only concept-specific (synset-level) if we only used glosses.
As such, sense embeddings generated by this method address the redundancy issue arising from the previously described propagation method, while simultaneously introducing representational information which is complementary to contextual embeddings extracted from sense-annotated sentences.

The method proceeds as follows.
For every lemma/sense pair (i.e., sensekey) in a sense inventory, we build the template ``$<$\textit{lemma}$>$ , $<$\textit{sense lemmas}$>$ - $<$\textit{sense gloss}$>$".
For instance, based on WordNet, the synset race$^v_2$ has the lemmas \textit{race} and \textit{run} which are provided with the following sensekey-specific fill-outs of the template:

\begin{itemize}
    \item \textit{race\%2:33:00::} - ``race - run, race - compete in a race"
    \item \textit{run\%2:33:01::} - ``run - run, race - compete in a race"
\end{itemize}

The initial ``$<$\textit{lemma}$>$" component of the template can be omitted if the target representation level is synsets, as it only serves the purpose of reinforcing the lemma which is specific to the sensekey.
The templated string is processed by $\Omega$, similarly to sentences S in Section \ref{met:embed}, but here we use the resulting set of contextual embeddings for every token $C_\star$.

Considering that we have a complete set of sense embedding $\Psi$, based on sense annotations and propagation as described in Sections \ref{met:embed} and \ref{met:prop}, we augment $\forall \vec{\psi} \in \Psi$ with gloss and lemma information as follows:

\begin{equation}
\vec{\psi} =  \frac{1}{2} ( ||\vec{\psi}||_2 + ||\frac{1}{|C_\star|}\sum_{l \in L}^{}\sum_{\vec{c} \in C_\star}^{}\vec{c}_l||_2)\textit{ , where }C_\star = \Omega(\text{Template}(\psi))
\tag{3}
\end{equation}

In contrast to \cite{loureiro-jorge-2019-language}, which proposed using concatenation to merge this new set of sense embeddings based on glosses and lemmas with the previously mentioned set, in this work we propose merging through averaging instead.
This departure is motivated by the fact that \cite{loureiro-jorge-2019-language} found that while concatenation outperformed averaging for WSD, the difference in performance was modest, and in this work we are interested in additional tasks which that work did not cover.
Merging representations through concatenation doubles the dimensionality of sense embeddings, increasing computational requirements and complicating comparison with contextual embeddings, among other potential applications.
On the other hand, merging representations through averaging allows for adding more components while retaining a similar vector, of equal dimensionality to contextual embeddings, and represented in the same vector space.

\subsection{Applying Sense Embeddings}
\label{met:apply}

In this section we address how sense embeddings can be employed for solving various tasks, grouped under two paradigms: disambiguation and matching.
Disambiguation assigns a word in context (i.e., in a sentence) to a particular sense out of a subset of candidate senses, restricted by the word's lemma and part-of-speech.
Matching also assigns specific senses to words, but imposes no restrictions, admitting every entry in the sense inventory for each assignment.

The different conditions for disambiguation and matching require sense representations with different degrees of lexical information and semantic coherence.
Whereas, for disambiguation, lexical information can be absent from sense representations, due to the subset restrictions, for matching, lexical information is essential to distinguish between word forms carrying identical or similar semantics.
Similarly, the disambiguation setting has no issues with sense representations displaying inconsistencies such as \textit{eat} being more similar to \textit{sleep} than to \textit{drink}, since these all belong to disjoint subsets, but the order and coherence of these similarities is relevant for sense matching applications.
This distinction leads us to specialize sense embeddings accordingly in Section \ref{met:lay}.

To disambiguate a word $w$ in context, we start by creating a set $\Psi_w$ of candidate senses based on its lemma and part-of-speech, using information provided with the sense inventory.
Afterwards, we compute the cosine similarities (denoted `cos') between the word's contextual embedding $\vec{c}_w$ and the pre-computed embeddings for each sense in this subset $\Psi_w$ (both using the same layer pooling).
Finally, we assign the sense whose similarity is highest (i.e., nearest neighbor):
\begin{gather*}
    \Psi_w = \{\vec{\psi} \in \Psi : \text{lemma and part-of-speech of  } \psi \text{  match  } w \}\\
    \text{Disambiguation}(w) = \argmax_{\vec{\psi} \in \Psi_w}(\text{cos}(\vec{c}_w, \vec{\psi}))
\tag{4}
\end{gather*}

To match a word $w$ in context, without restrictions, we follow the approach for disambiguation but simply consider the full sense inventory $\Psi$ instead of $\Psi_w$:
\begin{gather*}
    \text{Matching}(w) = \argmax_{\vec{\psi} \in \Psi}(\text{cos}(\vec{c}_w, \vec{\psi}))
\tag{5}
\end{gather*}

\subsection{Grounding Layer Pooling}
\label{met:lay}

Up until this point, we have described our method closely following our prior work in \cite{loureiro-jorge-2019-language}.
As we covered in earlier sections, NLMs can show substantial and, more importantly, unexpected variation in task performance across their layers.
Considering this work's focus on a more principled and grounded focus on sense representation with NLMs, our methodology also covers this important aspect.

In this section we present two methods targeting the layers composing NLMs.
The first method probes each layer's adeptness for sense representation.
Consequently, the second method in this section is designed to capitalize on that knowledge towards sense representations which better capture NLM's ability to represent senses over the current paradigm.

As we alluded to in Section \ref{met:apply}, sense representation should be viewed in light of the intended applications for these representations.
In particular, in this work we differentiate between representations used for disambiguating words, and for matching or comparing senses.
This distinction is motivated by the fact that disambiguation, which is the prevalent sense-related task on NLP, only requires that sense representations be adequately differentiated between the restricted set of senses which share the same lemmas and parts-of-speech.
However, there exist other potential applications where sense representations are matched without any constraints on the sense inventory, and thus require that senses be coherently represented across the semantic space.

\subsubsection{Sense Probing}
\label{met:prob}

In order to assess the contribution of individual layers of a pre-trained NLM for sense representation, we directly evaluate the performance of representations from these layers on tasks related to the previously described disambiguation or matching scenarios.
These tasks are solved using the nearest neighbors approaches described in Section \ref{met:apply}, comparing pre-computed sense embeddings with contextual embeddings obtained from the same layer.

For this probing experiment, we follow the method for learning sense representations described in Section \ref{met:embed}, but create multiple sets of senses $\Psi_l$ for each layer $l$ in the NLM. 
To maintain focus on assessing the performance of representations learned directly from specific layers, we ensure that test instances have all their senses represented in the sense-annotated corpora used to precompute $\Psi_l \text{,   } \forall l \in L$.
Thus, our probing experiments do not use techniques to infer or enrich sense representations, such as those we described in Section \ref{met:apply}, which could otherwise act as confounders.

The resulting performance scores for every layer $l \in L$ composing a specific $\Omega$, using a corresponding $\Psi_l$, not only reveal which layers perform best, but also inform the layer pooling method described next.

\subsubsection{Sense Profiling}
\label{met:prof}

We use the probing results described earlier as the basis for a pooling operation which is better grounded than the current paradigm of using the sum of the last four layers, and also better performing as we show later in this work.
We designate each set of model-specific layer weights as a `sense profile', and consider distinct sense profiles for disambiguation and matching, depending on the choice of disambiguation or matching tasks during layer probing.

These proposed sense profiles are a more immediate version of the Scalar Mixing used in \citet{tenney-etal-2019-bert}, being based on heuristically-derived sets of layer weights, instead of learning them through task optimization.
Considering this, we understand sense profiles to be closer to the extraction configurations of \citet{vulic-etal-2020-probing}.

Granted we have performance scores $s_l \text{   } \forall l \in L$, for a specific $\Omega$, we obtain layer specific weights $w_l \text{   } \forall l \in L$ by applying the softmax function:

\begin{equation}
w_l = \frac{\text{exp}(s_l/t)}{\sum_{l' \in L}^{}\text{exp}(s_{l'}/t)}
\tag{4}
\end{equation}

We use the temperature scaling parameter $t$ to skew the weight distribution towards highest performing layers.
While simple, temperature scaling has been found surprisingly effective at calibrating neural network predictions \citep{pmlr-v70-guo17a}.
This parameter is to be determined empirically and is only specific to application settings, not models.

In Table \ref{tab:weights_layers_f1s} we demonstrate the interaction between performance scores and layer weights conditioned on the temperature parameter.
In that table, and others found in this work, we use reverse layer indices so that we can consistently refer to the final layer of any model using the -1 index, regardless of the number of layers in the NLM.

Consequently, we employ sense profiles comprised of weights $w_l \text{   } \forall l \in L$ to retrieve contextual embeddings from $\Omega$, and generate our sense embeddings accordingly, updating formula (1) such that:

\begin{equation}
\vec{\psi} = \frac{1}{|C_\psi|}\sum_{l \in L}^{}\sum_{\vec{c} \in C_\psi}^{}w_l * \vec{c}_l\textit{ , where }C_\psi = \Omega(S_\psi)
\tag{5}
\end{equation}

This set of sense embeddings learned from annotations using sense profiles, undergoes the same extensions and augmentations described earlier ($\S$\ref{met:ext}).

To be clear, the process of probing layer performance and determining sense profiles to pool contextual embeddings from all layers (including when learning sense embeddings from annotations) is carried out for both the disambiguation and matching settings independently.
As a result, we produce two sets of sense embeddings for each NLM based on sense profiles, which we distinguish from the LMMS sense embeddings introduced in \citet{loureiro-jorge-2019-language} as LMMS-SP (Sense Profiles).
The SP-WSD (for Word Sense Disambiguation) and SP-USM (for Uninformed Sense Matching) abbreviations are used to refer to sense embeddings based on disambiguation and matching sense profiles respectively.

\begin{table}[ht]
\centering
\resizebox{0.98\textwidth}{!}{%
\begin{tabular}{p{6em}>{\centering\arraybackslash}p{1.3em}>{\centering\arraybackslash}p{1.3em}>{\centering\arraybackslash}p{1.3em}>{\centering\arraybackslash}p{1.3em}>{\centering\arraybackslash}p{1.3em}>{\centering\arraybackslash}p{1.3em}>{\centering\arraybackslash}p{1.3em}>{\centering\arraybackslash}p{1.3em}>{\centering\arraybackslash}p{1.3em}>{\centering\arraybackslash}p{1.3em}>{\centering\arraybackslash}p{1.3em}>{\centering\arraybackslash}p{1.3em}>{\centering\arraybackslash}p{1.3em}>{\centering\arraybackslash}p{1.3em}>{\centering\arraybackslash}p{1.3em}>{\centering\arraybackslash}p{1.3em}>{\centering\arraybackslash}p{1.3em}>{\centering\arraybackslash}p{1.3em}>{\centering\arraybackslash}p{1.3em}>{\centering\arraybackslash}p{1.3em}>{\centering\arraybackslash}p{1.3em}>{\centering\arraybackslash}p{1.3em}>{\centering\arraybackslash}p{1.3em}>{\centering\arraybackslash}p{1.3em}>{\centering\arraybackslash}p{1.3em}} \toprule
 & \textbf{\rotatebox{90}{\scriptsize{INIT}}} & \textbf{\rotatebox{90}{-24}} & \textbf{\rotatebox{90}{-23}} & \textbf{\rotatebox{90}{-22}} & \textbf{\rotatebox{90}{-21}} & \textbf{\rotatebox{90}{-20}} & \textbf{\rotatebox{90}{-19}} & \textbf{\rotatebox{90}{-18}} & \textbf{\rotatebox{90}{-17}} & \textbf{\rotatebox{90}{-16}} & \textbf{\rotatebox{90}{-15}} & \textbf{\rotatebox{90}{-14}} & \textbf{\rotatebox{90}{-13}} & \textbf{\rotatebox{90}{-12}} & \textbf{\rotatebox{90}{-11}} & \textbf{\rotatebox{90}{-10}} & \textbf{\rotatebox{90}{-9}} & \textbf{\rotatebox{90}{-8}} & \textbf{\rotatebox{90}{-7}} & \textbf{\rotatebox{90}{-6}} & \textbf{\rotatebox{90}{-5}} & \textbf{\rotatebox{90}{-4}} & \textbf{\rotatebox{90}{-3}} & \textbf{\rotatebox{90}{-2}} & \textbf{\rotatebox{90}{-1}} \\ \midrule \midrule
BERT-L & \cellcolor[HTML]{FFFFFF}{\large 53} & \cellcolor[HTML]{FFFFFF}{\large 58} & \cellcolor[HTML]{FFFFFF}{\large 62} & \cellcolor[HTML]{FFFFFF}{\large 63} & \cellcolor[HTML]{FFFFFF}{\large 65} & \cellcolor[HTML]{FFFFFF}{\large 67} & \cellcolor[HTML]{FFFFFF}{\large 68} & \cellcolor[HTML]{FFFFFF}{\large 68} & \cellcolor[HTML]{FFFFFF}{\large 69} & \cellcolor[HTML]{FFFFFF}{\large 70} & \cellcolor[HTML]{FFFFFF}{\large 71} & \cellcolor[HTML]{FFFFFF}{\large 71} & \cellcolor[HTML]{FFFFFF}{\large 71} & \cellcolor[HTML]{FFFEFD}{\large 71} & \cellcolor[HTML]{FEF5EC}{\large 72} & \cellcolor[HTML]{FEF1E4}{\large 72} & \cellcolor[HTML]{FFFEFD}{\large 71} & \cellcolor[HTML]{FDEDDC}{\large 72} & \cellcolor[HTML]{FDEBD9}{\large 73} & \cellcolor[HTML]{FDEDDC}{\large 72} & \cellcolor[HTML]{FCE5CE}{\large 73} & \cellcolor[HTML]{FBDBBB}{\large 74} & \cellcolor[HTML]{FACEA2}\textbf{{\large 75}} & \cellcolor[HTML]{F9CB9C}\textbf{{\large 75}} & \cellcolor[HTML]{FFFAF4}{\large 72} \\
XLNet-L & \cellcolor[HTML]{FFFFFF}{\large 51} & \cellcolor[HTML]{FFFFFF}{\large 57} & \cellcolor[HTML]{FFFFFF}{\large 65} & \cellcolor[HTML]{FFFFFF}{\large 67} & \cellcolor[HTML]{FFFFFF}{\large 68} & \cellcolor[HTML]{FFFFFF}{\large 70} & \cellcolor[HTML]{FFFFFF}{\large 71} & \cellcolor[HTML]{C1DAF1}{\large 72} & \cellcolor[HTML]{9FC5E8}{\large \textbf{73}} & \cellcolor[HTML]{ABCCEB}{\large \textbf{73}} & \cellcolor[HTML]{BCD7EF}{\large 72} & \cellcolor[HTML]{CDE1F3}{\large 72} & \cellcolor[HTML]{DEEBF7}{\large 72} & \cellcolor[HTML]{D2E4F5}{\large 72} & \cellcolor[HTML]{F4F9FD}{\large 71} & \cellcolor[HTML]{FFFFFF}{\large 71} & \cellcolor[HTML]{FFFFFF}{\large 71} & \cellcolor[HTML]{FFFFFF}{\large 71} & \cellcolor[HTML]{FFFFFF}{\large 71} & \cellcolor[HTML]{FFFFFF}{\large 71} & \cellcolor[HTML]{C1DAF1}{\large 72} & \cellcolor[HTML]{BCD7EF}{\large 72} & \cellcolor[HTML]{C1DAF1}{\large 72} & \cellcolor[HTML]{BCD7EF}{\large 72} & \cellcolor[HTML]{FFFFFF}{\large 68} \\
RoBERTa-L & \cellcolor[HTML]{FFFFFF}{\large 53} & \cellcolor[HTML]{FFFFFF}{\large 57} & \cellcolor[HTML]{FFFFFF}{\large 63} & \cellcolor[HTML]{FFFFFF}{\large 66} & \cellcolor[HTML]{FFFFFF}{\large 67} & \cellcolor[HTML]{FFFFFF}{\large 69} & \cellcolor[HTML]{FFFFFF}{\large 71} & \cellcolor[HTML]{FFFFFF}{\large 72} & \cellcolor[HTML]{FFFFFF}{\large 73} & \cellcolor[HTML]{FFFFFF}{\large 73} & \cellcolor[HTML]{FFFFFF}{\large 74} & \cellcolor[HTML]{F4F2F9}{\large 74} & \cellcolor[HTML]{EEEBF6}{\large 74} & \cellcolor[HTML]{D7D0E9}{\large 74} & \cellcolor[HTML]{B4A7D6}{\large \textbf{75}} & \cellcolor[HTML]{B4A7D6}{\large \textbf{75}} & \cellcolor[HTML]{C6BCE0}{\large \textbf{75}} & \cellcolor[HTML]{DDD7ED}{\large 74} & \cellcolor[HTML]{CCC3E3}{\large \textbf{75}} & \cellcolor[HTML]{E3DEF0}{\large 74} & \cellcolor[HTML]{E8E4F3}{\large 74} & \cellcolor[HTML]{DDD7ED}{\large 74} & \cellcolor[HTML]{FFFFFF}{\large 73} & \cellcolor[HTML]{D1C9E6}{\large 74} & \cellcolor[HTML]{FFFFFF}{\large 71} \\
ALBERT-XL & \cellcolor[HTML]{FFFFFF}{\large 54} & \cellcolor[HTML]{FFFFFF}{\large 65} & \cellcolor[HTML]{FFFFFF}{\large 67} & \cellcolor[HTML]{FFFFFF}{\large 68} & \cellcolor[HTML]{FFFFFF}{\large 69} & \cellcolor[HTML]{FCFEFB}{\large 70} & \cellcolor[HTML]{EAF4E5}{\large 70} & \cellcolor[HTML]{D4E8CB}{\large \textbf{71}} & \cellcolor[HTML]{C9E1BE}{\large \textbf{71}} & \cellcolor[HTML]{C5E0BA}{\large \textbf{71}} & \cellcolor[HTML]{BEDCB1}{\large \textbf{71}} & \cellcolor[HTML]{BADAAD}{\large \textbf{71}} & \cellcolor[HTML]{B6D7A8}{\large \textbf{71}} & \cellcolor[HTML]{BEDCB1}{\large \textbf{71}} & \cellcolor[HTML]{D0E6C7}{\large \textbf{71}} & \cellcolor[HTML]{DBEBD4}{\large 70} & \cellcolor[HTML]{F8FCF7}{\large 70} & \cellcolor[HTML]{FFFFFF}{\large 69} & \cellcolor[HTML]{FFFFFF}{\large 69} & \cellcolor[HTML]{FFFFFF}{\large 69} & \cellcolor[HTML]{FFFFFF}{\large 69} & \cellcolor[HTML]{FFFFFF}{\large 69} & \cellcolor[HTML]{FFFFFF}{\large 69} & \cellcolor[HTML]{FFFFFF}{\large 69} & \cellcolor[HTML]{FFFFFF}{\large 64} \\ \midrule
\includegraphics[width=250mm, height=175mm]{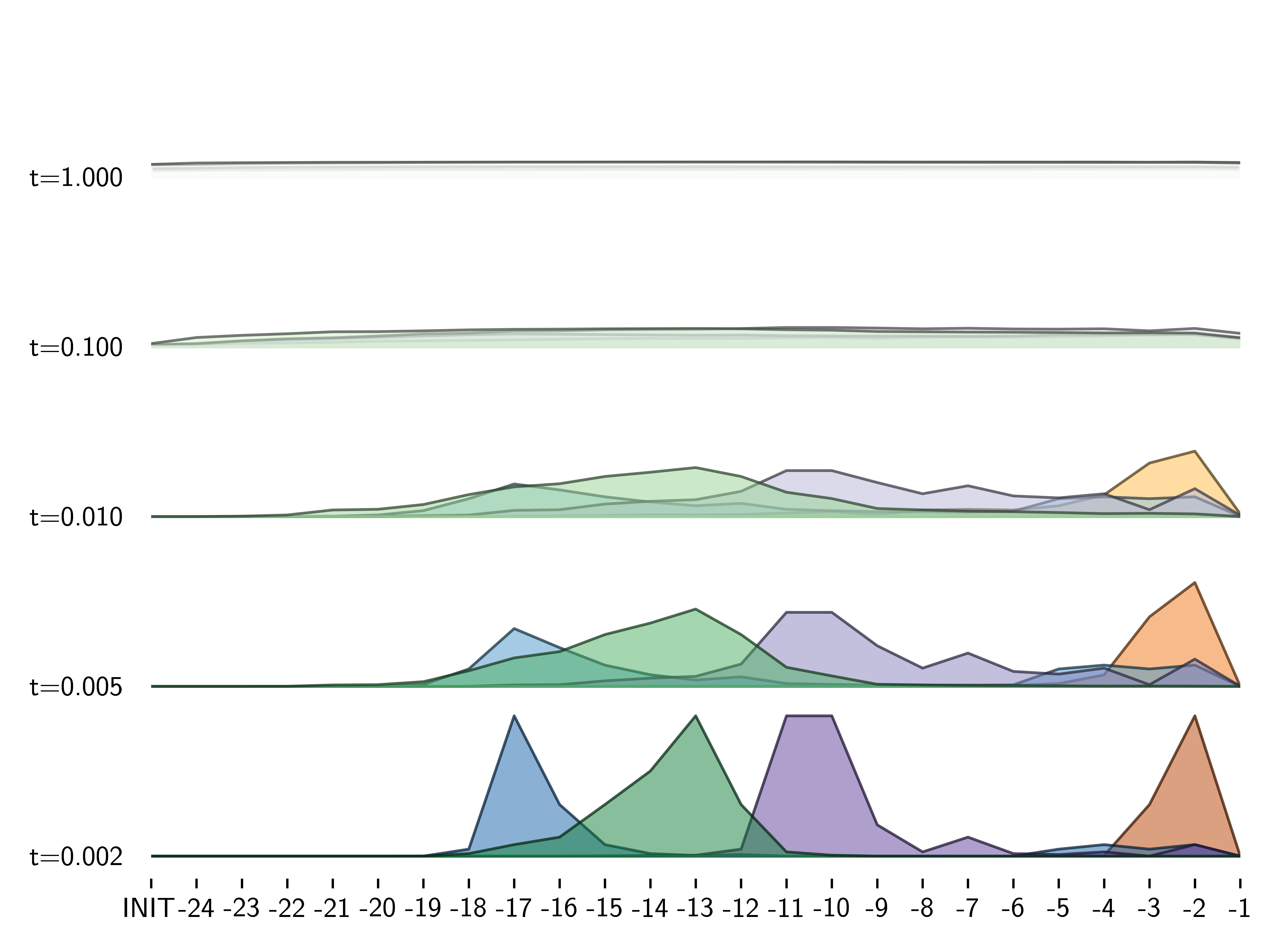}\\
\vspace{-50mm}\\
\end{tabular}%
}
\caption{Shows interaction between F1 scores (rounded) for 1NN WSD using each layer of four different NLMs, and respective weight distributions (matching colors) using decreasing temperature parameters. Lower temperatures induce higher skewness towards layers that perform best on the probing validation set. Distributions based on t=1.000 are almost uniform, while t$<$0.002 would place almost all mass on single best layer.}
\label{tab:weights_layers_f1s}
\end{table}

\clearpage

\section{Experimental Setting}
\label{exp:intro}

In this section we provide details about our experimental setting, including a description of the models ($\S$\ref{exp:models}) and datasets ($\S$\ref{exp:datasets}) used for learning sense representations.

\subsection{Transformer-based Language Models}
\label{exp:models}

In this work we experiment with several Transformer-based Language Models, including all the original English BERT models released by \citet{devlin-etal-2019-bert} as well as several other BERT-inspired alternatives, namely XLNet \citep{yang2019xlnet}, RoBERTa \citep{DBLP:journals/corr/abs-1907-11692} and ALBERT \citep{Lan2020ALBERT:}.
This section briefly describes the most relevant features of each of these models for our use case.
We summarize the differences between each variant of these models on Table \ref{tab:nlm_details}.

\paragraph{\textbf{BERT}}
The model released by \citet{devlin-etal-2019-bert} is first prominent Transformer-based NLM designed for language understanding.
It is pre-trained with two unsupervised modelling objectives, Masked Language Modelling (MLM) and Next Sentence Prediction (NSP), using English Wikipedia and BookCorpus \citep{zhu2015aligning}.
It uses WordPiece tokenization, splitting words into different components at the character-level (i.e., subwords).
BERT is available in several models differing not only on parameter size, but also tokenization and casing.
The `whole-word' models were released after publication, showing slightly improved benchmark performance when trained with whole words being masked instead of subwords resulting from WordPiece tokenization.

\paragraph{\textbf{XLNet}}
Based on a Transfomer-XL \citep{dai-etal-2019-transformer} architecture, \citet{yang2019xlnet} release XLNet featuring Permutation Language Modelling (PLM) as the only pre-training objective.
The motivation for PLM is that it does not rely on masked tokens, and thus makes pre-training closer to fine-tuning for downstream tasks.
It is also trained on much larger corpora than BERT, adding a large volume of web text from various sources to the corpora used for BERT.
Instead of using WordPiece for tokenization, XLNet uses SentencePiece \citep{kudo-richardson-2018-sentencepiece}, which is a very similar open-source version of WordPiece.

\paragraph{\textbf{RoBERTa}}
The model proposed by \citet{DBLP:journals/corr/abs-1907-11692} is explicitly designed as an optimized version of BERT.
RoBERTa does not use the NSP pre-training objective after finding that it deteriorates performance in the reported experimental setting, performing only MLM during pre-training.
It is also trained with some different choices of hyperparameters (e.g., larger batch sizes) that improve performance on downstream tasks.
The models released with RoBERTa are also trained on larger corpora composed mostly of web text, similarly to XLNet.
As for tokenization, RoBERTa opts for byte-level BPE, following \citet{radford2019language}, which makes retrieving embeddings for specific tokens more challenging (i.e., spacing must be explicitly encoded).

\paragraph{\textbf{ALBERT}}
Aiming for a lighter architecture, \citet{Lan2020ALBERT:} propose ALBERT as a more parameter-efficient version of BERT.
In spite of changes introduced to improve efficiency (e.g., cross-layer parameter sharing), ALBERT is based on a similar architecture to BERT.
Besides improving efficiency, ALBERT also improves performance on downstream tasks by replacing NSP with the more challenging Sentence Order Prediction (SOP) objective.
ALBERT uses the same SentencePiece tokenization as XLNet, and it is trained on similar corpora.
It is released in several configurations, showing benchmark performance comparable to BERT while using fewer parameters.

\begin{table}[htb]
\centering
\resizebox{\textwidth}{!}{%
\begin{tabular}{clccccccc} \toprule
\textbf{Model} & \textbf{Configuration} & \textbf{Params.} & \textbf{Layers} & \textbf{Heads} & \textbf{Dims.} & \textbf{Tokenization} & \textbf{Tasks} & \textbf{Corpus} \\ \midrule \midrule
\multirow{6}{*}{BERT} & B & 110M & 12 & 12 & 768 & WordPiece & MLM, NSP & 16\small{GB}\\
 & B-UNC & 110M & 12 & 12 & 768 & WordPiece, Unc. & MLM, NSP & 16\small{GB}\\
 & L & 340M & 24 & 16 & 1024 & WordPiece & MLM, NSP & 16\small{GB}\\
 & L-UNC & 340M & 24 & 16 & 1024 & WordPiece, Unc. & MLM, NSP & 16GB\\
 & L-WHL & 340M & 24 & 16 & 1024 & WordPiece & MLM, NSP & 16GB\\
 & L-UNC-WHL & 340M & 24 & 16 & 1024 & WordPiece, Unc. & MLM, NSP & 16GB\\ \midrule
\multirow{2}{*}{XLNet} & B & 110M & 12 & 12 & 768 & SentencePiece & PLM & 158GB\\
 & L & 340M & 24 & 16 & 1024 & SentencePiece & PLM & 158GB\\ \midrule
\multirow{2}{*}{RoBERTa} & B & 125M & 12 & 12 & 768 & Byte-level BPE & MLM & 160GB\\
 & L & 355M & 24 & 16 & 1024 & Byte-level BPE & MLM & 160GB\\ \midrule
\multirow{4}{*}{ALBERT} & B & 11M & 12 & 12 & 768 & SentencePiece & MLM, SOP & 160GB\\
 & L & 17M & 24 & 16 & 1024 & SentencePiece & MLM, SOP & 160GB\\
 & XL & 58M & 24 & 16 & 2048 & SentencePiece & MLM, SOP & 160GB\\
 & XXL & 223M & 12 & 64 & 4096 & SentencePiece & MLM, SOP & 160GB\\ \bottomrule
\end{tabular}%
}
\caption{Feature comparison for the NLMs used in this work. Configuration names are shortened for readability: B - Base; L - Large; XL - Extra Large; XXL - Extra Extra Large; UNC - Uncased; WHL - Whole-Word.}
\label{tab:nlm_details}
\end{table}

The full set of 14 model variants detailed on Table \ref{tab:nlm_details} are only used for layer-specific validation performance on WSD and USM tasks.
For task evaluation and analyses, we proceed with the single best performing model configuration from each model family, according to results from the validation experiments.

We use the Transformers package \citep{wolf-etal-2020-transformers} (v3.0.2) for experiments with BERT, XLNet and ALBERT, and the fairseq package \citep{ott-etal-2019-fairseq} (v0.9.0) for experiments with RoBERTa \footnote{Initial experiments with RoBERTa showed slightly better results using fairseq.}.

\subsection{Corpora for Training and Validation}
\label{exp:datasets}

We learn the initial set of sense representations described in Section \ref{met:embed} using sense-annotated corpora, namely SemCor \citep{Miller1994UsingAS} and the Unambiguous Word Annotations corpus \citep[UWA]{loureiro-camacho-collados-2020-dont}.

SemCor is a sense-annotated version of the Brown Corpus that still remains the largest corpus with manual sense-annotations despite its age.
It includes 226,695 annotations for 33,362 sensekeys (25,942 synsets), reaching a coverage of 16.1\% of WordNet's sense inventory. 
We use the version released in \citet{raganato-etal-2017-word}, which includes mappings updated to WordNet version 3.0.

UWA is our recently introduced corpus composed exclusively of annotations for unambiguous words from Wikipedia sentences.
Since WordNet is mostly composed of unambiguous words, UWA not only allows for representing the majority of WordNet senses (56.7\%, when combined with SemCor) from direct annotations, but also leads to improved sense representation for senses learned through propagation (as described in Section \ref{met:prop}), due to network effects.
UWA is released in several versions of different sizes, in this work we use the version with up to 10 examples per sense (denoted UWA10), which includes 867,252 annotations for 98,494 sensekeys (67,860 synsets).

In order to avoid interference with the standard test sets, we perform our layer analysis and probing using a custom validation set, based on the MASC\footnote{We use the version of the MASC corpus released in \citet{vial-etal-2018-ufsac}} corpus \citep{Ideetal:10}, following \citet{loureiro-etal-analysis-2021}.
Considering that our layer experiments are focused on intrinsic properties of NLMs, this custom version of the MASC corpus is restricted to only include annotations for senses that occur in SemCor.
Any sentence annotated with senses not occurring in SemCor is discarded, leaving a total of 14,645 annotations.
As such, our layer experiments use sense embeddings learned from SemCor and validated using this restricted version of MASC, without requiring strategies for inferring senses (e.g., ontological propagation), or fallbacks (e.g., Most Frequent Sense).


\section{Probing Analysis}
\label{probing:results}

In this section we present the outcome of the probing methodology described in Section \ref{met:lay} applied on the models detailed in Section \ref{exp:models}.
We report probing results in this section so they are presented and discussed before the evaluation and analysis sections, which report downstream task results using layer pooling informed by the probing analysis.

This section starts by covering our initial findings regarding layer performance variation patterns observed for all models ($\S$\ref{probing:variation}).
Second, we present validation results using our proposed sense profiles for both disambiguation and matching scenarios ($\S$\ref{probing:profiles}).
Finally, we present our rationale for choosing which sense profile should be used according to the type of task ($\S$\ref{probing:choosing}).

\subsection{Variation in Layer Performance Across NLMs}
\label{probing:variation}

As discussed in Section \ref{rel:lay}, it is well-understood that task performance varies considerably depending on which layers are used to retrieve embeddings from.
While some works have analysed task performance per layer specifically for the task of WSD \citep{coenen-etal-2019-visualizing,loureiro-etal-analysis-2021}, there is still lacking an in-depth cross-model comparison.

In Table \ref{tab:heatmap} we report WSD and USM performance for individual layers of each of the 14 models belonging to 4 different Transformer-based model families.
These results are obtained using the methodology described in Section \ref{met:prob}.
We observe that indeed, the final layer (-1) is never optimal for either WSD or USM performance.
More interestingly, we find that while the second-to-last layer (-2) always performs best for WSD in BERT models (with the exception of {\small BERT-L-UNC-WHL}), that pattern does not hold for the other models tested.
For some models, such as {\small XLNet-L} or {\small ALBERT-L}, we even find that the best performing layers are closer to the initialization layer (INIT) than to the final layer.
Another apparent pattern is that the best performing layers for USM are consistently lower than for WSD.
This can be explained by the fact that USM benefits from lexical information encoded in the lower layers, even though the initialization layer still performs worst, just as with WSD, demonstrating that it is not sufficient by itself.

These empirical results suggest that any layer pooling strategy based on a fixed set of layers, such as the often used sum of layers [-1,-4], cannot accurately capture the available sense information encoded in pre-trained NLMs.

\clearpage

\begin{center}
\scriptsize{Word Sense Disambiguation (WSD)}
\end{center}
\vspace{-0.5cm}

\begin{table}[ht]
\centering
\resizebox{0.77\textwidth}{!}{%
\begin{tabular}{cl>{\centering\arraybackslash}p{0.15em}>{\centering\arraybackslash}p{0.15em}>{\centering\arraybackslash}p{0.15em}>{\centering\arraybackslash}p{0.15em}>{\centering\arraybackslash}p{0.15em}>{\centering\arraybackslash}p{0.15em}>{\centering\arraybackslash}p{0.15em}>{\centering\arraybackslash}p{0.15em}>{\centering\arraybackslash}p{0.15em}>{\centering\arraybackslash}p{0.15em}>{\centering\arraybackslash}p{0.15em}>{\centering\arraybackslash}p{0.15em}>{\centering\arraybackslash}p{0.15em}>{\centering\arraybackslash}p{0.15em}>{\centering\arraybackslash}p{0.15em}>{\centering\arraybackslash}p{0.15em}>{\centering\arraybackslash}p{0.15em}>{\centering\arraybackslash}p{0.15em}>{\centering\arraybackslash}p{0.15em}>{\centering\arraybackslash}p{0.15em}>{\centering\arraybackslash}p{0.15em}>{\centering\arraybackslash}p{0.15em}>{\centering\arraybackslash}p{0.15em}>{\centering\arraybackslash}p{0.15em}>{\centering\arraybackslash}p{0.15em}} \toprule
\multicolumn{2}{c}{\textbf{Model}} & \rotatebox{90}{\textbf{\scriptsize{INIT}}} & \rotatebox{90}{\textbf{\scriptsize{-24}}} & \rotatebox{90}{\textbf{\scriptsize{-23}}} & \rotatebox{90}{\textbf{\scriptsize{-22}}} &\rotatebox{90}{\textbf{\scriptsize{-21}}} & \rotatebox{90}{\textbf{\scriptsize{-20}}} & \rotatebox{90}{\textbf{\scriptsize{-19}}} & \rotatebox{90}{\textbf{\scriptsize{-18}}} & \rotatebox{90}{\textbf{\scriptsize{-17}}} & \rotatebox{90}{\textbf{\scriptsize{-16}}} & \rotatebox{90}{\textbf{\scriptsize{-15}}} &\rotatebox{90}{\textbf{\scriptsize{-14}}} & \rotatebox{90}{\textbf{\scriptsize{-13}}} & \rotatebox{90}{\textbf{\scriptsize{-12}}} & \rotatebox{90}{\textbf{\scriptsize{-11}}} & \rotatebox{90}{\textbf{\scriptsize{-10}}} & \rotatebox{90}{\textbf{\scriptsize{-9}}} & \rotatebox{90}{\textbf{\scriptsize{-8}}} &\rotatebox{90}{\textbf{\scriptsize{-7}}} & \rotatebox{90}{\textbf{\scriptsize{-6}}} & \rotatebox{90}{\textbf{\scriptsize{-5}}} & \rotatebox{90}{\textbf{\scriptsize{-4}}} & \rotatebox{90}{\textbf{\scriptsize{-3}}} & \rotatebox{90}{\textbf{\scriptsize{-2}}} & \rotatebox{90}{\textbf{\scriptsize{-1}}} \\ \midrule \midrule
 & B & \cellcolor[HTML]{E67C73} & \multicolumn{12}{l}{\cellcolor[HTML]{D0D0D0}} & \cellcolor[HTML]{EDA49E} & \cellcolor[HTML]{F5CFCC} & \cellcolor[HTML]{FAE7E5} & \cellcolor[HTML]{FBEFEE} & \cellcolor[HTML]{FDF6F5} & \cellcolor[HTML]{FFFFFF} & \cellcolor[HTML]{E6F5EE} & \cellcolor[HTML]{D2EDE0} & \cellcolor[HTML]{B8E3CE} & \cellcolor[HTML]{80CCA7} & \cellcolor[HTML]{57BB8A}$\textcolor{white}{\star}$ & \cellcolor[HTML]{95D4B5} \\
 & B-UNC & \cellcolor[HTML]{E67C73} & \multicolumn{12}{l}{\cellcolor[HTML]{D0D0D0}} & \cellcolor[HTML]{F1B9B4} & \cellcolor[HTML]{F6D2CF} & \cellcolor[HTML]{FAE6E4} & \cellcolor[HTML]{FCEFEE} & \cellcolor[HTML]{FEFBFA} & \cellcolor[HTML]{FFFFFF} & \cellcolor[HTML]{ECF7F2} & \cellcolor[HTML]{D3EDE1} & \cellcolor[HTML]{BAE3CF} & \cellcolor[HTML]{7FCBA6} & \cellcolor[HTML]{57BB8A}$\textcolor{white}{\star}$ & \cellcolor[HTML]{84CEA9} \\
 & L & \cellcolor[HTML]{E67C73} & \cellcolor[HTML]{EDA19A} & \cellcolor[HTML]{F2BEBA} & \cellcolor[HTML]{F3C3BF} & \cellcolor[HTML]{F6D2CF} & \cellcolor[HTML]{F8DFDC} & \cellcolor[HTML]{FAE5E3} & \cellcolor[HTML]{FBEAE8} & \cellcolor[HTML]{FCF0EF} & \cellcolor[HTML]{FDF6F6} & \cellcolor[HTML]{FEFBFB} & \cellcolor[HTML]{FFFFFF} & \cellcolor[HTML]{FEFEFE} & \cellcolor[HTML]{FBFEFC} & \cellcolor[HTML]{DFF2E9} & \cellcolor[HTML]{D1EDDF} & \cellcolor[HTML]{FBFEFC} & \cellcolor[HTML]{C3E7D5} & \cellcolor[HTML]{BEE5D2} & \cellcolor[HTML]{C3E7D5} & \cellcolor[HTML]{ABDDC5} & \cellcolor[HTML]{8BD0AE} & \cellcolor[HTML]{61BF91} & \cellcolor[HTML]{57BB8A}$\textcolor{white}{\star}$ & \cellcolor[HTML]{EDF8F2} \\
 & L-UNC & \cellcolor[HTML]{E67C73} & \cellcolor[HTML]{EFAFAA} & \cellcolor[HTML]{F3C1BD} & \cellcolor[HTML]{F5CBC7} & \cellcolor[HTML]{F9E2E0} & \cellcolor[HTML]{FAE9E8} & \cellcolor[HTML]{FCF1F0} & \cellcolor[HTML]{FDF5F5} & \cellcolor[HTML]{FDF7F6} & \cellcolor[HTML]{FCF2F1} & \cellcolor[HTML]{FDF5F5} & \cellcolor[HTML]{FEFCFC} & \cellcolor[HTML]{FFFFFF} & \cellcolor[HTML]{E2F3EB} & \cellcolor[HTML]{C4E7D6} & \cellcolor[HTML]{B8E2CD} & \cellcolor[HTML]{A6DBC1} & \cellcolor[HTML]{8ED1B0} & \cellcolor[HTML]{88CFAC} & \cellcolor[HTML]{8ED1B0} & \cellcolor[HTML]{7CCAA4} & \cellcolor[HTML]{70C59B} & \cellcolor[HTML]{57BB8A}$\textcolor{white}{\star}$ & \cellcolor[HTML]{57BB8A}$\textcolor{white}{\star}$ & \cellcolor[HTML]{75C89F} \\
 & L-WHL & \cellcolor[HTML]{E67C73} & \cellcolor[HTML]{EA928B} & \cellcolor[HTML]{EFAFA9} & \cellcolor[HTML]{F2BEBA} & \cellcolor[HTML]{F3C1BC} & \cellcolor[HTML]{F8DBD8} & \cellcolor[HTML]{F9E1DF} & \cellcolor[HTML]{FBEAE9} & \cellcolor[HTML]{FCF2F1} & \cellcolor[HTML]{FDF5F4} & \cellcolor[HTML]{FDF6F6} & \cellcolor[HTML]{FEFEFE} & \cellcolor[HTML]{FFFFFF} & \cellcolor[HTML]{EFF9F4} & \cellcolor[HTML]{D4EEE1} & \cellcolor[HTML]{BEE5D2} & \cellcolor[HTML]{AEDFC7} & \cellcolor[HTML]{A9DCC3} & \cellcolor[HTML]{A3DABF} & \cellcolor[HTML]{A9DCC3} & \cellcolor[HTML]{A3DABF} & \cellcolor[HTML]{9ED8BC} & \cellcolor[HTML]{99D6B8} & \cellcolor[HTML]{57BB8A}$\textcolor{white}{\star}$ & \cellcolor[HTML]{B4E1CB} \\
\multirow{-6}{*}{BERT} & L-UNC-WHL & \cellcolor[HTML]{E67C73} & \cellcolor[HTML]{EB9891} & \cellcolor[HTML]{EFB0AA} & \cellcolor[HTML]{F2BEBA} & \cellcolor[HTML]{F5CBC8} & \cellcolor[HTML]{F7D7D5} & \cellcolor[HTML]{F8DEDC} & \cellcolor[HTML]{F9E4E2} & \cellcolor[HTML]{FAEAE8} & \cellcolor[HTML]{FDF5F4} & \cellcolor[HTML]{FEFDFD} & \cellcolor[HTML]{A0D9BD} & \cellcolor[HTML]{FFFFFF} & \cellcolor[HTML]{F3FBF7} & \cellcolor[HTML]{C4E7D6} & \cellcolor[HTML]{DCF1E6} & \cellcolor[HTML]{70C59B} & \cellcolor[HTML]{57BB8A}$\textcolor{white}{\star}$ & \cellcolor[HTML]{A0D9BD} & \cellcolor[HTML]{DCF1E6} & \cellcolor[HTML]{DCF1E6} & \cellcolor[HTML]{CFECDE} & \cellcolor[HTML]{CFECDE} & \cellcolor[HTML]{CFECDE} & \cellcolor[HTML]{FEFCFC} \\ \midrule
 & B & \cellcolor[HTML]{E67C73} & \multicolumn{12}{l}{\cellcolor[HTML]{D0D0D0}} & \cellcolor[HTML]{F4C5C1} & \cellcolor[HTML]{FBEDEB} & \cellcolor[HTML]{FDF6F6} & \cellcolor[HTML]{E1F3EA} & \cellcolor[HTML]{85CEAA} & \cellcolor[HTML]{76C8A0} & \cellcolor[HTML]{A4DAC0} & \cellcolor[HTML]{FFFFFF} & \cellcolor[HTML]{D2EDE0} & \cellcolor[HTML]{57BB8A}$\textcolor{white}{\star}$ & \cellcolor[HTML]{FCF1F0} & \cellcolor[HTML]{FEFDFD} \\
\multirow{-2}{*}{XLNet} & L & \cellcolor[HTML]{E67C73} & \cellcolor[HTML]{EDA29C} & \cellcolor[HTML]{F7D6D3} & \cellcolor[HTML]{F9E3E1} & \cellcolor[HTML]{FBEBEA} & \cellcolor[HTML]{FDF6F5} & \cellcolor[HTML]{FFFFFF} & \cellcolor[HTML]{93D4B4} & \cellcolor[HTML]{57BB8A}$\textcolor{white}{\star}$ & \cellcolor[HTML]{6BC498} & \cellcolor[HTML]{89CFAD} & \cellcolor[HTML]{A7DBC2} & \cellcolor[HTML]{C4E8D6} & \cellcolor[HTML]{B0E0C8} & \cellcolor[HTML]{ECF7F2} & \cellcolor[HTML]{FFFFFF} & \cellcolor[HTML]{FEFDFD} & \cellcolor[HTML]{FEFDFD} & \cellcolor[HTML]{FEFDFC} & \cellcolor[HTML]{FEFEFE} & \cellcolor[HTML]{93D4B4} & \cellcolor[HTML]{89CFAD} & \cellcolor[HTML]{93D4B4} & \cellcolor[HTML]{89CFAD} & \cellcolor[HTML]{FBEAE9} \\ \midrule
 & B & \cellcolor[HTML]{E67C73} & \multicolumn{12}{l}{\cellcolor[HTML]{D0D0D0}} & \cellcolor[HTML]{F1B8B3} & \cellcolor[HTML]{F7D9D6} & \cellcolor[HTML]{FBEEED} & \cellcolor[HTML]{FEFCFB} & \cellcolor[HTML]{B5E1CB} & \cellcolor[HTML]{DAF0E5} & \cellcolor[HTML]{DAF0E5} & \cellcolor[HTML]{FFFFFF} & \cellcolor[HTML]{FFFFFF} & \cellcolor[HTML]{57BB8A}$\textcolor{white}{\star}$ & \cellcolor[HTML]{7DCBA5} & \cellcolor[HTML]{FDF9F8} \\
\multirow{-2}{*}{RoBERTa} & L & \cellcolor[HTML]{E67C73} & \cellcolor[HTML]{EA948D} & \cellcolor[HTML]{F2BAB6} & \cellcolor[HTML]{F5CECB} & \cellcolor[HTML]{F7D6D3} & \cellcolor[HTML]{F9E1DF} & \cellcolor[HTML]{FBEDEC} & \cellcolor[HTML]{FCF1F0} & \cellcolor[HTML]{FEFAFA} & \cellcolor[HTML]{FEFBFA} & \cellcolor[HTML]{FFFFFF} & \cellcolor[HTML]{E6F5ED} & \cellcolor[HTML]{D9F0E4} & \cellcolor[HTML]{A5DBC1} & \cellcolor[HTML]{57BB8A}$\textcolor{white}{\star}$ & \cellcolor[HTML]{57BB8A}$\textcolor{white}{\star}$ & \cellcolor[HTML]{7ECBA6} & \cellcolor[HTML]{B2E0C9} & \cellcolor[HTML]{8BD0AF} & \cellcolor[HTML]{BFE5D3} & \cellcolor[HTML]{CCEBDB} & \cellcolor[HTML]{B2E0C9} & \cellcolor[HTML]{FEFBFA} & \cellcolor[HTML]{98D6B7} & \cellcolor[HTML]{FCF0EF} \\ \midrule
 & B & \cellcolor[HTML]{E67C73} & \multicolumn{12}{l}{\cellcolor[HTML]{D0D0D0}} & \cellcolor[HTML]{F3C2BE} & \cellcolor[HTML]{F8DEDC} & \cellcolor[HTML]{FCF1F1} & \cellcolor[HTML]{FDF8F8} & \cellcolor[HTML]{FFFFFF} & \cellcolor[HTML]{57BB8A}$\textcolor{white}{\star}$ & \cellcolor[HTML]{74C79E} & \cellcolor[HTML]{57BB8A}$\textcolor{white}{\star}$ & \cellcolor[HTML]{C8E9D9} & \cellcolor[HTML]{E3F4EC} & \cellcolor[HTML]{FFFFFF} & \cellcolor[HTML]{C8E9D9} \\
 & L & \cellcolor[HTML]{E67C73} & \cellcolor[HTML]{F1B8B3} & \cellcolor[HTML]{F6D3D0} & \cellcolor[HTML]{F9DFDD} & \cellcolor[HTML]{FAE9E8} & \cellcolor[HTML]{FCF0EF} & \cellcolor[HTML]{FDF5F4} & \cellcolor[HTML]{FEFBFA} & \cellcolor[HTML]{ABDDC5} & \cellcolor[HTML]{57BB8A}$\textcolor{white}{\star}$ & \cellcolor[HTML]{57BB8A}$\textcolor{white}{\star}$ & \cellcolor[HTML]{C8E9D9} & \cellcolor[HTML]{ABDDC5} & \cellcolor[HTML]{90D2B2} & \cellcolor[HTML]{90D2B2} & \cellcolor[HTML]{ABDDC5} & \cellcolor[HTML]{FFFFFF} & \cellcolor[HTML]{FEFEFE} & \cellcolor[HTML]{FFFFFF} & \cellcolor[HTML]{FFFFFF} & \cellcolor[HTML]{C8E9D9} & \cellcolor[HTML]{E4F4EC} & \cellcolor[HTML]{FEFDFD} & \cellcolor[HTML]{FDF9F9} & \cellcolor[HTML]{FDF8F8} \\
 & XL & \cellcolor[HTML]{E67C73} & \cellcolor[HTML]{F7D5D3} & \cellcolor[HTML]{FAE7E6} & \cellcolor[HTML]{FCF3F2} & \cellcolor[HTML]{FFFFFF} & \cellcolor[HTML]{F7FCFA} & \cellcolor[HTML]{CDEBDC} & \cellcolor[HTML]{9BD7B9} & \cellcolor[HTML]{81CCA8} & \cellcolor[HTML]{79C9A2} & \cellcolor[HTML]{68C296} & \cellcolor[HTML]{60BF90} & \cellcolor[HTML]{57BB8A}$\textcolor{white}{\star}$ & \cellcolor[HTML]{68C296} & \cellcolor[HTML]{92D3B3} & \cellcolor[HTML]{ABDDC5} & \cellcolor[HTML]{EFF9F4} & \cellcolor[HTML]{FFFFFF} & \cellcolor[HTML]{FEFDFD} & \cellcolor[HTML]{FEFCFC} & \cellcolor[HTML]{FEFAFA} & \cellcolor[HTML]{FDF8F7} & \cellcolor[HTML]{FDF9F8} & \cellcolor[HTML]{FDF7F6} & \cellcolor[HTML]{F6D3D0} \\
\multirow{-4}{*}{ALBERT} & XXL & \cellcolor[HTML]{E67C73} & \multicolumn{12}{l}{\cellcolor[HTML]{D0D0D0}} & \cellcolor[HTML]{F6D0CD} & \cellcolor[HTML]{FAE5E3} & \cellcolor[HTML]{FDF7F6} & \cellcolor[HTML]{FFFFFF} & \cellcolor[HTML]{B5E1CC} & \cellcolor[HTML]{90D2B2} & \cellcolor[HTML]{57BB8A}$\textcolor{white}{\star}$ & \cellcolor[HTML]{90D2B2} & \cellcolor[HTML]{90D2B2} & \cellcolor[HTML]{90D2B2} & \cellcolor[HTML]{FEFCFC} & \cellcolor[HTML]{F8DCDA} \\ \bottomrule
\end{tabular}%
}
\label{tab:wsd_heatmap}
\end{table}

\vspace{-0.5cm}

\begin{center}
\scriptsize{Uninformed Sense Matching (USM)}
\end{center}
\vspace{-0.5cm}

\begin{table}[ht]
\centering
\resizebox{0.77\textwidth}{!}{%
\begin{tabular}{cl>{\centering\arraybackslash}p{0.15em}>{\centering\arraybackslash}p{0.15em}>{\centering\arraybackslash}p{0.15em}>{\centering\arraybackslash}p{0.15em}>{\centering\arraybackslash}p{0.15em}>{\centering\arraybackslash}p{0.15em}>{\centering\arraybackslash}p{0.15em}>{\centering\arraybackslash}p{0.15em}>{\centering\arraybackslash}p{0.15em}>{\centering\arraybackslash}p{0.15em}>{\centering\arraybackslash}p{0.15em}>{\centering\arraybackslash}p{0.15em}>{\centering\arraybackslash}p{0.15em}>{\centering\arraybackslash}p{0.15em}>{\centering\arraybackslash}p{0.15em}>{\centering\arraybackslash}p{0.15em}>{\centering\arraybackslash}p{0.15em}>{\centering\arraybackslash}p{0.15em}>{\centering\arraybackslash}p{0.15em}>{\centering\arraybackslash}p{0.15em}>{\centering\arraybackslash}p{0.15em}>{\centering\arraybackslash}p{0.15em}>{\centering\arraybackslash}p{0.15em}>{\centering\arraybackslash}p{0.15em}>{\centering\arraybackslash}p{0.15em}} \toprule
\multicolumn{2}{c}{\textbf{Model}} & \rotatebox{90}{\textbf{\scriptsize{INIT}}} & \rotatebox{90}{\textbf{\scriptsize{-24}}} & \rotatebox{90}{\textbf{\scriptsize{-23}}} & \rotatebox{90}{\textbf{\scriptsize{-22}}} &\rotatebox{90}{\textbf{\scriptsize{-21}}} & \rotatebox{90}{\textbf{\scriptsize{-20}}} & \rotatebox{90}{\textbf{\scriptsize{-19}}} & \rotatebox{90}{\textbf{\scriptsize{-18}}} & \rotatebox{90}{\textbf{\scriptsize{-17}}} & \rotatebox{90}{\textbf{\scriptsize{-16}}} & \rotatebox{90}{\textbf{\scriptsize{-15}}} &\rotatebox{90}{\textbf{\scriptsize{-14}}} & \rotatebox{90}{\textbf{\scriptsize{-13}}} & \rotatebox{90}{\textbf{\scriptsize{-12}}} & \rotatebox{90}{\textbf{\scriptsize{-11}}} & \rotatebox{90}{\textbf{\scriptsize{-10}}} & \rotatebox{90}{\textbf{\scriptsize{-9}}} & \rotatebox{90}{\textbf{\scriptsize{-8}}} &\rotatebox{90}{\textbf{\scriptsize{-7}}} & \rotatebox{90}{\textbf{\scriptsize{-6}}} & \rotatebox{90}{\textbf{\scriptsize{-5}}} & \rotatebox{90}{\textbf{\scriptsize{-4}}} & \rotatebox{90}{\textbf{\scriptsize{-3}}} & \rotatebox{90}{\textbf{\scriptsize{-2}}} & \rotatebox{90}{\textbf{\scriptsize{-1}}} \\ \midrule \midrule
 & B & \cellcolor[HTML]{E67C73} & \multicolumn{12}{l}{\cellcolor[HTML]{D0D0D0}} & \cellcolor[HTML]{EEA9A3} & \cellcolor[HTML]{FAE5E3} & \cellcolor[HTML]{FFFFFF} & \cellcolor[HTML]{9BD7B9} & \cellcolor[HTML]{57BB8A}$\textcolor{white}{\star}$ & \cellcolor[HTML]{57BB8A}$\textcolor{white}{\star}$ & \cellcolor[HTML]{57BB8A}$\textcolor{white}{\star}$ & \cellcolor[HTML]{92D3B3} & \cellcolor[HTML]{F7FCFA} & \cellcolor[HTML]{FDF7F7} & \cellcolor[HTML]{FCEFEE} & \cellcolor[HTML]{FDF9F8} \\
 & B-UNC & \cellcolor[HTML]{E67C73} & \multicolumn{12}{l}{\cellcolor[HTML]{D0D0D0}} & \cellcolor[HTML]{F5CECB} & \cellcolor[HTML]{FDF6F5} & \cellcolor[HTML]{B0DFC8} & \cellcolor[HTML]{78C9A1} & \cellcolor[HTML]{57BB8A}$\textcolor{white}{\star}$ & \cellcolor[HTML]{66C194} & \cellcolor[HTML]{7DCBA5} & \cellcolor[HTML]{B5E1CB} & \cellcolor[HTML]{FFFFFF} & \cellcolor[HTML]{FBECEB} & \cellcolor[HTML]{FBECEB} & \cellcolor[HTML]{F9E3E1} \\
 & L & \cellcolor[HTML]{E67C73} & \cellcolor[HTML]{EEAAA4} & \cellcolor[HTML]{F5CDCA} & \cellcolor[HTML]{F6D4D1} & \cellcolor[HTML]{FCF0EF} & \cellcolor[HTML]{FFFFFF} & \cellcolor[HTML]{D9F0E5} & \cellcolor[HTML]{B7E2CD} & \cellcolor[HTML]{9FD9BD} & \cellcolor[HTML]{74C79F} & \cellcolor[HTML]{61BF91} & \cellcolor[HTML]{57BB8A}$\textcolor{white}{\star}$ & \cellcolor[HTML]{61BF91} & \cellcolor[HTML]{66C195} & \cellcolor[HTML]{61BF91} & \cellcolor[HTML]{74C79F} & \cellcolor[HTML]{CBEADB} & \cellcolor[HTML]{D9F0E5} & \cellcolor[HTML]{FEFCFB} & \cellcolor[HTML]{FCF2F1} & \cellcolor[HTML]{FCF0EF} & \cellcolor[HTML]{FCF1F0} & \cellcolor[HTML]{FBEDEB} & \cellcolor[HTML]{F9E3E2} & \cellcolor[HTML]{FCF0EF} \\
 & B-UNC & \cellcolor[HTML]{E67C73} & \cellcolor[HTML]{F1B8B3} & \cellcolor[HTML]{F6D1CE} & \cellcolor[HTML]{F9E2E0} & \cellcolor[HTML]{FFFFFF} & \cellcolor[HTML]{C1E6D4} & \cellcolor[HTML]{96D5B6} & \cellcolor[HTML]{7ECBA5} & \cellcolor[HTML]{74C79F} & \cellcolor[HTML]{9FD9BD} & \cellcolor[HTML]{7ECBA5} & \cellcolor[HTML]{57BB8A}$\textcolor{white}{\star}$ & \cellcolor[HTML]{6FC59B} & \cellcolor[HTML]{74C79F} & \cellcolor[HTML]{83CDA9} & \cellcolor[HTML]{9BD7B9} & \cellcolor[HTML]{BCE4D1} & \cellcolor[HTML]{FEFEFE} & \cellcolor[HTML]{FDF8F7} & \cellcolor[HTML]{FBEDEC} & \cellcolor[HTML]{FAE8E6} & \cellcolor[HTML]{F9E0DE} & \cellcolor[HTML]{F8DEDC} & \cellcolor[HTML]{F7D6D3} & \cellcolor[HTML]{F8DCDA} \\
 & L-WHL & \cellcolor[HTML]{E67C73} & \cellcolor[HTML]{EB9890} & \cellcolor[HTML]{F1B8B3} & \cellcolor[HTML]{F5CDC9} & \cellcolor[HTML]{F6D3D0} & \cellcolor[HTML]{FBEFEE} & \cellcolor[HTML]{FEF9F9} & \cellcolor[HTML]{EDF8F2} & \cellcolor[HTML]{B5E1CB} & \cellcolor[HTML]{A2DABF} & \cellcolor[HTML]{90D2B2} & \cellcolor[HTML]{57BB8A}$\textcolor{white}{\star}$ & \cellcolor[HTML]{7DCBA4} & \cellcolor[HTML]{86CEAB} & \cellcolor[HTML]{90D2B2} & \cellcolor[HTML]{86CEAB} & \cellcolor[HTML]{99D6B8} & \cellcolor[HTML]{A2DABF} & \cellcolor[HTML]{EDF8F2} & \cellcolor[HTML]{FEFDFD} & \cellcolor[HTML]{FFFFFF} & \cellcolor[HTML]{FDF9F8} & \cellcolor[HTML]{FCF0EF} & \cellcolor[HTML]{FEF9F9} & \cellcolor[HTML]{FCF2F1} \\
\multirow{-6}{*}{BERT} & L-UNC-WHL & \cellcolor[HTML]{E67C73} & \cellcolor[HTML]{EC9F99} & \cellcolor[HTML]{F2BCB7} & \cellcolor[HTML]{F6D4D1} & \cellcolor[HTML]{FAE7E6} & \cellcolor[HTML]{FDF6F6} & \cellcolor[HTML]{FFFFFF} & \cellcolor[HTML]{C6E8D8} & \cellcolor[HTML]{A1D9BE} & \cellcolor[HTML]{70C59C} & \cellcolor[HTML]{64C093} & \cellcolor[HTML]{57BB8A}$\textcolor{white}{\star}$ & \cellcolor[HTML]{7CCAA4} & \cellcolor[HTML]{91D3B2} & \cellcolor[HTML]{A1D9BE} & \cellcolor[HTML]{9DD8BB} & \cellcolor[HTML]{99D6B8} & \cellcolor[HTML]{A1D9BE} & \cellcolor[HTML]{CAEADA} & \cellcolor[HTML]{FEFDFD} & \cellcolor[HTML]{FDF9F8} & \cellcolor[HTML]{FCF3F3} & \cellcolor[HTML]{FAE8E7} & \cellcolor[HTML]{FCEFEE} & \cellcolor[HTML]{FCEFEE} \\ \midrule
 & B & \cellcolor[HTML]{E98E86} & \multicolumn{12}{l}{\cellcolor[HTML]{D0D0D0}} & \cellcolor[HTML]{D5EEE2} & \cellcolor[HTML]{76C8A0} & \cellcolor[HTML]{57BB8A}$\textcolor{white}{\star}$ & \cellcolor[HTML]{57BB8A}$\textcolor{white}{\star}$ & \cellcolor[HTML]{74C79E} & \cellcolor[HTML]{B8E3CE} & \cellcolor[HTML]{FFFFFF} & \cellcolor[HTML]{F4C8C4} & \cellcolor[HTML]{F0B2AD} & \cellcolor[HTML]{F1B6B1} & \cellcolor[HTML]{E67C73} & \cellcolor[HTML]{EFADA7} \\
\multirow{-2}{*}{XLNet} & L & \cellcolor[HTML]{F2BEB9} & \cellcolor[HTML]{FCF2F1} & \cellcolor[HTML]{9FD8BC} & \cellcolor[HTML]{83CDA8} & \cellcolor[HTML]{74C79E} & \cellcolor[HTML]{65C194} & \cellcolor[HTML]{59BC8B} & \cellcolor[HTML]{57BB8A}$\textcolor{white}{\star}$ & \cellcolor[HTML]{66C195} & \cellcolor[HTML]{74C79E} & \cellcolor[HTML]{85CEAA} & \cellcolor[HTML]{97D5B6} & \cellcolor[HTML]{B3E0CA} & \cellcolor[HTML]{C6E8D7} & \cellcolor[HTML]{FFFFFF} & \cellcolor[HTML]{FDF6F5} & \cellcolor[HTML]{FBEEED} & \cellcolor[HTML]{F9E2E0} & \cellcolor[HTML]{F6D4D1} & \cellcolor[HTML]{F4C5C1} & \cellcolor[HTML]{F2BEB9} & \cellcolor[HTML]{F3C2BE} & \cellcolor[HTML]{F3C5C1} & \cellcolor[HTML]{F9E1DE} & \cellcolor[HTML]{E67C73} \\ \midrule
 & B & \cellcolor[HTML]{E67C73} & \multicolumn{12}{l}{\cellcolor[HTML]{D0D0D0}} & \cellcolor[HTML]{F6D0CC} & \cellcolor[HTML]{FDF5F4} & \cellcolor[HTML]{C0E6D4} & \cellcolor[HTML]{57BB8A}$\textcolor{white}{\star}$ & \cellcolor[HTML]{66C194} & \cellcolor[HTML]{97D5B6} & \cellcolor[HTML]{A4DBC0} & \cellcolor[HTML]{DDF1E7} & \cellcolor[HTML]{FFFFFF} & \cellcolor[HTML]{FDF6F6} & \cellcolor[HTML]{FBEEED} & \cellcolor[HTML]{FCF2F1} \\
\multirow{-2}{*}{RoBERTa} & L & \cellcolor[HTML]{E67C73} & \cellcolor[HTML]{EB9A93} & \cellcolor[HTML]{F6D0CD} & \cellcolor[HTML]{FAE7E6} & \cellcolor[HTML]{FCF4F3} & \cellcolor[HTML]{FEFDFC} & \cellcolor[HTML]{D5EEE2} & \cellcolor[HTML]{96D5B6} & \cellcolor[HTML]{57BB8A}$\textcolor{white}{\star}$ & \cellcolor[HTML]{7BCAA3} & \cellcolor[HTML]{9ED8BB} & \cellcolor[HTML]{8FD2B1} & \cellcolor[HTML]{B2E0CA} & \cellcolor[HTML]{ACDEC5} & \cellcolor[HTML]{B2E0CA} & \cellcolor[HTML]{F8FDFB} & \cellcolor[HTML]{FEF9F9} & \cellcolor[HTML]{FEF9F9} & \cellcolor[HTML]{FEFCFC} & \cellcolor[HTML]{FEFBFA} & \cellcolor[HTML]{DCF1E7} & \cellcolor[HTML]{9ED8BB} & \cellcolor[HTML]{FEFBFB} & \cellcolor[HTML]{FFFFFF} & \cellcolor[HTML]{FBEDEB} \\ \midrule
 & B & \cellcolor[HTML]{E67C73} & \multicolumn{12}{l}{\cellcolor[HTML]{D0D0D0}} & \cellcolor[HTML]{F8DBD8} & \cellcolor[HTML]{FFFFFF} & \cellcolor[HTML]{81CCA8} & \cellcolor[HTML]{65C194} & \cellcolor[HTML]{57BB8A}$\textcolor{white}{\star}$ & \cellcolor[HTML]{65C194} & \cellcolor[HTML]{96D5B6} & \cellcolor[HTML]{B9E3CF} & \cellcolor[HTML]{FFFFFF} & \cellcolor[HTML]{FDF8F8} & \cellcolor[HTML]{FDF9F9} & \cellcolor[HTML]{FDF8F8} \\
 & L & \cellcolor[HTML]{E67C73} & \cellcolor[HTML]{F6D0CD} & \cellcolor[HTML]{FCF1F0} & \cellcolor[HTML]{FFFFFF} & \cellcolor[HTML]{C2E7D5} & \cellcolor[HTML]{95D4B5} & \cellcolor[HTML]{8AD0AE} & \cellcolor[HTML]{73C79E} & \cellcolor[HTML]{5DBE8E} & \cellcolor[HTML]{57BB8A}$\textcolor{white}{\star}$ & \cellcolor[HTML]{63C092} & \cellcolor[HTML]{95D4B5} & \cellcolor[HTML]{A6DBC1} & \cellcolor[HTML]{B7E2CD} & \cellcolor[HTML]{C2E7D5} & \cellcolor[HTML]{EFF9F4} & \cellcolor[HTML]{FEFDFD} & \cellcolor[HTML]{FDF8F8} & \cellcolor[HTML]{FDF4F4} & \cellcolor[HTML]{FCF2F1} & \cellcolor[HTML]{FBEEED} & \cellcolor[HTML]{FBECEB} & \cellcolor[HTML]{FBEBEA} & \cellcolor[HTML]{FCF2F1} & \cellcolor[HTML]{FEFBFB} \\
 & XL & \cellcolor[HTML]{E67C73} & \cellcolor[HTML]{FDF6F6} & \cellcolor[HTML]{AFDFC8} & \cellcolor[HTML]{83CDA9} & \cellcolor[HTML]{5FBE90} & \cellcolor[HTML]{5BBD8D} & \cellcolor[HTML]{5BBD8D} & \cellcolor[HTML]{57BB8A}$\textcolor{white}{\star}$ & \cellcolor[HTML]{62C092} & \cellcolor[HTML]{6AC397} & \cellcolor[HTML]{78C9A1} & \cellcolor[HTML]{96D5B6} & \cellcolor[HTML]{B3E0CA} & \cellcolor[HTML]{D4EEE1} & \cellcolor[HTML]{FFFFFF} & \cellcolor[HTML]{FDF5F4} & \cellcolor[HTML]{FBEAE9} & \cellcolor[HTML]{F9DFDD} & \cellcolor[HTML]{F7DAD7} & \cellcolor[HTML]{F7D6D3} & \cellcolor[HTML]{F6D2CF} & \cellcolor[HTML]{F5CECB} & \cellcolor[HTML]{F6D0CD} & \cellcolor[HTML]{F5CDC9} & \cellcolor[HTML]{EB9992} \\
\multirow{-4}{*}{ALBERT} & XXL & \cellcolor[HTML]{E67C73} & \multicolumn{12}{l}{\cellcolor[HTML]{D0D0D0}} & \cellcolor[HTML]{FAE6E5} & \cellcolor[HTML]{FEFCFC} & \cellcolor[HTML]{57BB8A}$\textcolor{white}{\star}$ & \cellcolor[HTML]{57BB8A}$\textcolor{white}{\star}$ & \cellcolor[HTML]{A2DABE} & \cellcolor[HTML]{EDF8F3} & \cellcolor[HTML]{FFFFFF} & \cellcolor[HTML]{FEFDFD} & \cellcolor[HTML]{FEFDFD} & \cellcolor[HTML]{EDF8F3} & \cellcolor[HTML]{EDF8F3} & \cellcolor[HTML]{F7D9D6} \\ \bottomrule
\end{tabular}%
}
\caption{Performance variation on the development set across all models and configurations considered in this work. Green represents best performing layers (best is marked with $\star$), red represents worst performing layers, and grey stands for layers missing in shallower variants.}
\label{tab:heatmap}
\end{table}

\clearpage

\subsection{Sense Profiles for Disambiguation and Matching}
\label{probing:profiles}

In Section \ref{met:prof} we described our method for uncovering a model-specific set of layer weights which informs a weighted layer pooling that results in improved sense representations.
We have applied this method to all our models, for both disambiguation and matching scenarios, in order to verify whether our proposed method reliably improves performance on WSD and USM tasks when compared to conventional pooling approaches.
Additionally, we also compare against different values for the temperature $t$ parameter.
In order to understand whether our recommended $t$ values actually result in improved test-time performance for these tasks, we run a limited evaluation on the ALL test set of \citet{raganato-etal-2017-word} where we compare NLMs using only our method as described until Section \ref{met:prop} (without using glosses) and trained solely with SemCor annotations.
Later, in Sections \ref{eval:wsd} and \ref{eval:usm}, we report WSD and USM results using our final solution in comparison with the current state-of-the-art.

The conventional layer choices we considered are the following: last/final ($L_{-1}$); second-to-last ($L_{-2}$); sum of last 4 ($L_{-1} + L_{-2} + L_{-3} + L_{-4}$); integer weighted sum of last 4 ($L_{-1} + 2 * L_{-2} + 3 * L_{-3} + 4 * L_{-4}$); fractional weighted sum of last 4 ($ \frac{1}{4} * L_{-1} + \frac{1}{3} * L_{-2} + \frac{1}{2} * L_{-3} + L_{-4}$).
We tested temperature values $t \in \{0.002, 0.005, 0.01, 0.1, 1.0\}$\footnote{$t=0.001$ results in large exponents that cause overflow errors.}.
Below we discuss our findings regarding the impact of sense profiles specific to each task.

The WSD validation results on Table \ref{tab:wsd_dev} reveal that the single best layer (which varies depending on model, see Table \ref{tab:heatmap}) consistently outperforms the sum of last 4 layers.
We also find that WSD sense profiles with $t=0.002$ and $t=0.005$ perform comparably to the single best layer, with $t=0.002$ being slightly closer on average.
Given the close performance, we opt for recommending $t=0.005$ as higher values are less likely to overfit on the validation set (bias-variance tradeoff).
In the limited evaluation results on Table \ref{tab:wsd_test} we compare the performance of conventional layer choices against WSD sense profiles with the recommended temperature value.
We observe that for 11 out of 14 models, WSD sense profiles with the recommended temperature reliably outperform any of the conventional choices, of which none stands out as a reliable cross-model choice.
Moreover, on Table \ref{tab:wsd_test} we also see that WSD sense profiles with the recommended temperature generally match or outperform both the single layers which performed best on the validation set, and the WSD sense profiles using the temperature value that showed best performance on the validation set.

Our findings regarding performance of USM sense profiles largely follow the previously mentioned findings for WSD sense profiles.
In the case of USM, validation results on Table \ref{tab:usm_dev} more clearly show that $t=0.100$ performs better, although $t=1.000$ also performs well.
As for the limited evaluation results, Table \ref{tab:usm_test} shows that conventional layer choices significantly underperform any of the alternatives introduced in this work, with the USM sense profile with recommended temperature ($t=0.100$) showing overall best performance.

\subsection{Choosing Sense Profiles for Different Tasks}
\label{probing:choosing}

Having established that our proposed sense profiles improve WSD and USM performance over conventional layer choices, the question remains of whether to choose WSD or USM sense profiles to represent sense embeddings.
In this work we propose choosing sense profiles based on the probing task that shares most similar constraints to the downstream task of interest.
More specifically, tasks requiring comparison of different senses for the same word fit the disambiguation profile, such as classical WSD \citep{Navigli2009WordSD} or WiC \citep{pilehvar-camacho-collados-2019-wic}, and benefit less from information in lower layers.
On the other hand, tasks without lexical constraints, not only USM but also synset similarity \citep{colla-etal-2020-lesslex} or semantic change \citep{hamilton-etal-2016-diachronic}, are better suited to the matching profile, which uses information from more layers.
In Section \ref{eval:intro} we evaluate sense embeddings learned using sense profiles according to each task's constraints, and in Section \ref{analysis:profiles} we analyse the performance gap when using alternate sense profiles.

\clearpage

\begin{table}
\centering
\resizebox{0.95\textwidth}{!}{%
\begin{tabular}{clccccccc} \toprule
\multicolumn{2}{c}{\multirow{2}{*}{\textbf{Model}}} & \textbf{Sum} & \textbf{Layer} & \multicolumn{5}{c}{\textbf{Weighted Sum (WS)}} \\  \cline{5-9}
\multicolumn{2}{c}{} & \textbf{\small{LST4}} & \textbf{\small{Best}} & \textbf{\small{t=0.002}} & \textbf{\small{t=0.005}} & \textbf{\small{t=0.010}} & \textbf{\small{t=0.100}} & \textbf{\small{t=1.000}} \\ \midrule \midrule
\multirow{6}{*}{BERT} & {B} & {\large 71.6} & {\large \textbf{72.5}} \small{(-2)} & {\large 72.4} & {\large 72.2} & {\large 71.8} & {\large 69.6} & {\large 68.6} \\
 & {B-UNC} & {\large 71.9} & {\large \textbf{73.0}} \small{(-2)} & {\large 72.9} & {\large 72.8} & {\large 72.3} & {\large 70.3} & {\large 69.1} \\
 & {L} & {\large 73.8} & {\large \textbf{74.7}} \small{(-2)} & {\large \textbf{74.7}} & {\large 74.5} & {\large 74.3} & {\large 72.2} & {\large 70.9} \\
 & {L-UNC} & {\large 72.7} & {\large \textbf{72.9}} \small{(-2)} & {\large \textbf{72.9}} & {\large 72.7} & {\large 72.7} & {\large 71.5} & {\large 70.8} \\
 & {L-WHL} & {\large 72.0} & {\large 73.0} \small{(-2)} & {\large \textbf{73.1}} & {\large 72.6} & {\large 71.8} & {\large 70.4} & {\large 69.1} \\
 & {L-UNC-WHL} & {\large 71.5} & {\large \textbf{72.7}} \small{(-8)} & {\large 72.6} & {\large 72.1} & {\large 72.0} & {\large 71.5} & {\large 70.7} \\ \midrule
\multirow{2}{*}{XLNet} & {B} & {\large 66.6} & {\large 70.9} \small{(-3)} & {\large 71.0} & {\large \textbf{71.2}} & {\large \textbf{71.2}} & {\large 70.3} & {\large 69.9} \\
 & {L} & {\large 66.5} & {\large 72.7} \small{(-17)} & {\large 73.0} & {\large \textbf{73.4}} & {\large 73.3} & {\large 72.4} & {\large 71.8} \\ \midrule
\multirow{2}{*}{RoBERTa} & {B} & {\large 72.5} & {\large \textbf{72.9}} \small{(-3)} & {\large 72.8} & {\large 72.6} & {\large 72.2} & {\large 71.6} & {\large 71.2} \\
 & {L} & {\large 74.1} & {\large \textbf{74.9}}  \small{(-10)} & {\large \textbf{74.9}} & {\large 74.7} & {\large 74.4} & {\large 73.6} & {\large 73.1} \\ \midrule
\multirow{4}{*}{ALBERT} & {B} & {\large 68.3} & {\large \textbf{68.9}} \small{(-5)} & {\large 68.3} & {\large 68.2} & {\large 68.1} & {\large 67.6} & {\large 67.3} \\
 & {L} & {\large 69.4} & {\large \textbf{70.5}} \small{(-15)} & {\large 70.2} & {\large 70.0} & {\large 69.9} & {\large 69.3} & {\large 69.3} \\
 & {XL} & {\large 68.2} & {\large \textbf{71.4}} \small{(-13)} & {\large 71.1} & {\large 71.1} & {\large 71.0} & {\large 70.5} & {\large 70.4} \\
 & {XXL} & {\large 72.4} & {\large \textbf{73.8}} \small{(-6)} & {\large 73.5} & {\large 73.4} & {\large 73.3} & {\large 72.8} & {\large 72.4} \\ \bottomrule
\end{tabular}%
}
\caption{WSD validation results (F1). Reports best single layer and weighted sums using specific sense profiles, with different $t$ values, for each model configuration. Sense representations for this experiment were learned from SemCor (no propagation required).}
\label{tab:wsd_dev}
\end{table}

\begin{table}
\centering
\resizebox{0.95\textwidth}{!}{%
\begin{tabular}{clccccclccc}  \toprule
\multicolumn{2}{c}{\multirow{3}{*}{\textbf{Model}}} & \multicolumn{5}{c}{\textbf{Standard}} &  & \multicolumn{3}{c}{\textbf{Proposed}} \\ \cline{3-7} \cline{9-11} 
\multicolumn{2}{c}{} & \textbf{Layer} & \textbf{Layer} & \textbf{Sum} & \textbf{WS (I)} & \textbf{WS (F)} &  & \textbf{Layer} & \textbf{WS} & \textbf{WS} \\
\multicolumn{2}{c}{} & \textbf{\small{-1}} & \textbf{\small{-2}} & \textbf{\small{LST4}} & \textbf{\small{LST4}} & \textbf{\small{LST4}} &  & \textbf{\small{Best Dev}} & \textbf{\small{Rec. t}} & \textbf{\small{Best t}} \\ \hline \hline
\multirow{6}{*}{BERT} & B & {\large 72.1} & {\large \textbf{72.9}} & {\large 72.6} & {\large 72.5} & {\large 72.5} &  & {\large \textbf{72.9}} \small{(-2)} & {\large 72.8} & {\large \textbf{72.9}} \small{(.002)} \\
 & B-UNC & {\large \textbf{73.5}} & {\large \textbf{73.5}} & {\large 73.0} & {\large 73.3} & {\large 73.3} &  & {\large \textbf{73.5}} \small{(-2)} & {\large 73.4} & {\large \textbf{73.5}} \small{(.002)} \\
 & L & {\large 73.3} & {\large 73.9} & {\large 74.0} & {\large 74.0} & {\large 74.0} &  & {\large 73.9} \small{(-2)} & {\large \textbf{74.2}} & {\large 74.0} \small{(.002)} \\
 & L-UNC & {\large 73.4} & {\large 73.6} & {\large 73.9} & {\large \textbf{74.0}} & {\large \textbf{74.0}} &  & {\large 73.6} \small{(-2)} & {\large \textbf{74.0}} & {\large 73.8} \small{(.002)} \\
 & L-WHL & {\large 72.0} & {\large 73.4} & {\large 73.2} & {\large 73.1} & {\large 73.0} &  & {\large 73.4} \small{(-2)} & {\large \textbf{73.5}} & {\large 73.4} \small{(.002)} \\
 & L-UNC-WHL & {\large 72.0} & {\large 73.0} & {\large 72.9} & {\large 72.8} & {\large 72.8} &  & {\large 65.4} \small{(-8)} & {\large \textbf{73.1}} & {\large \textbf{73.1}} \small{(.002)} \\ \hline
\multirow{2}{*}{XLNet} & B & {\large 69.1} & {\large 67.4} & {\large 64.8} & {\large 62.7} & {\large 63.7} &  & 55.4 \small{(-3)} & {\large \textbf{72.3}} & {\large \textbf{72.3}} \small{(.005)} \\
 & L & {\large 66.2} & {\large 70.4} & {\large 65.7} & {\large 64.8} & {\large 66.1} &  & 57.5 \small{(-17)} & {\large \textbf{73.8}} & {\large \textbf{73.8}} \small{(.005)} \\ \hline
\multirow{2}{*}{RoBERTa} & B & {\large 71.9} & {\large 73.3} & {\large 73.3} & {\large 73.4} & {\large 73.3} &  & {\large 73.5} \small{(-3)} & {\large 73.6} & {\large \textbf{73.7}} \small{(.002)} \\
 & L & {\large 71.2} & {\large 74.0} & {\large 74.1} & {\large 74.0} & {\large 73.9} &  & {\large 66.3} \small{(-10)} & {\large \textbf{74.7}} & {\large \textbf{74.7}} \small{(.002)} \\ \hline
\multirow{4}{*}{ALBERT} & B & {\large \textbf{70.6}} & {\large 69.6} & {\large 70.1} & {\large 70.1} & {\large 70.3} &  & {\large 67.3} \small{(-5)} & {\large 69.7} & {\large 69.7} \small{(.002)} \\
 & L & {\large 70.1} & {\large 70.5} & {\large 70.5} & {\large 70.6} & {\large 70.4} &  & {\large 67.7} \small{(-15)} & {\large \textbf{71.1}} & {\large 70.7} \small{(.002)} \\
 & XL & {\large 64.3} & {\large 69.0} & {\large 68.8} & {\large 67.8} & {\large 67.2} &  & {\large 66.6} \small{(-12)} & {\large \textbf{73.0}} & {\large \textbf{73.0}} \small{(.002)} \\
 & XXL & {\large 69.4} & {\large 73.7} & {\large 73.9} & {\large 73.1} & {\large 72.5} &  & {\large 74.8} \small{(-6)} & {\large \textbf{75.1}} & {\large \textbf{75.1}} \small{(.002)} \\ \bottomrule
\end{tabular}%
}
\caption{WSD test results (F1 on ALL). Reports conventional layer choices and alternatives using sense profiles. Recommended $t$ for WSD is 0.005. Sense representations for this experiment were learned from SemCor (with propagation).}
\label{tab:wsd_test}
\end{table}

\clearpage

\begin{table}
\centering
\resizebox{0.95\textwidth}{!}{%
\begin{tabular}{clccccccc} \toprule
\multicolumn{2}{c}{\multirow{2}{*}{\textbf{Model}}} & \textbf{Sum} & \textbf{Layer} & \multicolumn{5}{c}{\textbf{Weighted Sum (WS)}} \\  \cline{5-9}
\multicolumn{2}{c}{} & \textbf{\small{LST4}} & \textbf{\small{Best}} & \textbf{\small{t=0.002}} & \textbf{\small{t=0.005}} & \textbf{\small{t=0.010}} & \textbf{\small{t=0.100}} & \textbf{\small{t=1.000}} \\ \midrule \midrule
\multirow{6}{*}{BERT} & B & {\large 59.2} & {\large 61.2} \small{(-6)} & {\large 61.5} & {\large 61.5} & {\large 61.8} & {\large \textbf{62.2}} & {\large 62.0} \\
 & B-UNC & {\large 58.6} & {\large 63.1} \small{(-8)} & {\large 63.1} & {\large 63.2} & {\large 63.5} & {\large \textbf{63.7}} & {\large \textbf{63.7}} \\
 & L & {\large 57.7} & {\large 63.8} \small{(-14)} & {\large 63.9} & {\large 64.0} & {\large 63.9} & {\large \textbf{64.5}} & {\large \textbf{64.5}} \\
 & L-UNC & {\large 56.8} & {\large 64.3} \small{(-14)} & {\large 64.4} & {\large 64.6} & {\large 64.6} & {\large 65.4} & {\large \textbf{65.7}} \\
 & L-WHL & {\large 59.7} & {\large 62.1} \small{(-14)} & {\large 62.1} & {\large 62.0} & {\large 62.1} & {\large \textbf{62.7}} & {\large 62.5} \\
 & L-UNC-WHL & {\large 59.1} & {\large \textbf{64.8 }\small{(-14)}} & {\large 64.7} & {\large 64.5} & {\large 64.5} & {\large 64.7} & {\large \textbf{64.8}} \\ \midrule
\multirow{2}{*}{XLNet} & B & {\large 34.7} & {\large 61.2} \small{(-9)} & {\large 61.3} & {\large 61.4} & {\large 61.4} & {\large \textbf{61.9}} & {\large 60.4} \\
 & L & {\large 28.0} & {\large 63.6} \small{(-18)} & {\large 63.5} & {\large 63.6} & {\large 63.7} & {\large \textbf{64.4}} & {\large 64.1} \\ \midrule
\multirow{2}{*}{RoBERTa} & B & {\large 61.6} & {\large \textbf{64.2 }\small{(-9)}} & {\large 64.0} & {\large 63.9} & {\large 64.0} & {\large 64.0} & {\large 63.9} \\
 & L & {\large 64.1} & {\large 65.3} \small{(-17)} & {\large 65.2} & {\large 65.4} & {\large 65.8} & {\large 66.1} & {\large \textbf{66.2}} \\ \midrule
\multirow{4}{*}{ALBERT} & B & {\large 60.0} & {\large 61.8} \small{(-8)} & {\large \textbf{61.9}} & {\large 61.8} & {\large 61.7} & {\large 61.6} & {\large 61.6} \\
 & L & {\large 60.0} & {\large \textbf{64.1 }\small{(-16)}} & {\large 63.5} & {\large 63.5} & {\large 63.2} & {\large 63.3} & {\large 63.2} \\
 & XL & {\large 54.9} & {\large 64.5} \small{(-18)} & {\large 64.2} & {\large 64.1} & {\large 64.2} & {\large 64.5} & {\large \textbf{64.6}} \\
 & XXL & {\large 65.8} & {\large 65.8} \small{(-9)} & {\large 65.7} & {\large 66.1} & {\large 66.1} & {\large 66.2} & {\large \textbf{66.3}} \\ \bottomrule
\end{tabular}%
}
\caption{USM validation results (F1). Reports best single layer and weighted sums using specific sense profiles, with different $t$ values, for each model configuration. Sense representations for this experiment were learned from SemCor (no propagation required).}
\label{tab:usm_dev}
\end{table}

\begin{table}
\centering
\resizebox{0.95\textwidth}{!}{%
\begin{tabular}{clccccclccc}  \toprule
\multicolumn{2}{c}{\multirow{3}{*}{\textbf{Model}}} & \multicolumn{5}{c}{\textbf{Standard}} &  & \multicolumn{3}{c}{\textbf{Proposed}} \\ \cline{3-7} \cline{9-11} 
\multicolumn{2}{c}{} & \textbf{Layer} & \textbf{Layer} & \textbf{Sum} & \textbf{WS (I)} & \textbf{WS (F)} &  & \textbf{Layer} & \textbf{WS} & \textbf{WS} \\
\multicolumn{2}{c}{} & \textbf{\small{-1}} & \textbf{\small{-2}} & \textbf{\small{LST4}} & \textbf{\small{LST4}} & \textbf{\small{LST4}} &  & \textbf{\small{Best Dev}} & \textbf{\small{Rec. t}} & \textbf{\small{Best t}} \\ \hline \hline
\multirow{6}{*}{BERT} & B & \large{53.1} & {\large 51.2} & {\large 53.7} & {\large 53.0} & {\large 53.0} &  & {\large 57.0} \small{(-6)} & {\large \textbf{57.7}} & {\large \textbf{57.7}} \small{(.100)}\\
 & B-UNC & {\large 50.4} & {\large 50.4} & {\large 53.0} & {\large 51.8} & {\large 52.3} &  & {\large 57.9} \small{(-8)} & {\large \textbf{58.8}} & {\large \textbf{58.8}} \small{(.100)}\\
 & L & {\large 53.9} & {\large 50.3} & {\large 52.5} & {\large 52.8} & {\large 53.4} &  & {\large 58.9} \small{(-14)} & {\large \textbf{60.0}} & {\large \textbf{60.0}} \small{(.100)}\\
 & L-UNC & {\large 48.3} & {\large 46.7} & {\large 49.0} & {\large 48.8} & {\large 48.8} &  & {\large 58.5} \small{(-14)} & {\large 60.3} & {\large \textbf{60.4}} \small{(1.00)}\\
 & L-WHL & {\large 53.3} & {\large 54.3} & {\large 54.5} & {\large 54.2} & {\large 54.5} &  & {\large 57.6} \small{(-14)} & {\large \textbf{58.4}} & {\large \textbf{58.4}} \small{(.100)}\\
 & L-UNC-WHL & {\large 53.7} & {\large 52.4} & {\large 53.3} & {\large 53.3} & {\large 53.3} &  & {\large 58.3} \small{(-14)} & {\large 59.4} & {\large \textbf{59.6}} \small{(1.00)}\\ \midrule
\multirow{2}{*}{XLNet} & B & {\large 38.1} & {\large 36.9} & {\large 31.4} & {\large 27.8} & {\large 28.9} &  & {\large \textbf{57.3}} \small{(-9)} & {\large \textbf{57.3}} & {\large \textbf{57.3}} \small{(.100)}\\
 & L & {\large 27.9} & {\large 41.3} & {\large 28.7} & {\large 28.2} & {\large 29.6} &  & {\large 59.0} \small{(-18)} & {\large \textbf{60.4}} & {\large \textbf{60.4}} \small{(.100)}\\ \midrule
\multirow{2}{*}{RoBERTa} & B & {\large 53.7} & {\large 53.2} & {\large 55.1} & {\large 55.3} & {\large 55.5} &  & {\large 58.3} \small{(-9)} & {\large \textbf{59.2}} & {\large \textbf{59.2}} \small{(.100)}\\
 & L & {\large 56.3} & {\large 56.7} & {\large 57.9} & {\large 58.1} & {\large 58.0} &  & {\large 60.6} \small{(-17)} & {\large \textbf{61.2}} & {\large \textbf{61.2}} \small{(.100)}\\ \midrule
\multirow{4}{*}{ALBERT} & B & {\large 53.8} & {\large 53.6} & {\large 54.7} & {\large 54.2} & {\large 54.8} &  & {\large 55.8} \small{(-8)} & {\large 56.2} & {\large \textbf{56.3}} \small{(.002)}\\
 & L & {\large 55.7} & {\large 55.5} & {\large 56.0} & {\large 56.1} & {\large 56.1} &  & {\large 57.4} \small{(-16)} & {\large \textbf{58.4}} & {\large 57.3} \small{(.005)}\\
 & XL & {\large 41.2} & {\large 48.5} & {\large 49.8} & {\large 48.0} & {\large 47.4} &  & {\large 59.9} \small{(-18)} & {\large \textbf{60.1}} & {\large 59.8} \small{(1.00)}\\
 & XXL & {\large 55.1} & {\large 60.6} & {\large 61.7} & {\large 61.0} & {\large 60.5} &  & {\large 60.6} \small{(-9)} & {\large \textbf{62.3}} & {\large \textbf{62.3}} \small{(1.00)}\\ \bottomrule
\end{tabular}%
}
\caption{USM test results (F1 on ALL). Reports conventional layer choices and alternatives using sense profiles. Recommended $t$ for USM is 0.1. Sense representations for this experiment were learned from SemCor (with propagation).}
\label{tab:usm_test}
\end{table}

\clearpage

\section{Evaluation}
\label{eval:intro}

In this work we address several sense-related tasks selected to investigate the versatility of the proposed sense embeddings, covering disambiguation ($\S$\ref{eval:wsd} - WSD), matching ($\S$\ref{eval:usm} - USM), meaning change detection ($\S$\ref{eval:wic} and $\S$\ref{eval:gcs} - WiC and GWCS) and sense similarity ($\S$\ref{eval:sid} - SID).
For each task, we report our new results (LMMS-SP) in comparison with the state-of-the-art and the original LMMS \citeyearpar{loureiro-jorge-2019-language} sense embeddings.

For brevity, we only consider the variant from each model family that showed best results in our probing analysis ($\S$\ref{probing:results}).
In our comparisons, we omit LMMS$_{2348}$ because those sense embeddings are concatenated with fastText \citep{bojanowski-etal-2017-enriching} word embeddings, thus not exclusively based on representations from particular NLMs, as focused in this work.

All tasks are solved essentially using cosine similarity between contextual embeddings and LMMS-SP precomputed sense embeddings represented using the same NLM.
Each task's subsection provides more details about how these similarities are used to produce task-specific predictions.
No additional task-specific training or validation datasets are used asides from those referred in Section \ref{exp:datasets}, and all NLMs are employed in the same exact fashion - simply retrieving contextualized representations from each layer (following $\S$\ref{met:embed}).

As such, LMMS-SP performance on these tasks should be indicative of each NLM's intrinsic ability to approximate meaning representations learned during pre-training with language modelling objectives alone.

\subsection{Word Sense Disambiguation (WSD)}
\label{eval:wsd}

\begin{table}[hbt]
\centering
\resizebox{1.0\textwidth}{!}{
\begin{tabular}{lccc} \toprule
\textbf{Sentence} & \textbf{Lemma} & \textbf{POS} & \textbf{Gold Sensekey} \\ \midrule \midrule
\makecell[l]{Eyes that were clear , but also bright with a strange \\ intensity , a sort of cold \textbf{fire} burning behind them .} & fire & NOUN & fire\%1:12:00:: \\ \bottomrule
\end{tabular}
}
\caption{Example WSD instance from \citet{raganato-etal-2017-word}. Sentence, lemma and part-of-speech (POS) are provided. The goal is to predict the correct sensekey (sense from WordNet).}
\label{tab:wsd_examples}
\end{table}

WSD is the most popular and obvious task for evaluating sense embeddings.
This task has been researched since the early days of Artificial Intelligence and constitutes an AI-complete task \citep{Navigli2009WordSD}.
It is usually formulated as choosing the correct sense for a word in context out of a list of possible senses given the word's lemma and part-of-speech tag (see Table \ref{tab:wsd_examples}).
Several test sets have been proposed over the years, and the compilation of \citet{raganato-etal-2017-word} has emerged as the de facto evaluation framework for English WSD, which we also use.
Naturally, this task suits sense profiles for WSD, and we follow the method described in Section \ref{met:apply}.


\subsubsection{Results}
On Table \ref{tab:wsd_comparison} we report performance on the standard test sets of the WSD Evaluation Framework \citep{raganato-etal-2017-word}. 
Given the breadth of recent WSD solutions, we make results more comparable by separating solutions using only SemCor annotations, and solutions augmenting SemCor with other sense-annotated datasets.
In the case of LMMS-SP, we combine SemCor with the unambiguous annotations from UWA, which are easily retrieved from unlabeled corpora.
We also report which solutions use glosses and relations, besides sense annotations, as well as which solutions are based on 1NN (first nearest neighbor) with precomputed sense embeddings represented in the space of NLMs.

When considering SemCor as the only source of annotations, LMMS and LMMS-SP remain the best solutions based on 1NN in NLM-space.
Most notably, LMMS-SP\textsubscript{ALBERT-XXL} is able to match the performance of LMMS$_{2048}$ on the combination of test sets (ALL) without concatenating gloss embeddings.

As could be expected, task-specific classifiers show best results, particularly BEM \citep{blevins-zettlemoyer-2020-moving} and ConSeC \citep{barba-etal-2021-consec}.
Generally, solutions fine-tuning NLMs, or combining them with other classifiers trained for the WSD task show improved performance over 1NN.
Despite this, in \citet{loureiro-etal-analysis-2021} we have shown that 1NN solutions offer other advantages, such as better sample efficiency and less frequency biases, besides the versatility advocated in this current work.

Allowing for additional annotations, we find that LMMS-SP results improve slightly when using {\small BERT-L} and {\small ALBERT-XXL}.
ARES \citep{scarlini-etal-2020-contexts} uses semi-supervised annotations to increase coverage of the sense inventory with sense embeddings represented on the space of {\small BERT-L}.
Results show the ARES dataset leads to improved WSD performance in comparison to LMMS-SP on the reported test sets, particularly on SE13 and SE15.\footnote{We expect the annotations in ARES to produce further performance gains for LMMS-SP but do not use this resource due to its large size (13x the annotations in SemCor+UWA{\small 10}).}

\begin{table}[htb]
\centering
\resizebox{0.96\textwidth}{!}{
\begin{tabular}{clccccccccc} \toprule
 & \multirow{2}{*}{\textbf{\large{Model}}} & \multirow{2}{*}{\textbf{1NN}} & \multirow{2}{*}{\textbf{Defs.}} & \multirow{2}{*}{\textbf{Rels.}} & \textbf{SE2} & \textbf{SE3} & \textbf{SE07} & \textbf{SE13} & \textbf{SE15} & \textbf{ALL} \\
 & & & & & \scriptsize{(n=2,282)} & \scriptsize{(n=1,850)} & \scriptsize{(n=445)} & \scriptsize{(n=1,644)} & \scriptsize{(n=1,022)} & \scriptsize{(n=7,253)} \\ \midrule \midrule
\multirow{15}{*}{\rotatebox{90}{\textbf{\small{SemCor (SC)}}}} & MFS        &  &  &  & \large{65.6} & \large{66.0} & \large{54.5} & \large{63.8} & \large{67.1} & \large{64.8} \\
 & context2vec \citeyearpar{melamud-etal-2016-context2vec}             & \checkmark & & & \large{71.8} & \large{69.1} & \large{61.3} & \large{65.6} & \large{71.9} & \large{69.0} \\
 & ELMo \citeyearpar{peters-etal-2018-deep}                            & \checkmark & & & \large{71.6} & \large{69.6} & \large{62.2} & \large{66.2} & \large{71.3} & \large{69.0} \\
 & BERT-L \citeyearpar{loureiro-jorge-2019-language}                   & \checkmark &  &  & \large{76.3} & \large{73.2} & \large{66.2} & \large{71.7} & \large{74.1} & \large{73.5} \\
 & SVC \citeyearpar{vial-etal-2019-sense}                    &  &  & \checkmark & \large{76.6} & \large{76.9} & \large{69.0} & \large{73.8} & \large{75.4} & \large{75.4} \\
 & GlossBERT \citeyearpar{huang-etal-2019-glossbert}                   &  & \checkmark &  & \large{77.7} & \large{75.2} & \large{72.5*} & \large{76.1} & \large{80.4} & \large{77.0*} \\
 & EWISER \citeyearpar{bevilacqua-navigli-2020-breaking}                   &  & \checkmark & \checkmark & \large{78.9} & \large{78.4} & \large{71.0} & \large{78.9} & \large{79.3*} & \large{78.3*} \\
 & BEM \citeyearpar{blevins-zettlemoyer-2020-moving}                   &  & \checkmark &  & \large{79.4} & \large{77.4} & \large{74.5*} & \large{79.7} & \large{81.7} & \large{79.0*} \\
 & ConSeC \citeyearpar{barba-etal-2021-consec}                   &  & \checkmark &  & \large{82.3} & \large{79.9} & \large{77.4*} & \large{83.2} & \large{85.2} & \large{82.0*} \\ \cline{2-11}
 & LMMS\textsubscript{1024} \citeyearpar{loureiro-jorge-2019-language} & \checkmark &  & \checkmark & \large{75.4} & \large{74.0} & \large{66.4} & \large{72.7} & \large{75.3} & \large{73.8} \\
 & LMMS\textsubscript{2048} \citeyearpar{loureiro-jorge-2019-language} & \checkmark & \checkmark & \checkmark & \large{76.3} & \underline{\large{75.6}} & \large{68.1} & \large{75.1} & \large{77.0} & \underline{\large{75.4}} \\
 & LMMS-SP\textsubscript{BERT-L}                                       & \checkmark & \checkmark & \checkmark & \large{76.1} & \large{74.0} & \large{67.0} & \large{75.2} & \underline{\large{77.4}} & \large{75.0} \\
 & LMMS-SP\textsubscript{XLNet-L}                                      & \checkmark & \checkmark & \checkmark & \large{76.0} & \large{73.1} & \large{66.4} & \large{74.2} & \large{74.9} & \large{74.1} \\
 & LMMS-SP\textsubscript{RoBERTa-L}                                    & \checkmark & \checkmark & \checkmark & \large{77.2} & \large{73.5} & \large{67.9} & \underline{\large{75.5}} & \large{76.4} & \large{75.2} \\
 & LMMS-SP\textsubscript{ALBERT-XXL}                                   & \checkmark & \checkmark & \checkmark & \underline{\large{77.4}} & \large{74.8} & \underline{\large{71.0}} & \large{74.7} & \large{74.8} & \underline{\large{75.4}} \\ \midrule \midrule
\multirow{4}{*}{\rotatebox{90}{\textbf{\small{SC+Others}}}} & SVC \citeyearpar{vial-etal-2019-sense} &  &  & \checkmark & \large{79.4} & \large{78.1} & \large{71.4} & \large{77.8} & \large{81.4} & \large{78.5} \\
 & KnowBERT \citeyearpar{peters-etal-2019-knowledge}                   &  &  & \checkmark & \large{76.4} & \large{76.0} & \large{71.4} & \large{73.1} & \large{75.4} & \large{75.1} \\
 & EWISER \citeyearpar{bevilacqua-navigli-2020-breaking}               &  & \checkmark & \checkmark & \large{80.8} & \large{79.0} & \large{75.2} & \large{80.7} & \large{81.8*} & \large{80.1*} \\
 & ARES \citeyearpar{scarlini-etal-2020-contexts}                      & \checkmark & \checkmark &  & \large{78.0} & \large{77.1} & \large{71.0} & \large{77.3} & \large{83.2} & \large{77.9} \\ \cline{1-11}
\multirow{4}{*}{\rotatebox{90}{\textbf{\small{SC+UWA}}}} & LMMS-SP\textsubscript{BERT-L}                                       & \checkmark & \checkmark & \checkmark & \large{76.7} & \large{74.1} & \large{66.4} & \large{75.2} & \underline{\large{77.6}} & \large{75.2} \\
 & LMMS-SP\textsubscript{XLNet-L}                                      & \checkmark & \checkmark & \checkmark & \large{76.1} & \large{73.1} & \large{65.9} & \large{74.2} & \large{75.0} & \large{74.1} \\
 & LMMS-SP\textsubscript{RoBERTa-L}                                    & \checkmark & \checkmark & \checkmark & \large{77.4} & \large{73.5} & \large{67.7} & \underline{\large{75.3}} & \large{76.7} & \large{75.2} \\
 & LMMS-SP\textsubscript{ALBERT-XXL}                                   & \checkmark & \checkmark & \checkmark & \underline{\large{77.7}} & \underline{\large{75.0}} & \underline{\large{70.5}} & \large{74.7} & \large{74.9}& \underline{\large{75.5}} \\ \bottomrule
\end{tabular}
}
\caption{F1 scores (\%) for each test set in the WSD Evaluation Framework \citep{raganato-etal-2017-word}. Top rows show results for models using sense annotations exclusively from SemCor (SC). Bottom rows show results for models augmenting SC with annotations from additional sources. Results marked with * correspond to development sets (and therefore ALL). For each group of results, we underline the best from LMMS.}
\label{tab:wsd_comparison}
\end{table}

\subsection{Uninformed Sense Matching (USM)}
\label{eval:usm}

\begin{table}[ht]
\centering
\resizebox{\textwidth}{!}{%
\begin{tabular}{@{}lcc@{}} \toprule
\textbf{Sentence} & \textbf{Gold Sensekey} & \textbf{Gold Synset} \\ \midrule \midrule
\multirow{2}{*}{\makecell[l]{Eyes that were clear , but also bright with a strange \\ intensity , a sort of cold \textbf{fire} burning behind them .}} & \multirow{2}{*}{fire\%1:12:00::} & 06711159n \\
 &  & (fire$_n^9$) \\ \bottomrule
\end{tabular}%
}
\caption{Example USM instance adapted from \citet{raganato-etal-2017-word}. The correct sensekey or synset must be predicted, in separate evaluations.}
\label{tab:usm_examples}
\end{table}

We introduced the USM task in \citet{loureiro-jorge-2019-language} as a variation on WSD that can more accurately represent the extent to which NLMs can associate words or phrases to senses from the WordNet inventory.   
The crucial difference in relation to WSD is that in the USM task we do not use any supplemental information to restrict candidates in the sense inventory (compare examples in Table \ref{tab:wsd_examples} and Table \ref{tab:usm_examples}).
Conveniently, this allows for USM to use the same test sets as WSD.
As expected, we address USM using the sense profile of the same name, and follow the method described in Section \ref{met:apply}.
In this work we evaluate USM from both the sensekey and synset perspective, to provide a clearer account of the impact of lexical information on task performance.

\subsubsection{Results}
Following \citet{loureiro-camacho-collados-2020-dont}, we evaluate performance considering two additional metrics besides F1: Precision at 5 (P@5) and Mean Reciprocal Rank (MRR).
To our knowledge, ARES \citep{scarlini-etal-2020-contexts} is the only other publicly available set of full-coverage sense embeddings represented in the space of a Transformer-based NLM, so we also compare LMMS-SP against those sense embeddings.
Since our prior LMMS sense embeddings and the ARES sense embeddings are released using sensekey representations, USM synset evaluation requires converting those sensekey embeddings to synset embeddings.
We perform this conversion by simply averaging sensekey embeddings that belong to the same synset.
In Section \ref{analysis:annotations} we analyse the impact this conversion can have on task performance.

On Table \ref{tab:usm_comparison} it can be observed that LMMS-SP dramatically improves performance over LMMS on all three metrics considered.
The poor performance of LMMS$_{2048}$ in comparison to LMMS$_{1024}$ suggests that concatenating gloss embeddings is detrimental to USM performance, particularly on the F1 metric.
In this comparison we do not consider LMMS$_{2348}$ because those sense embeddings are concatenated with fastText static embeddings, resulting in 300 dimensions having the same exact distribution for sense embeddings corresponding to identical lemmas.
This property of LMMS$_{2348}$ makes the comparison inequitable and diverts from this work's focus on the intrinsic capabilities of Transformer NLMs.

Interestingly, we find that, when targeting sensekeys, LMMS-SP\textsubscript{ALBERT-XXL} shows best performance on all metrics, and ARES (based on BERT-L) only outperforms LMMS-SP\textsubscript{BERT-L} on the F1 metric.
However, when targeting synsets, the additional contexts of ARES prove more advantageous, and we do not observe a similar performance gap between sensekeys and synset as we do with LMMS-SP, which can be expected considering that the additional contexts of ARES are targeted at the synset-level.

\begin{table}[ht]
\centering
\resizebox{0.8\textwidth}{!}{%
\begin{tabular}{l>{\centering\arraybackslash}p{3em}>{\centering\arraybackslash}p{3em}>{\centering\arraybackslash}p{3em}l>{\centering\arraybackslash}p{3em}>{\centering\arraybackslash}p{3em}>{\centering\arraybackslash}p{3em}} \toprule
\multirow{2}{*}{\textbf{Model}} & \multicolumn{3}{c}{\textbf{Sensekeys}} &  & \multicolumn{3}{c}{\textbf{Synsets}} \\ \cline{2-4} \cline{6-8} 
 & \textbf{F1} & \textbf{P@5} & \textbf{MRR} &  & \textbf{F1} & \textbf{P@5} & \textbf{MRR} \\ \midrule \midrule
ARES & \large{61.4} & \large{84.7} & \large{71.8} &  & \large{\textbf{60.7}}$^\dagger$ & \large{\textbf{86.5}}$^\dagger$ & \large{\textbf{71.8}}$^\dagger$ \\
LMMS\textsubscript{1024} \citeyearpar{loureiro-jorge-2019-language} & \large{52.2} & \large{66.9} & \large{59.0} &  & \large{29.4}$^\dagger$ & \large{53.9}$^\dagger$ & \large{40.7}$^\dagger$ \\
LMMS\textsubscript{2048} \citeyearpar{loureiro-jorge-2019-language} & \large{34.8} & \large{60.3} & \large{46.3} &  & \large{32.5}$^\dagger$ & \large{58.9}$^\dagger$ & \large{44.5}$^\dagger$ \\
LMMS-SP\textsubscript{BERT-L}     & \large{60.8} & \large{86.7} & \large{72.2} &  & \large{51.0} & \large{81.7} & \large{64.3} \\
LMMS-SP\textsubscript{XLNet-L}    & \large{60.1} & \large{87.3} & \large{71.9} &  & \large{51.7} & \large{82.7} & \large{65.1} \\
LMMS-SP\textsubscript{RoBERTa-L}  & \large{62.2} & \large{86.9} & \large{73.1} &  & \large{50.2} & \large{80.1} & \large{63.3} \\
LMMS-SP\textsubscript{ALBERT-XXL} & \large{\textbf{62.9}} & \large{\textbf{87.6}} & \large{\textbf{73.7}} &  & \large{52.7} & \large{81.9} & \large{65.5} \\ \bottomrule
\end{tabular}%
}
\caption{USM results on the ALL test set of the WSD Evaluation Framework \citep{raganato-etal-2017-word}, at sense and synset-level. Results marked with $\dagger$ are obtained from synset embeddings converted from sensekey embeddings.}
\label{tab:usm_comparison}
\end{table}

\subsection{Word-in-Context (WiC)}
\label{eval:wic}

\begin{table}[ht]
\centering
\resizebox{0.9\textwidth}{!}{
\begin{tabular}{lccc} \toprule
\textbf{Sentence Pairs} & \textbf{Lemma} & \textbf{POS} & \textbf{Boolean} \\ \midrule \midrule
\makecell[l]{You must \textbf{carry} your camping gear .\\Sound \textbf{carries} well over water .} & carry & VERB & False \\ \midrule
\makecell[l]{He wore a jock strap with a metal \textbf{cup} .\\Bees filled the waxen \textbf{cups} with honey .} & cup & NOUN & True \\ \bottomrule
\end{tabular}
}
\caption{Examples from the WiC training set. Showing two independent instances.}
\label{tab:wic_examples}
\end{table}

The Word-in-Context \citep[WiC]{pilehvar-camacho-collados-2019-wic} task is designed to assess how context impacts word representations produced by contextual NLMs.
It is a binary classification task that simply requires determining whether a particular word is used with the same meaning or not in a  pair of sentences, also given lemma and POS provided in WSD tasks (see Table \ref{tab:wic_examples} for examples).
The dataset is balanced and performance is measured with accuracy.

\subsubsection{Solution}
In this work, we tackle the WiC task using our proposed sense embeddings following the unsupervised approach from \citet{loureiro-jorge-2019-liaad}, which essentially applies the 1NN method for disambiguating the target word in both sentences and checks whether they are equal or not.
Even though we also explored a supervised approach in \citet{loureiro-jorge-2019-liaad}, based on Logistic Regression, in this work we focus on the unsupervised approach as its performance is more revealing of the inherent representational abilities of NLMs.
Given the close relation to disambiguation, we use WSD sense profiles for WiC.

\subsubsection{Results}
WiC is a benchmark NLU task, being part of SuperGLUE \citep{NEURIPS2019_4496bf24}, therefore most state-of-the-art NLMs have reported results for this task.
The initial baseline methods proposed with WiC were based on cosine similarity with thresholds learned from the validation set.

Most recent solutions, however, involve fine-tuning the NLM (as performed for other sentence classification tasks in SuperGLUE) using the training and validation sets provided with WiC.
One notable exception is \citet{scarlini-etal-2020-contexts} which proposed a method that leverages ARES sense embeddings to improve the fine-tuning process.
As such, on Table \ref{tab:wic_comparison} we compare results from these solutions to our unsupervised LMMS and LMMS-SP, as well as an unsupervised result based on the same 1NN approach using the ARES embeddings.

Starting with our unsupervised results, we confirm that LMMS-SP\textsubscript{BERT-L} surpasses the performance of LMMS\textsubscript{2048} (based on {\small BERT-L}), and once again LMMS-SP\textsubscript{ALBERT-XXL} displays the best performance.
Nevertheless, supervised solutions using NLMs fine-tuned for this task show best performance overall, particularly T5 \citep{JMLR:v21:20-074} which is currently the largest NLM with reported results on this task, at over 11B parameters.
KnowBERT \citep{peters-etal-2019-knowledge} and SenseBERT \citep{levine-etal-2020-sensebert} are both NLMs based on BERT that have been augmented with sense information from WordNet and SemCor, among other resources, showing improved performance in comparison to fine-tuning the original {\small BERT-L}.

The method used by \citet{scarlini-etal-2020-contexts} to employ sense embeddings while fine-tuning {\small BERT-L} for WiC resulted in a notable improvement similar to SenseBERT.
In the unsupervised setting, however, we found that ARES embeddings outperform LMMS-SP\textsubscript{BERT-L}, but underperform both LMMS-SP\textsubscript{RoBERTa-L} and LMMS-SP\textsubscript{ALBERT-XXL}. 
We expect following the same method to assist supervised fine-tuning with LMMS-SP sense embeddings may produce improved results, but consider that experiment out of scope for this work.

As for solutions using the threshold method, all reported models substantially underperform unsupervised results using any Transformer-based NLM.

\begin{table}[ht]
\centering
\resizebox{0.8\textwidth}{!}{%
\begin{tabular}{llllc} \toprule
\textbf{} & \textbf{Method} & \textbf{Language Model} & \textbf{Sense Embeddings} & \multicolumn{1}{c}{\textbf{Acc.}} \\ \midrule \midrule
\multirow{7}{*}{\rotatebox{90}{\textbf{\small{Supervised}}}} & Fine-Tuning & BERT-L \citeyearpar{NEURIPS2019_4496bf24} & - & \large{69.6*} \\
 & Logistic Reg. & BERT-L & LMMS\textsubscript{2048} \citeyearpar{loureiro-jorge-2019-liaad} & \large{68.1} \\
 & Fine-Tuning & RoBERTa-L \citeyearpar{DBLP:journals/corr/abs-1907-11692} & - & \large{69.9*} \\
 & Fine-Tuning & KnowBERT \citeyearpar{peters-etal-2019-knowledge} & - & \large{70.9} \\
 & Fine-Tuning & SenseBERT \citeyearpar{levine-etal-2020-sensebert} & - & \large{72.1} \\
 & Fine-Tuning & T5 \citeyearpar{JMLR:v21:20-074} & - & \textbf{\large{76.9*}} \\
 & Fine-Tuning & BERT-L & ARES \citeyearpar{scarlini-etal-2020-contexts} & \large{72.2*} \\ \midrule
\multirow{6}{*}{\rotatebox{90}{\textbf{\small{Threshold}}}} & 1NN WSD & - & JBT \citeyearpar{pelevina-etal-2016-making} & \large{53.6} \\
 & 1NN WSD & - & DeConf \citeyearpar{pilehvar-collier-2016-de} & \large{58.7} \\
 & 1NN WSD & context2vec \citeyearpar{melamud-etal-2016-context2vec} & - & \large{\textbf{59.3}} \\
 & 1NN WSD & - & SW2V \citeyearpar{mancini-etal-2017-embedding} & \large{58.1} \\
 & 1NN WSD & ELMo \citeyearpar{peters-etal-2018-deep} & - & \large{57.7} \\
 & 1NN WSD & - & LessLex \citeyearpar{COLLA2020106346} & \large{59.2} \\ \midrule
\multirow{6}{*}{\rotatebox{90}{\textbf{\small{Unsupervised}}}} & 1NN WSD & BERT-L & ARES (2020) & \large{67.6} \\
 & 1NN WSD & BERT-L & LMMS\textsubscript{2048} \citeyearpar{loureiro-jorge-2019-language} & \large{66.3} \\
 & 1NN WSD & BERT-L & LMMS-SP\textsubscript{BERT-L} & \large{67.4} \\ 
 & 1NN WSD & XLNet-L & LMMS-SP\textsubscript{XLNet-L} & \large{66.1} \\
 & 1NN WSD & RoBERTa-L & LMMS-SP\textsubscript{RoBERTa-L} & \large{67.8} \\
 & 1NN WSD & ALBERT-XXL & LMMS-SP\textsubscript{ALBERT-XXL} & \textbf{\large{67.9}} \\ \bottomrule
\end{tabular}%
}
\caption{Results (Accuracy) on the test set of the WiC task comparing our unsupervised approach to the state-of-art. Best results for each approach reported in bold. Our results were obtained from the Codalab online platform. Results marked with * used the SuperGLUE version of the WiC test set, which has minor preprocessing differences.}
\label{tab:wic_comparison}
\end{table}


\subsection{Graded Word Similarity in Context (GWCS)}
\label{eval:gcs}

\begin{table}[ht]
\centering
\resizebox{\textwidth}{!}{%
\begin{tabular}{@{}clcc@{}} \toprule
\multicolumn{2}{c}{\textbf{Contexts}} & \textbf{Sim.} & \textbf{Change} \\ \midrule \midrule
\multirow{3}{*}{A}
 & Tim Drake \textbf{keeps} a memorial for her in his cave hideout underneath Titans & \multirow{3}{*}{\large{4.44}} & \multirow{6}{*}{\large{-0.52}} \\
 & Tower in San Francisco. [...] It is later revealed that Dr. Leslie Thompkins &  &  \\
 & had faked her death after the gang war in an effort to \textbf{protect} her. &  &  \\ \cmidrule(r){1-3}
\multirow{3}{*}{B}
 & Shisa are wards, believed to \textbf{protect} from various evils. When found in & \multirow{3}{*}{\large{3.92}} &  \\
 & pairs, the shisa on the left traditionally [...] The open mouth to ward off &  &  \\
 & evil spirits, and the closed mouth to \textbf{keep} good spirits in. &  & \\ \bottomrule
\end{tabular}%
}
\caption{Example from the practice set of GWCS (single instance). Contexts A and B each have corresponding similarity ratings for the same `keep'-`protect' word pair.}
\label{tab:gwcs_examples}
\end{table}

For evaluating graded contextual similarity, in contrast to the binary contextual similarity assignments of WiC, we address SemEval 2020 Task 3: Graded Word Similarity in Context (GWCS) \citep{armendariz-etal-2020-semeval}.
This task, based on the CoSimLex resource \citep{armendariz-etal-2020-cosimlex}, targets word pairs used for evaluating distributional semantic models (not necessarily polysemous words) in contexts spanning multiple sentences.
The task is divided into two sub-tasks derived from human-annotated similarity ratings: 1) predict the change in similarity between two different contexts for each word pair; 2) predict the similarity ratings themselves.
Table \ref{tab:gwcs_examples} shows a single example from GWCS, featuring two contexts each with occurrences of the same pair of words, context specific similarity ratings, and the associated similarity change.

\subsubsection{Solution}
While the sub-tasks are independently evaluated, we employ essentially the same method for both, based on our straightforward approach for the WiC task covered in Section \ref{eval:wic}, with minor adjustments to quantify the observed change in similarity.
Given contexts $A$ and $B$, we disambiguate target words (each instance's word pair) in the corresponding contexts, and compute sense similarities $sim^A_{wsd}$ and $sim^B_{wsd}$ as the cosine similarity between the embeddings of the predicted senses.
Considering that disambiguation may predict the same senses, thus resulting in $sim^A_{wsd} = sim^B_{wsd}$ for many instances, we also compute contextual similarities $sim^A_{ctx}$ and $sim^B_{ctx}$ as the cosine similarity between the contextual embeddings of the target words.
Thus, we determine similarity scores specific to context A as $sim^A = \frac{1}{2}(sim^A_{wsd} + sim^A_{ctx})$, and similarity scores specific to context B as $sim^B = \frac{1}{2}(sim^B_{wsd} + sim^B_{ctx})$.
These context-specific similarities constitute our solutions to sub-task 2.
We determine the semantic change scores for sub-task 1 trivially as $sim^B - sim^A$.
Considering that this solution closely follows our solution for WiC, and that word pairs contained in this dataset tend to be closely related, we use the WSD sense profile for GWCS.

\subsubsection{Results}
Performance on sub-task 1 is measured with Pearson Uncentered Correlation between the system's scores and the average human annotations, and performance on sub-task 2 is measured with the harmonic mean of the Spearman and Pearson correlations between the system’s scores and the average human annotations.
On Table \ref{tab:gwcs_comparison} we report results using our sense embeddings (LMMS and LMMS-SP), using the ARES sense embeddings with our scoring method (and BERT-Large, pooling with the sum of last 4 layers), and the best reported results from other task participants (including post-evaluation, until 03/2021).
Similarly to WiC, the scores for the test sets are hidden from participants, both during evaluation (ended 03/2020) and post-evaluation periods (extends indefinitely), so all reported results are obtained from the online platform used by SemEval after submitting each system's predictions.

We observe that our straightforward method combining similarity between sense and contextual embeddings is able to outperform the solutions of other task participants (Leaderboard Best), most of which also relied on Transformer-based NLMs \citep{armendariz-etal-2020-semeval}.
Interestingly, GWCS shows wide variation in performance from the choice of NLM, with  LMMS-SP\textsubscript{XLNet-L} standing out with clearly best results on both sub-tasks\footnote{Complete leaderboard results are available on Appendix \ref{appendix:semeval20}. Additionally, Appendix \ref{appendix:scws} reports performance on the Stanford Contextual Word Similarities \citep{huang-etal-2012-improving} task, which inspired the GWCS and WiC tasks, with similar conclusions.}.

\begin{table}[ht]
\centering  
\resizebox{0.5\textwidth}{!}{%
\begin{tabular}{lcc} \toprule
\textbf{Model} & \textbf{Subtask1} & \textbf{Subtask2} \\ \midrule \midrule
Leaderboard Best$^\dagger$ & {\large 77.4}$^1$ & {\large 74.6}$^2$ \\ \midrule
ARES (\citeyear{scarlini-etal-2020-contexts}) & {\large 76.9} & {\large 74.5} \\
LMMS\textsubscript{1024} (\citeyear{loureiro-jorge-2019-language}) & {\large 74.1} & {\large 74.2} \\
LMMS\textsubscript{2048} (\citeyear{loureiro-jorge-2019-language}) & {\large 75.7} & {\large 74.5} \\
LMMS-SP\textsubscript{BERT-L} & {\large 76.2} & {\large 74.4} \\
LMMS-SP\textsubscript{XLNet-L} & \textbf{{\large 78.7}} & \textbf{{\large 76.6}} \\
LMMS-SP\textsubscript{RoBERTa-L} & {\large 75.7} & {\large 74.9} \\
LMMS-SP\textsubscript{ALBERT-XXL} & {\large 75.2} & {\large 71.8} \\ \bottomrule
\end{tabular}%
}
\caption{Results on both subtasks of SemEval 2020 Task 3. $^\dagger$ as of 03/2021, considering evaluation and post-evaluation submissions (Users: $^1$Ferryman, $^2$Alexa). ARES results obtained using the same method as LMMS, only replacing the corresponding sense embeddings.}
\label{tab:gwcs_comparison}
\end{table}

\subsection{Sense Similarity}
\label{eval:sid}

\begin{table}[ht]
\centering
\resizebox{0.8\textwidth}{!}{
\begin{tabular}{ccc} \toprule
\textbf{Synset 1} & \textbf{Synset 2} & \textbf{Similarity} \\ \midrule \midrule
08570634n (hayfield$_n^1$) & 08598301n (grassland$_n^1$) & \large{3.58} \\ \midrule
03169390n (decoration$_n^1$) & 03291741n (envelope$_n^2$) & \large{0.08} \\ \bottomrule
\end{tabular}
}
\caption{Two examples of paired synsets with human similarity ratings from the SID dataset. Showing synset identifiers after conversion to WordNet (more readable format in parenthesis).}
\label{tab:sid_examples}
\end{table}

All the tasks we considered so far (WSD, USM, WiC and GWCS) have evaluated sense embeddings by their utility for accurately matching or distinguishing word senses in particular contexts.
In this last task, we address intrinsic evaluation of sense embeddings, directly comparing distributional similarity between sense pairs against human similarity ratings.

We perform this evaluation using the Sense Identification Dataset \citep[SID]{COLLA2020106267}, which is based on the word pairs (nouns only) and human similarity ratings from SemEval-2017 Task 2 \citep{camacho-collados-etal-2017-semeval}, with the addition of mapping word pairs to particular senses in the BabelNet sense inventory (see examples on Table \ref{tab:sid_examples}).

\subsubsection{Task Adaptation}

We convert the BabelNet sense identifiers to synsets from WordNet 3.0 using the mapping provided by \citet{NavigliPonzetto:10}.
However, some instances cannot be mapped due to missing entries in WordNet, or, in rare cases, their mapping results in the two senses of the pair being equal, leading to a reduction of 492 instances to 377 mapped to WordNet.
We further split SID into different groups for additional insights.
We first separate the 354 pairs for which both senses are represented in the related works we compare against (overlapping), considering these are not always complete sets of WordNet sense embeddings.
Next, we breakdown the overlapping pairs into a set of the most polarized word pairs (i.e., similarity ratings $\leq$ 1 or $\geq$ 3), and another set containing only pairs with senses that are annotated in SemCor+UWA{\small 10} (observed).

\subsubsection{Solution}

We use cosine similarity between synset embeddings to correlate with human similarity ratings.
Since we are directly comparing embeddings of very different dimensionality, we apply truncated SVD to normalize them to 300 dimensions\footnote{We verified that SVD-reduced embeddings always outperform original embeddings.} (including related work).
The senses being compared range from completely unrelated (e.g., polyhedron$_n^1$; actor$_n^1$) to highly related or similar (e.g., actor$_n^1$; actress$_n^1$), so we use USM sense profiles for SID.

\subsubsection{Results}

Performance on SID is measured with Pearson correlation.
For completeness, we report performance of synset embeddings that are not based on contextual NLMs, including new results based on fastText  embeddings (trained on CommonCrawl)\footnote{fastText embeddings for a given synset are computed by averaging the word embeddings for each lemma that belongs to the input synset in WordNet.}.
Results for related works are based on sense embeddings provided by the authors, converting sensekeys to synsets by averaging the corresponding embeddings whenever required (as in Section \ref{eval:usm}).
The inter-annotator agreement on our full set (n=377) reaches 87.9, measured as averaged pairwise Pearson correlation of the original SemEval-2017 human similarity scores.

Results for the WordNet-subset of SID are shown on Table \ref{tab:sid_comparison}.
As can be observed, LMMS-SP substantially outperforms LMMS and related works.
As with GWCS, LMMS-SP\textsubscript{XLNet-L} stands out with clearly best results.
While LMMS-SP also outperforms most non-contextual embeddings, it still underperforms LessLex \citep{COLLA2020106346} embeddings, which are based on ensembles and learned using BabelNet.
We also note that LMMS-SP performs particularly well on the `Observed' set corresponding to senses learned from annotated corpora.
The performance gap between ARES and LMMS-SP\textsubscript{BERT-L} suggests that additional semi-supervised annotations for more senses may not suffice.
Finally, the `Polarized' set seems consistently easier than the full set, indicating that the most challenging pairs are those with moderate similarity ratings.

\begin{table}[ht]
\centering
\resizebox{0.9\textwidth}{!}{%
\begin{tabular}{llc>{\centering\arraybackslash}p{5em}>{\centering\arraybackslash}p{5em}>{\centering\arraybackslash}p{5em}>{\centering\arraybackslash}p{5em}} \toprule
&  &  & \multirow{2}{*}{\textbf{All}} & \multicolumn{3}{c}{\textbf{Overlapping}} \\ \cline{5-7} 
& \textbf{Synset} & \textbf{WN Full} &  & \textbf{All} & \textbf{Polarized} & \textbf{Observed} \\
& \textbf{Embeddings} & \textbf{Coverage} & \textbf{{\small (n=377)}} & \textbf{{\small (n=354)}} & \textbf{{\small (n=182)}} & \textbf{{\small (n=297)}} \\ \midrule \midrule
\multirow{4}{*}{\rotatebox{90}{\textbf{\small{Static}}}} & fastText \citeyearpar{bojanowski-etal-2017-enriching} & \checkmark & {\large 64.4} & {\large 63.5} & {\large 69.3} & {\large 65.4} \\
& NASARI\textsubscript{UMBC} \citeyearpar{camacho-collados-etal-2015-nasari} & & - & {\large 71.6} & {\large 79.1} & {\large 74.4} \\
& DeConf$\dagger$ \citeyearpar{pilehvar-collier-2016-de} & & {\large 75.1} & {\large 74.9} & {\large 80.6} & {\large 76.9} \\
& LessLex \citeyearpar{COLLA2020106346} & & \textbf{{\large 82.5}} & \textbf{{\large 82.3}} & \textbf{{\large 85.5}} & \textbf{{\large 85.1}} \\ \midrule
\multirow{7}{*}{\rotatebox{90}{\textbf{\small{Contextual}}}} & SensEmBERT \citeyearpar{scarlinietal:2020} & & {\large 66.9} & {\large 66.8} & {\large 74.6} & {\large 69.5} \\
& ARES$\dagger$ & \checkmark & {\large 70.6} & {\large 70.4} & {\large 80.5} & {\large 73.3} \\
& LMMS\textsubscript{2048}$\dagger$ (\citeyear{loureiro-jorge-2019-language}) & \checkmark & {\large 71.2} & {\large 72.2} & {\large 76.2} & {\large 76.3} \\
& LMMS-SP\textsubscript{BERT-L} & \checkmark & {\large 77.8} & {\large 77.8} & {\large 80.4} & {\large 83.1} \\
& LMMS-SP\textsubscript{XLNet-L} & \checkmark & \textbf{{\large 79.5}} & \textbf{{\large 79.6}} & \textbf{{\large 81.2}} & \textbf{{\large 84.5}} \\
& LMMS-SP\textsubscript{RoBERTa-L} & \checkmark & {\large 74.1} & {\large 74.2} & {\large 79.0} & {\large 80.9} \\
& LMMS-SP\textsubscript{ALBERT-XXL} & \checkmark & {\large 77.4} & {\large 77.2} & {\large 80.5} & {\large 81.4} \\ \bottomrule
\end{tabular}%
}
\caption{Performance (Pearson Correlation) on the adapted SID dataset. All reported embeddings feature 300 dimensions. Embeddings marked with $\dagger$ have been converted from sensekeys. LessLex and NASARI embeddings were converted from BabelNet to WordNet using the same mapping applied to the SID adaptation.}
\label{tab:sid_comparison}
\end{table}


\section{Analysis}
\label{analysis:intro}

In this section, we perform several ablation studies to better understand the impact of individual contributions we have introduced in this work's extension of LMMS.
These experiments target the same NLMs and tasks\footnote{Due to leaderboard submission limits, ablations for WiC use the validation set.} that we addressed in the previous evaluation section.
Our ablation analyses cover the impact of sense profiles ($\S$\ref{analysis:profiles}), UWA annotations ($\S$\ref{analysis:annotations}), merging gloss representations ($\S$\ref{analysis:glosses}) and indirect representation of synsets ($\S$\ref{analysis:indirect}).
Considering that part-of-speech is an important factor in disambiguation (and sense representation), we also report performance per part-of-speech using both LMMS and LMMS-SP sense embeddings on WSD and USM tasks ($\S$\ref{analysis:pos}).

\subsection{Choice of Sense Profiles}
\label{analysis:profiles}

On Table \ref{tab:profiles_tasks} we report performance according to the sense profile used for weighted pooling of contextual embeddings from NLMs (described in Section \ref{probing:profiles}), and using the sum of the last 4 layers (Sum-LST4), as commonly used in related work and the original LMMS \citep{loureiro-jorge-2019-language}.

Our first conclusion is that Sum-LST4 pooling is only appropriate for particular models and tasks (i.e., WSD and WiC w/{\small BERT-L}; WiC w/{\small RoBERTa-L}), but detrimental for most (specially any task w/{\small XLNet-L}; any model for USM).
However, our recommended choice of sense profile not only appears beneficial for WSD and USM tasks across all models (expected since the sense profiles are based on those tasks), but also for WiC, GWCS and SID.
In fact, out of 20 model-task combinations, we only find 3 exceptions: {\small RoBERTa-L} on WiC and GWCS, and {\small ALBERT-XXL} on GWCS (to a lesser extent).
Moreover, we confirm that tasks are sensitive to the choice between SP-WSD and SP-USM, which validate our task-specific recommendations.

\begin{table}[ht]
\centering
\resizebox{0.95\textwidth}{!}{%
\begin{tabular}{cC{5em}C{7.6em}C{7.6em}C{7.6em}C{7.6em}} \toprule
\textbf{Task (Metric)} & \textbf{Pooling} & \textbf{BERT-L} & \textbf{XLNet-L} & \textbf{RoBERTa-L} & \textbf{ALBERT-XXL} \\ \midrule \midrule
\multirow{3}{*}{\textbf{\makecell{WSD\\{\small (F1 on ALL)}}}} & Sum-LST4 & \underline{{\large 75.2}} & {\large 56.4} & {\large 74.8} & {\large 73.8} \\
 & SP-WSD $\star$ & \underline{{\large 75.2}} & \underline{{\large 74.1}} & \underline{{\large 75.2}} & \textbf{\underline{{\large 75.5}}} \\
 & SP-USM & {\large 72.9} & {\large 73.4} & {\large 74.2} & {\large 74.5} \\ \midrule
\multirow{3}{*}{\textbf{\makecell{USM\\{\small (P@5 on ALL)}}}} & Sum-LST4 & \large{74.6} & \large{65.9} & \large{74.6} & \large{74.3} \\
 & SP-WSD & \large{73.6} & \large{81.6} & \large{83.1} & \large{85.7} \\
 & SP-USM $\star$ & \underline{\large{86.7}} & \underline{\large{87.3}} & \underline{\large{86.9}} & \textbf{\underline{\large{87.6}}} \\ \midrule
\multirow{3}{*}{\textbf{\makecell{WiC\\{\small (ACC on Val.)}}}} & Sum-LST4 & \underline{\large{71.8}} & \large{61.0} & \textbf{\underline{\large{72.1}}} & \large{66.8} \\
 & SP-WSD $\star$ & \underline{\large{71.8}} & \underline{\large{67.9}} & \large{68.8} & \underline{\large{68.7}} \\
 & SP-USM & \large{67.7} & \large{65.0} & \large{68.5} & \large{67.9} \\ \midrule
\multirow{3}{*}{\textbf{\makecell{GWCS\\{\small (COR on ST2)}}}} & Sum-LST4 & \large{74.4} & \large{54.9} & \large{73.3} & \large{71.5} \\
 & SP-WSD $\star$ & \underline{\large{76.3}} & \textbf{\underline{\large{78.7}}} & \large{75.7} & \large{75.2} \\
 & SP-USM & \large{73.4} & \large{75.4} & \underline{\large{77.2}} & \underline{\large{75.9}} \\ \midrule
\multirow{3}{*}{\textbf{\makecell{SID\\{\small (COR on ALL)}}}} & Sum-LST4 & \large{77.4} & \large{41.1} & \large{73.8} & \large{72.8} \\
 & SP-WSD & \large{76.3} & \large{77.7} & \large{73.5} & \large{75.3} \\
 & SP-USM $\star$ & \underline{\large{77.8}} & \textbf{\underline{\large{79.5}}} & \underline{\large{74.1}} & \underline{\large{77.4}} \\ \bottomrule
\end{tabular}%
}
\caption{Impact of pooling operation on task performance. Underline highlights pooling operation that performed best for each NLM and task. Bold highlights NLM and pooling operation that performed best for each task. $\star$ denotes default choice of LMMS-SP.}
\label{tab:profiles_tasks}
\end{table}

\subsection{Unambiguous Word Annotations}
\label{analysis:annotations}

In this work we learnt our initial set of sense embeddings (as described in Sections \ref{met:embed} and \ref{met:prop}) using SemCor, the only source of sense annotations used for LMMS \citeyearpar{loureiro-jorge-2019-language}, in combination with UWA \citep{loureiro-camacho-collados-2020-dont}, a set of sense annotations exclusively targeting unambiguous words.

On Table \ref{tab:prop_uwa} we present results showing the impact of UWA 
on task performance.
As noted in \citet{loureiro-camacho-collados-2020-dont}, the increase in WordNet coverage using UWA allows for disentangling dense clusters that coarsen the semantic space when relying on SemCor and network propagation alone.
Consequently, we expect UWA to benefit sense matching tasks, which is confirmed by our results showing substantial improvements in USM and SID (the two tasks using SP-USM).
We also find that UWA does not hinder performance on the remaining tasks for most model-task combinations (improves in most cases), with the exceptions of GWCS with {\small RoBERTa-L} and WiC with {\small ALBERT-XXL} (the former is also an exception observed in the sense profile ablation on $\S$\ref{analysis:profiles}).

As future work, we will also explore WordNet-independent procedures to discover monosemous words, such as the method introduced by \citet{soler2021let}, which may lead to further improvements.

\begin{table}[ht]
\centering
\resizebox{1.0\textwidth}{!}{%
\begin{tabular}{cC{7.6em}C{7.6em}C{7.6em}C{7.6em}C{7.6em}} \toprule
\textbf{Task (Metric)} & \textbf{Annotations} & \textbf{BERT-L} & \textbf{XLNet-L} & \textbf{RoBERTa-L} & \textbf{ALBERT-XXL} \\ \midrule \midrule
\multirow{2}{*}{\textbf{\makecell{WSD\\{\small (F1 on ALL)}}}} & SemCor & {\large 75.0} & \underline{{\large 74.1}} & \underline{{\large 75.2}} & {\large 75.4} \\
 & SemCor+UWA $\star$ & \underline{{\large 75.2}} & \underline{{\large 74.1}} & \underline{{\large 75.2}} & \textbf{\underline{{\large 75.5}}} \\ \midrule
\multirow{2}{*}{\textbf{\makecell{USM\\{\small (P@5 on ALL)}}}} & SemCor & {\large 76.3} & {\large 76.5} & {\large 76.1} & {\large 77.4} \\
 & SemCor+UWA $\star$ & \underline{{\large 86.7}} & \underline{{\large 87.3}} & \underline{{\large 86.9}} & \textbf{\underline{{\large 87.6}}} \\ \midrule
\multirow{2}{*}{\textbf{\makecell{WiC\\{\small (ACC on Val.)}}}} & SemCor & {\large 71.2} & {\large 67.1} & {\large 68.5} & \underline{{\large 69.1}} \\
 & SemCor+UWA $\star$ & \textbf{\underline{{\large 71.8}}} & \underline{{\large 67.9}} & \underline{{\large 68.8}} & {\large 68.7} \\ \midrule
\multirow{2}{*}{\textbf{\makecell{GWCS\\{\small (COR on ST2)}}}} & SemCor & {\large 76.1} & \underline{\textbf{{\large 78.7}}} & \underline{{\large 76.3}} & {\large 72.7} \\
 & SemCor+UWA $\star$ & \underline{{\large 76.3}} & \textbf{\underline{{\large 78.7}}} & {\large 75.7} & \underline{{\large 75.2}} \\ \midrule
\multirow{2}{*}{\textbf{\makecell{SID\\{\small (COR on ALL)}}}} & SemCor & {\large 72.1} & {\large 75.2} & {\large 65.3} & {\large 73.5} \\
 & SemCor+UWA $\star$ & \underline{{\large 77.8}} & \textbf{\underline{{\large 79.5}}} & \underline{{\large 74.1}} & \underline{{\large 77.4}} \\ \bottomrule
\end{tabular}%
}
\caption{Impact of sense annotations on task performance. Underline highlights pooling operation that performed best for each NLM and task. Bold highlights NLM and pooling operation that performed best for each task. $\star$ denotes default choice of LMMS-SP.}
\label{tab:prop_uwa}
\end{table}

\subsection{Merging Gloss Representations}
\label{analysis:glosses}

Another aspect of LMMS-SP that differs from LMMS \citeyearpar{loureiro-jorge-2019-language} is merging gloss embeddings by averaging with sense embeddings, instead of through concatenation (described in Section \ref{met:gls}).

Results on Table \ref{tab:gloss_analysis} reveal that concatenation only benefits WSD, with minor improvements over averaging.
Alternatively, averaging shows clear improvements for all other tasks (again, the exception is GWCS w/RoBERTa-L).

We also report performance using exclusively gloss representations, and sense embeddings without gloss information.
Surprisingly, these two sets of results are very close on WiC, GWCS and SID, showing that unsupervised representations learned from glosses can be competitive on particular tasks.

\begin{table}[ht]
\centering
\resizebox{1.0\textwidth}{!}{%
\begin{tabular}{cC{7em}C{7.6em}C{7.6em}C{7.6em}C{7.6em}} \toprule
\textbf{Task (Metric)} & \textbf{Glosses} & \textbf{BERT-L} & \textbf{XLNet-L} & \textbf{RoBERTa-L} & \textbf{ALBERT-XXL} \\ \midrule \midrule
\multirow{4}{*}{\textbf{\makecell{WSD\\{\small (F1 on ALL)}}}} & Without & {\large 74.6} & \underline{{\large 74.3}} & \underline{{\large 75.3}} & \textbf{\underline{{\large 75.5}}} \\
 & Exclusively & {\large 57.1} & {\large 55.2} & {\large 55.1} & {\large 54.3} \\
 & Averaged $\star$ & {\large 75.2} & {\large 74.1} & {\large 75.2} & \textbf{\underline{{\large 75.5}}} \\
 & Concatenated & \textbf{\underline{{\large 75.5}}} & \underline{{\large 74.3}} & \underline{{\large 75.3}} & {\large 74.8} \\ \midrule
\multirow{4}{*}{\textbf{\makecell{USM\\{\small (P@5 on ALL)}}}} & Without & {\large 83.5} & {\large 83.9} & {\large 83.7} & {\large 84.2} \\
 & Exclusively & {\large 46.4} & {\large 44.6} & {\large 40.5} & {\large 43.4} \\
 & Averaged $\star$ & \underline{{\large 86.7}} & \underline{{\large 87.3}} & \underline{{\large 86.9}} & \textbf{\underline{{\large 87.6}}} \\
 & Concatenated & {\large 85.0} & {\large 85.7} & {\large 85.8} & {\large 86.1} \\ \midrule
\multirow{4}{*}{\textbf{\makecell{WiC\\{\small (ACC on Val.)}}}} & Without & {\large 66.8} & {\large 64.6} & {\large 68.7} & {\large 67.1} \\
 & Exclusively & {\large 66.3} & {\large 62.2} & {\large 66.8} & {\large 64.1} \\
 & Averaged $\star$ & \textbf{\underline{{\large 71.8}}} & \underline{{\large 67.9}} & \underline{{\large 68.8}} & \underline{{\large 68.7}} \\
 & Concatenated & {\large 69.3} & {\large 66.5} & {\large 68.5} & {\large 67.7} \\ \midrule
\multirow{4}{*}{\textbf{\makecell{GWCS\\{\small (COR on ST2)}}}} & Without & {\large 75.6} & {\large 77.3} & {\large 75.0} & {\large 74.4} \\
 & Exclusively & {\large 75.1} & {\large 72.9} & {\large 75.0} & {\large 68.6} \\
 & Averaged $\star$ & \underline{{\large 76.3}} & \textbf{\underline{{\large 78.7}}} & {\large 75.7} & \underline{{\large 75.2}} \\
 & Concatenated & {\large 75.7} & {\large 77.2} & \underline{{\large 75.8}} & {\large 74.8} \\ \midrule
\multirow{4}{*}{\textbf{\makecell{SID\\{\small (COR on ALL)}}}} & Without & {\large 69.5} & {\large 72.5} & {\large 62.0} & {\large 68.3} \\
 & Exclusively & {\large 68.3} & {\large 70.0} & {\large 65.1} & {\large 65.9} \\
 & Averaged $\star$ & \underline{{\large 77.8}} & \textbf{\underline{{\large 79.5}}} & \underline{{\large 74.1}} & \underline{{\large 77.4}} \\
 & Concatenated & {\large 76.7} & {\large 77.9} & {\large 69.9} & {\large 73.8} \\ \bottomrule
\end{tabular}%
}
\caption{Impact of merging gloss representations on task performance. Underline highlights pooling operation that performed best for each NLM and task. Bold highlights NLM and pooling operation that performed best for each task. $\star$ denotes default choice of LMMS-SP.}
\label{tab:gloss_analysis}
\end{table}

\subsection{Learning Synsets Directly}
\label{analysis:indirect}

The SID task, as well as the synset version of USM, require synset-level embeddings.
In Section \ref{met:embed}, we explain that LMMS-SP synset embeddings are learned directly from sensekey annotations that are converted to synsets.
However, in our evaluation we compare LMMS-SP with other works that are only available as sensekey embeddings, so we converted these representations into synset embeddings learned as the average of corresponding sensekey embeddings (i.e., learned indirectly).

On Table \ref{tab:synset_repr} we compare LMMS-SP embeddings learned directly and indirectly, showing that learning these representations directly leads to an average improvement across models of 7.3\% on USM, and 1.5\% on SID.
The fact that indirect representation of synsets has a reduced impact on SID performance, in comparison to USM, suggests that indirect representation leads to more intermingled synset embeddings (i.e., harder to rank), but nearly as globally coherent as those learned from direct representation.

\begin{table}[ht]
\centering
\resizebox{1.0\textwidth}{!}{%
\begin{tabular}{cC{7em}C{7.6em}C{7.6em}C{7.6em}C{7.6em}} \toprule
\textbf{Task (Metric)} & \textbf{Synset Repr.} & \textbf{BERT-L} & \textbf{XLNet-L} & \textbf{RoBERTa-L} & \textbf{ALBERT-XXL} \\ \midrule \midrule
\multirow{2}{*}{\textbf{\makecell{USM\\{\small (P@5 on ALL)}}}} & Indirect & \large{74.5} & \large{76.1} & \large{77.3} & \large{76.3} \\
 & Direct $\star$ & \underline{\large{81.7}} & \textbf{\underline{\large{82.7}}} & \underline{\large{80.1}} & \underline{\large{81.9}} \\ \midrule
\multirow{2}{*}{\textbf{\makecell{SID\\{\small (COR on ALL)}}}} & Indirect & \large{77.3} & \large{78.6} & \large{73.0} & \large{75.5} \\
 & Direct $\star$ & \underline{\large{77.8}} & \textbf{\underline{\large{79.5}}} & \underline{\large{74.1}} & \underline{\large{77.4}} \\ \bottomrule
\end{tabular}%
}
\caption{Impact of learning synset representations directly from annotations, or indirectly as the average of corresponding sensekey embeddings. Underline highlights pooling operation that performed best for each NLM and task. Bold highlights NLM and pooling operation that performed best for each task. $\star$ denotes default choice of LMMS-SP.}
\label{tab:synset_repr}
\end{table}

\subsection{Part-of-Speech Performance}
\label{analysis:pos}

In \citet{loureiro-jorge-2019-language} we presented an error analysis targeting part-of-speech mismatch between predicted and ground-truth senses, which showed that verbs were particularly challenging.
In this work, we complement those results by reporting performance by part-of-speech, while comparing LMMS \citeyearpar{loureiro-jorge-2019-language} with LMMS-SP.

Our results on Table \ref{tab:pos_comparison} confirm that verbs remain the most challenging part-of-speech to disambiguate correctly, although {\small ALBERT-XXL} shows appreciably better verb results than the other NLMs used in this work.
Considering ranked USM matches, however, we find a much narrower gap between verbs and other parts-of-speech using LMMS-SP, with verbs performing comparably with adjectives and nouns, and only {\small BERT-L} showing differences larger than 1\%.

It is also interesting to note that {\small XLNet-L} outperforms or equals {\small ALBERT-XXL} on USM for all parts-of-speech with the exception of adverbs, providing better insight into the overall performance differences reported in USM evaluation ($\S$\ref{eval:usm}), where {\small ALBERT-XXL} outperforms {\small XLNet-L}.

\begin{table}[htb]
\centering
\resizebox{\textwidth}{!}{%
\begin{tabular}{lcclcclcclcc} \toprule
\multirow{2}{*}{\textbf{Model}} & \multicolumn{2}{c}{\textbf{Nouns}} &  & \multicolumn{2}{c}{\textbf{Verbs}} &  & \multicolumn{2}{c}{\textbf{Adjectives}} &  & \multicolumn{2}{c}{\textbf{Adverbs}} \\ \cline{2-3} \cline{5-6} \cline{8-9} \cline{11-12} 
 & \textbf{WSD} & \textbf{USM} &  & \textbf{WSD} & \textbf{USM} &  & \textbf{WSD} & \textbf{USM} &  & \textbf{WSD} & \textbf{USM} \\ \midrule \midrule
{MFS} & 67.6 & N/A &  & 49.6 & N/A &  & 78.3 & N/A &  & 80.5 & N/A \\ \midrule
{LMMS\textsubscript{1024} (2019)} & 75.6 & 48.2 &  & 63.6 & 65.3 &  & 79.8 & 75.6 &  & \textbf{85.0} & 78.6 \\
{LMMS\textsubscript{2048} (2019)} & 78.0 & 54.3 &  & 64.0 & 64.6 &  & \textbf{80.7} & 74.0 &  & 83.5 & 77.2 \\ \midrule
LMMS-SP\textsubscript{BERT-L}        & 78.0 & 87.2 &  & 63.0 & 84.2 &  & 80.3 & 85.3 &  & 83.8 & 96.5 \\
LMMS-SP\textsubscript{XLNet-L}       & 76.8 & \textbf{87.5} &  & 63.3 & 86.6 &  & 76.9 & \textbf{86.8} &  & \textbf{85.0} & 96.5 \\
LMMS-SP\textsubscript{RoBERTa-L}     & \textbf{78.2} & 86.9 &  & 63.1 & \textbf{86.8} &  & 79.2 & 86.1 &  & 84.7 & 96.2 \\
LMMS-SP\textsubscript{ALBERT-XXL}    & 77.8 & 87.3 &  & \textbf{65.6} & 86.6 &  & 79.1 & 86.7 &  & 84.1 & \textbf{97.1} \\ \bottomrule
\end{tabular}%
}
\caption{Performance on the combined set of \citet{raganato-etal-2017-word}, grouped by part-of-speech. Reporting F1 for WSD and P@5 for USM. MFS not applicable for USM.}
\label{tab:pos_comparison}
\end{table}

\section{Discussion}
\label{disc:intro}

In this section we discuss the main findings of our work.
More specifically, we discuss sense representation at specific layers of NLMs ($\S$\ref{disc:layers}), differences observed across models and variants ($\S$\ref{disc:meaning}), and finally, how our sense embeddings may benefit downstream tasks ($\S$\ref{disc:apps}).

\subsection{Layer Distribution}
\label{disc:layers}

Throughout this article we have provided empirical evidence supporting that there is substantial non-monotonic variation in the adeptness of specific layers of Transformer-based NLMs for sense representation.
This evidence is available from both our probing analysis and the improvements in several sense-related tasks obtained from using our proposed sense profiles, which are based on non-monotonic pooling from all layers (most clearly shown in Table \ref{tab:heatmap}).

The cause for this variation remains elusive, calling for controlled experiments where different NLMs are tested under comparable circumstances, particularly with regards to training data and modelling objectives, although such an experimental setup may be cost-prohibitive for models of this scale.
Nevertheless, seeking to better understand this variation, we conducted two qualitative experiments targeting representations of the same sentences at different layers.

\begin{figure}[htb]
  \centering
  \includegraphics[width=1.0\textwidth]{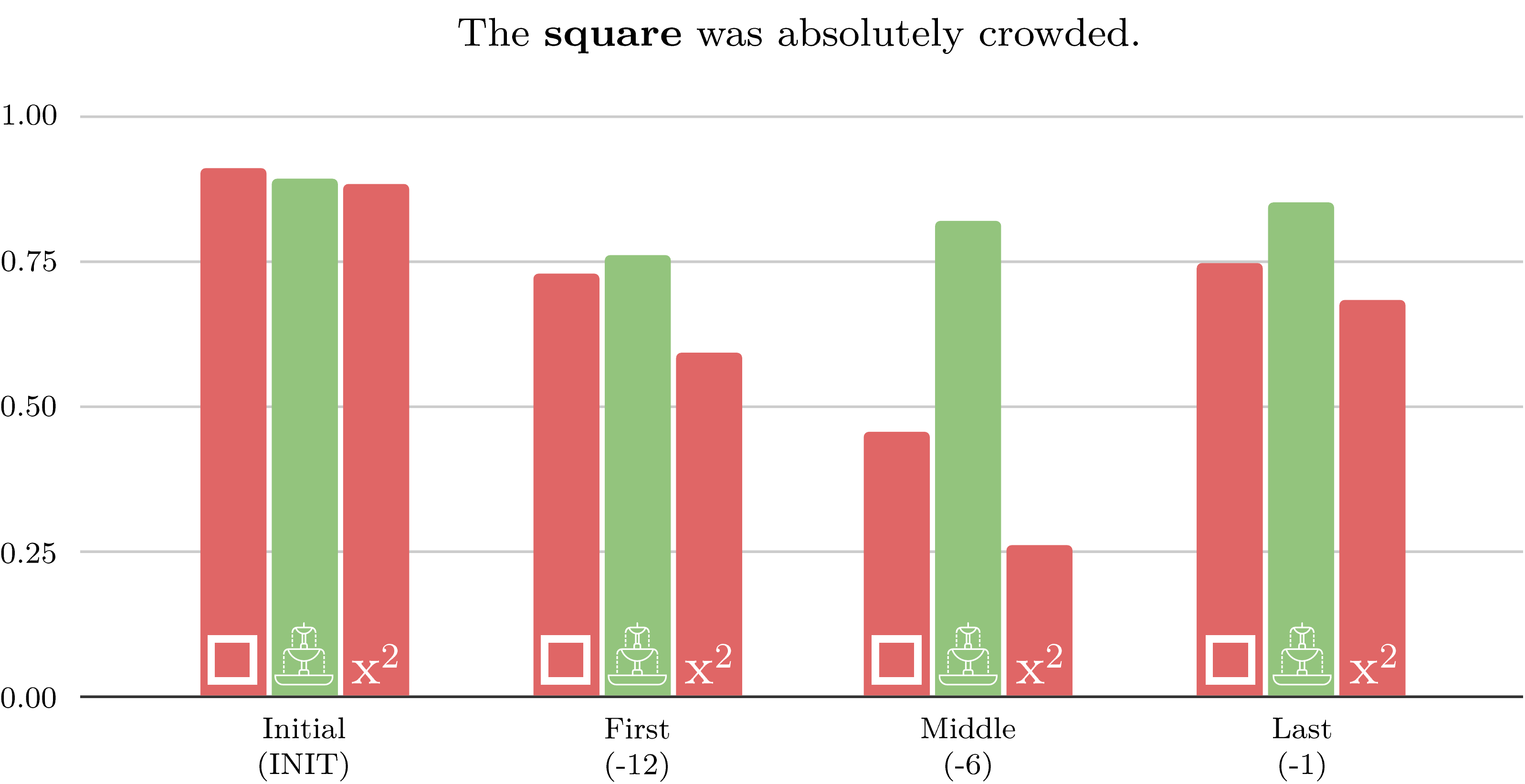}
  \caption{Cosine similarity at specific layers for the word `square' in context and the 3 senses annotated in SemCor, represented with {\small ALBERT-XXL}. The 3 senses of square correspond to the shape, public square (correct, green bar chart) and mathematical operation, in order. Initial layer similarities are influenced by the word forms used with each sense in SemCor.}
  \label{fig:square}
\end{figure}

In our first qualitative experiment, we compared sense similarity at different layers for the same word in context. We found some evidence potentially in support of the hypothesis advanced by \citet{voita-etal-2019-analyzing}, with the distribution of final layers resembling the distribution of the first layers moreso than the distribution of middlemost layers, where the difference between correct and incorrect senses is more marked (see example in Figure \ref{fig:square}).

\begin{figure}[htb]
  \centering
  \includegraphics[width=1.0\textwidth]{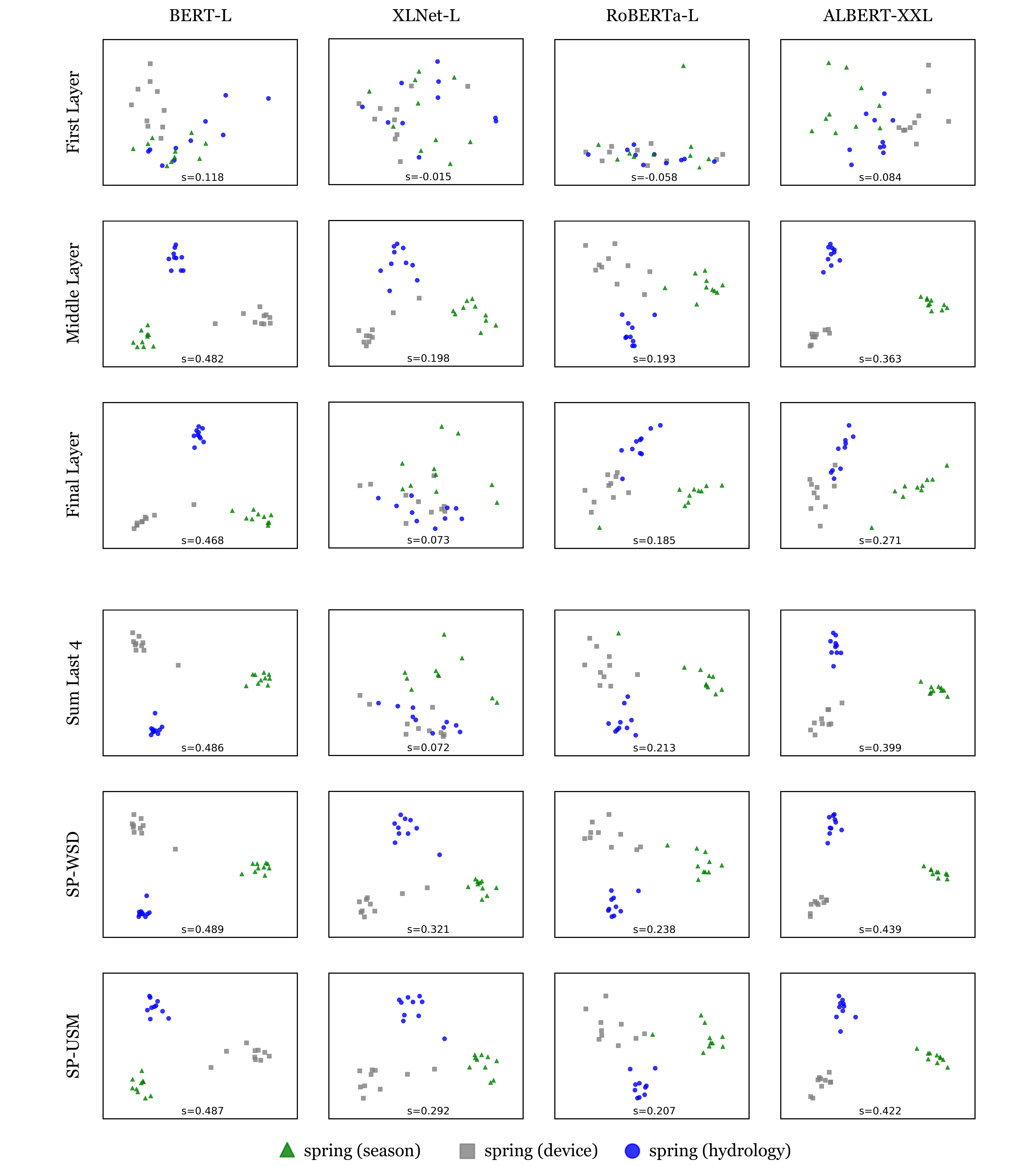}
  \caption{Visualization of embedding spaces using different pooling strategies. The last two rows correspond to our proposed pooling strategies (see $\S$\ref{met:lay} and $\S$\ref{probing:results}). Each point corresponds to an embedding for the word ``spring" in context, as provided in the 10-shot set of CoarseWSD-20 \citep{loureiro-etal-analysis-2021}. Using PCA for dimensionality reduction. Silhouette scores \textit{s} are computed before reduction.}
  \label{fig:pca}
\end{figure}

We further extended the previous experiment to a cluster-level comparison of the embedding space. For this experiment, we focus on the words present in the CoarseWSD-20 dataset \citep{loureiro-etal-analysis-2021}, both in aggregate for measuring correct sense clustering, as well as targeting ``spring" and its three distinct senses for visualization.
Considering silhouette scores\footnote{We use the mean silhouette coefficients of all embeddings for a particular word to measure how well each model and pooling strategy can assign embeddings to the correct sense cluster. Silhouette coefficients are based on intra- and nearest-cluster cosine similarities. Low values represent overlapping clusters.} \citep{rousseeuw} and PCA visualizations of the embedding space (Table \ref{tab:silhouette_scores} and Figure \ref{fig:pca}), we arrived at similar conclusions, namely that final layers tend to produce less accurate representations than layers closer to the middle, while the first layer show lowest scores.
Our proposed layer pooling methods also show generally improved clustering in comparison to the sum of the last four layers.
In addition, this experiment further confirms the unexpected finding regarding a different pattern of semantic representation across layers for XLNet-L, with representations from its final layer showing atypical dispersion.

We leave a more thorough large-scale analysis of this phenomenon for future work, alongside how to appropriately account for measuring the granularity of the different senses of a word, among other confounding factors.

\begin{table}[htb]
\centering
\resizebox{1.0\textwidth}{!}{%
\begin{tabular}{lC{7.6em}C{7.6em}C{7.6em}C{7.6em}} \toprule
\textbf{Pooling} & \textbf{BERT-L} & \textbf{XLNet-L} & \textbf{RoBERTa-L} & \textbf{ALBERT-XXL} \\ \midrule \midrule
First Layer  & 0.156 & 0.064 & 0.010 & 0.137 \\
Middle Layer & 0.384 & 0.167 & 0.180 & 0.328 \\
Final Layer  & 0.369 & 0.049 & 0.210 & 0.273 \\ \midrule
Sum Last 4   & 0.376 & 0.088 & \textbf{0.218} & 0.347 \\
SP-WSD       & 0.377 & 0.250 & 0.203 & 0.387 \\
SP-USM       & \textbf{0.388} & \textbf{0.255} & 0.196 & \textbf{0.390} \\ \bottomrule
\end{tabular}
}
\caption{Mean silhouette scores for all 20 words of the 10-shot training instances (balanced) of CoarseWSD-20 \citep{loureiro-etal-analysis-2021}. Top rows report scores for specific layers, bottom rows report scores when pooling the sum of last 4 layers and our proposed pooling strategies.}
\label{tab:silhouette_scores}
\end{table}

\clearpage

\subsection{NLM Idiosyncrasies}
\label{disc:meaning}

Besides unexpected results regarding the performance of particular layers of NLMs, we also find intriguing differences in the patterns of layer performance observed across models, and even variants of the same model.
Looking at our results on Table \ref{tab:heatmap}, we find many intriguing examples of this variation.

For the WSD task, the most striking examples are the differences between {\small BERT-L-UNC-WHL} and any other BERT model, and the bi-modal distribution for XLNet. For example, {\small XLNet-B} exhibited its best-performing layer near the top of the model, while the best-performing layer for {\small XLNet-L} is in the bottom-half of the model.
While results for USM are more consistent, we also find some peculiarities there, such as XLNet models showing worst-performing layers at the top, and {\small ALBERT-XL} showing a more biased distribution than other ALBERT variants.

The reasons for these differences in patterns across models and variants are not straightforward, specially considering many of these models are trained on very similar data and architectures.
Still, among several technical differences, we highlight the differences in modelling objectives covered in Section \ref{exp:models}.
Out of the 4 the models we considered in this work (see performance summary on Table \ref{tab:perf_overview}), XLNet is in fact simultaneously the model that appears most distinctive, with particularly strong performance on graded similarity tasks, and whose objectives are most different (being the only model not using MLM).
Another interesting finding is that we obtain best results on WSD, USM and WiC using {\small ALBERT-XXL}, which has half the layers of the other models, but much larger embedding dimensionality (model details are available in Table \ref{tab:nlm_details}).
As for differences in variants of the same model (same objectives) we consider the possibility that trivial run-time parameters may have an impact on this variation, akin to the unexpected influence of random seeds on fine-tuning BERT models \citep{dodge2020finetuning}.

\begin{table}[htb]
\centering
\resizebox{0.8\textwidth}{!}{%
\begin{tabular}{lccccc} \toprule
\multirow{2}{*}{\textbf{Model}} & \textbf{WSD} & \textbf{USM} & \textbf{WiC} & \textbf{GWCS} & \textbf{SID} \\
 & \textbf{{\small (F1)}} & \textbf{{\small (P@5)}} & \textbf{{\small (ACC)}} & \textbf{{\small (COR)}} & \textbf{{\small (COR)}} \\ \midrule \midrule
LMMS-SP\textsubscript{BERT-L}     & 75.2 & 86.7 & 67.4 & 76.3 & 77.8 \\
LMMS-SP\textsubscript{XLNet-L}    & 74.1 & 87.3 & 66.1 & \textbf{78.7} & \textbf{79.5} \\
LMMS-SP\textsubscript{RoBERTa-L}  & 75.2 & 86.9 & 67.8 & 75.7 & 74.1 \\
LMMS-SP\textsubscript{ALBERT-XXL} & \textbf{75.5} & \textbf{87.6} & \textbf{67.9} & 75.2 & 77.4 \\ \bottomrule
\end{tabular}%
}
\caption{Summary comparison between different NLMs using our LMMS-SP approach.}
\label{tab:perf_overview}
\end{table}

\subsection{Knowledge Integration}
\label{disc:apps}

The ability of matching WordNet synsets to any fragment of text allows downstream applications to easily leverage the manually curated relations available on WordNet.
At the same time, these sense embeddings can also serve as an entry point to many other knowledge bases linked to WordNet, such as the multilingual knowledge graph of BabelNet \citep{NavigliPonzetto:10}, the common-sense triples of ConceptNet \citep{conceptnet55} or WebChild \citep{tandon-etal-2017-webchild}, the semantic frames of VerbNet \citep{KipperSchuler2006}, and even the images of ImageNet \citep{ILSVRC15} or Visual Genome \citep{krishnavisualgenome}.
Several recent works have used the symbolic relations expressed in these knowledge bases to improve neural solutions to Natural Language Inference \citep{Kapanipathi_2020}, Commonsense Reasoning \citep{lin-etal-2019-kagnet}, Story Generation \citep{Ammanabrolu_Tien_Cheung_Luo_Ma_Martin_Riedl_2020}, among others.

\begin{figure}[htb]
  \centering
  \includegraphics[width=0.9\textwidth]{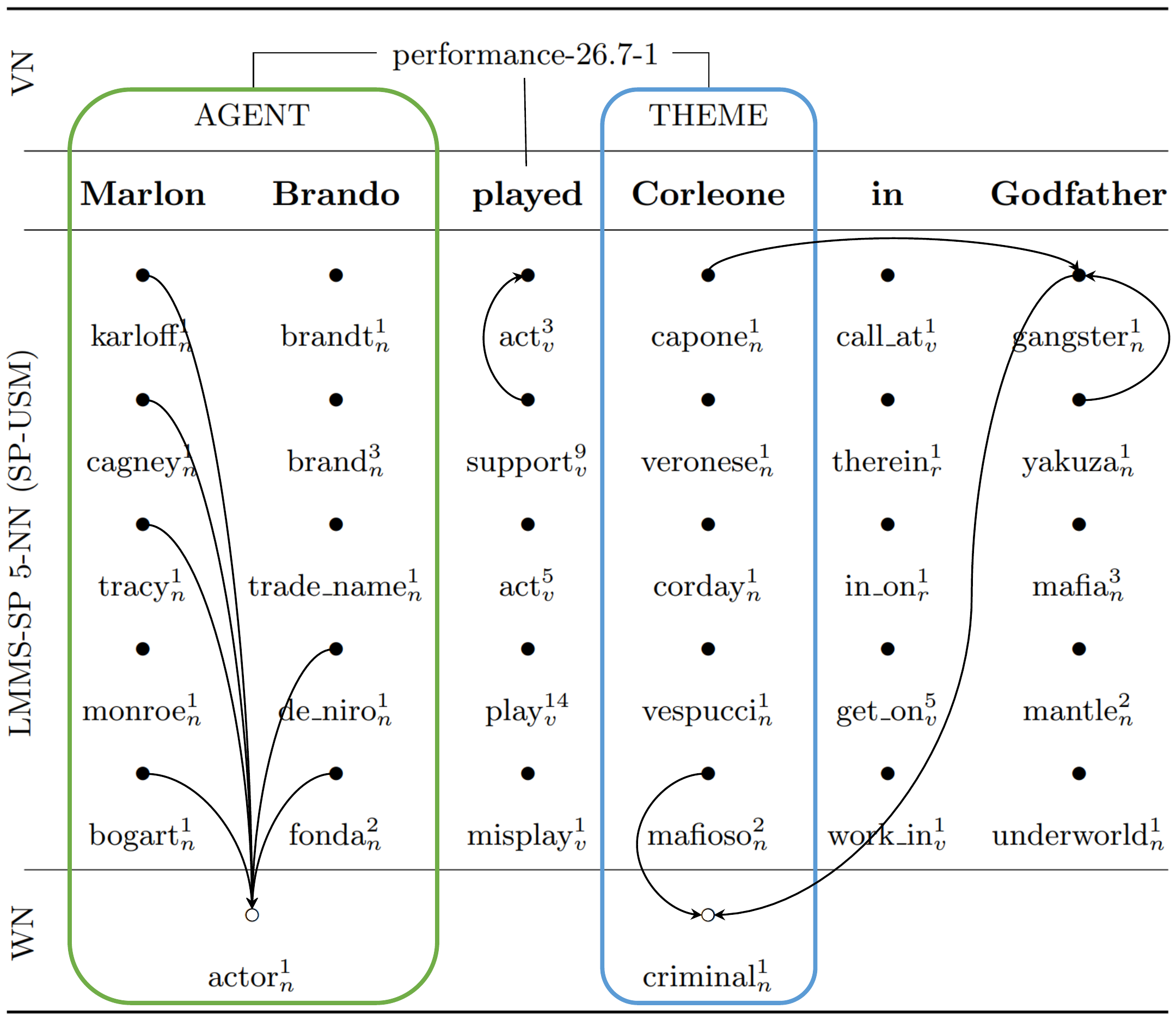}
  \caption{Example sentence with each token matched to LMMS-SP\textsubscript{ALBERT-XXL} sense embeddings, presenting synsets for the 5 nearest neighbors, using the SP-USM sense profile. Shows direct hypernymy relations (i.e., Is-A), included in WordNet (WN), between matched synsets, as well as hypernymy relations shared between more than one matched and unmatched synset (i.e., deducible generalizations, not in top 5 matches). Finally, at the top, we show a VerbNet (VN) semantic frame matched to this sentence, highlighting how LMMS-SP enables generalization of argument spans.}
  \label{fig:apps}
\end{figure}

As an example of how using LMMS-SP to bridge natural language and symbolic knowledge can be beneficial, in Figure \ref{fig:apps} we demonstrate how these sense embeddings allow for generalization of argument spans, predicted by a semantic parser, exploiting WordNet relations between matched synsets.
The matches shown in Figure \ref{fig:apps} also illustrate how sense embeddings may be used for probing world knowledge encoded in pre-trained NLMs, as already suggested in \citet{loureiro-jorge-2019-language}.

\clearpage

\section{Conclusion}
\label{conclusion}

Leveraging neural language models in combination with sense-annotated corpora (and complementary resources such as glosses or relations), this work has shown that it is possible to produce sense embeddings applicable beyond mere disambiguation, with relevant implications for long-standing challenges in Artificial Intelligence such as symbol grounding.

This extension of \citet{loureiro-jorge-2019-language} proposes a more principled approach for learning distributional representations of word senses using pre-trained NLMs, focusing on state-of-the-art Transformer-based models. From extensive evaluation on several sense-related tasks, we demonstrated that the LMMS-SP approach is more effective than prior work at approximating precise word sense representations in the same vector space of NLMs.

The broad probing analysis of the many variants of popular NLMs endeavored in this work provides new evidence supporting further research on the interplay between pre-training objectives, layer specialization, and model size.
The conclusions of this probing analysis are indeed expected to be applicable in tasks outside WSD, and for learning representations other than sense embeddings, which we leave for future work.

Effectively, there are known limitations to meaning representation based on language modelling objectives alone \citep{bender-koller-2020-climbing,merrill2021provable}. Nonetheless, we believe our work shows there is still much to understand about how to best leverage NLMs for meaning representation, in addition to more thoroughly testing the effectiveness of current approaches centered on self-supervision.

\paragraph{Release}
This work is accompanied by the release of the following resources: sensekey and synset embeddings with full-coverage of WordNet based on {\small BERT-L}, {\small XLNet-L}, {\small RoBERTa-L} and {\small ALBERT-XXL}; scripts to generate embeddings following our method, using the same NLMs or others supported by the Transformers package; and scripts to run task evaluations.
These resources are released under a GNU General Public License (v3) and available from this public repository: 
\url{https://github.com/danlou/lmms}

\clearpage

\section{Acknowledgements}
Daniel Loureiro is supported by the EU and Fundação para a Ciência e Tecnologia through contract DFA/BD/9028/2020 (Programa Operacional Regional Norte).
Jose Camacho-Collados is supported by a UKRI Future Leaders Fellowship.

\begin{appendices}

\section{Full results for SemEval 2020 - Task 3}
\label{appendix:semeval20}

On Table \ref{tab:se20_leaderboard} we report complete leaderboard results for subtasks 1 and 2 of SemEval 2020 Task 3 (including other languages besides English), during the evaluation period.

\begin{table}[ht]
\centering
\resizebox{\textwidth}{!}{%
\begin{tabular}{lcllcllcllc} \toprule
\multicolumn{5}{c}{\textbf{English}} &  & \multicolumn{5}{c}{\textbf{Hungarian}} \\ \cline{1-5} \cline{7-11} 
\textbf{Team} & \textbf{Sub1} &  & \textbf{Team} & \textbf{Sub2} &  & \textbf{Team} & \textbf{Sub1} &  & \textbf{Team} & \textbf{Sub2} \\
1. Ferryman & 0.774 &  & 1. MineriaUNAM & 0.723 &  & 1. N+S & 0.740 &  & 1. N+S & 0.658 \\
2. will\_go & 0.768 &  & \textbf{2.} LMMS\textsubscript{RoBERTa-L} & 0.720 &  & 2. Hitachi & 0.681 &  & 2. Hitachi & 0.616 \\
3. MULTISEM & 0.760 &  & 3. somaia & 0.719 &  & 3. InfoMiner & 0.754 &  & 3. MineriaUNAM & 0.613 \\
\textbf{4.} LMMS\textsubscript{RoBERTa-L} & 0.754 &  & 4. MULTISEM & 0.718 &  & 4. Ferryman & 0.774 &  & \textbf{4.} LMMS\textsubscript{XLMR-L} & 0.565 \\
5. InfoMiner & 0.754 &  & 5. InfoMiner & 0.715 &  & \textbf{5.} LMMS\textsubscript{XLMR-L} & 0.754 &  & 5. InfoMiner & 0.545 \\
 & \multicolumn{1}{l}{} &  &  & \multicolumn{1}{l}{} &  &  & \multicolumn{1}{l}{} &  &  & \multicolumn{1}{l}{} \\
\multicolumn{5}{c}{\textbf{Finnish}} &  & \multicolumn{5}{c}{\textbf{Slovenian}} \\ \cline{1-5} \cline{7-11} 
\textbf{Team} & \textbf{Sub1} &  & \textbf{Team} & \textbf{Sub2} &  & \textbf{Team} & \textbf{Sub1} &  & \textbf{Team} & \textbf{Sub2} \\
1. will\_go & 0.772 &  & 1. InfoMiner & 0.645 &  & 1. Hitachi & 0.654 &  & 1. N+S & 0.579 \\
2. Ferryman & 0.745 &  & 2. N+S & 0.611 &  & 2. InfoMiner & 0.648 &  & 2. InfoMiner & 0.573 \\
3. N+S & 0.726 &  & 3. MineriaUNAM & 0.597 &  & 3. N+S & 0.646 &  & 3. CitiusNLP & 0.538 \\
4. RTM & 0.671 &  & 4. MULTISEM & 0.492 &  & 4. CitiusNLP & 0.624 &  & 4. tthhanh & 0.516 \\
\textbf{11.} LMMS\textsubscript{XLMR-L} & 0.360 &  & \textbf{7.} LMMS\textsubscript{XLMR-L} & 0.354 &  & \textbf{8.} LMMS\textsubscript{XLMR-L} & 0.560 &  & \textbf{9.} LMMS\textsubscript{XLMR-L} & 0.483 \\ \bottomrule
\end{tabular}%
}
\caption{Results from the leaderboard of subtasks 1 and 2 of SemEval 2020 Task 3 - Predicting the (Graded) Effect of Context in Word Similarity. Rank reported in team names. At the time of this evaluation, we did not use the sense profiles proposed in this paper, so our reported results on this table are based on senses embeddings pooled from the last 4 layers of the specified models, following \citet{loureiro-jorge-2019-language}.}
\label{tab:se20_leaderboard}
\end{table}

\section{Stanford Contextual Word Similarities (SCWS)}
\label{appendix:scws}

On Table \ref{tab:scws_comparison} we report our results on the Stanford Contextual Word Similarities \citep[SCWS]{huang-etal-2012-improving} task.
We address this task similarly to GWCS (see Section \ref{eval:gcs}). Given two words in context, each within an independent sentence, we disambiguate both occurrences and score each pair as the average of similarities between corresponding sense and contextual embeddings.

Results on SCWS follow performance on GWCS, with XLNet-L outperforming other NLMs as well as results from related works.
Analysing performance by Part-of-Speech (POS), we find that nouns appear most challenging for this task, particularly when being compared against other nouns.

\begin{table}[ht]
\centering
\resizebox{\textwidth}{!}{%
\begin{tabular}{@{}lccccccc@{}}
\toprule
                & \textbf{ALL} & \textbf{N-N} & \textbf{N-V} & \textbf{N-A} & \textbf{V-V} & \textbf{V-A} & \textbf{A-A} \\
\textbf{System} & \small{($n$=2003)} & \small{($n$=1328)} & \small{($n$=140)} & \small{($n$=30)} & \small{($n$=399)} & \small{($n$=9)} & \small{($n$=97)} \\ \midrule
\cite{huang-etal-2012-improving}  & 65.7 & -- & -- & -- & -- & -- & -- \\
SensEmbed (\citeyear{iacobacci-etal-2015-sensembed})                   & 62.4 & -- & -- & -- & -- & -- & -- \\
NASARI (\citeyear{CamachoCollados2016NasariIE})                        & -- & 47.1 & -- & -- & -- & -- & -- \\
DeConf (\citeyear{pilehvar-collier-2016-de})                           & 71.5 & -- & -- & -- & -- & -- & -- \\
LessLex (\citeyear{colla-etal-2020-lesslex})                           & 69.5 & 69.2 & 69.6 & 82.0 & 64.1 & 73.6 & 63.8 \\
ARES (\citeyear{scarlini-etal-2020-contexts})                          & 67.9 & 66.6 & 68.6 & \textbf{87.9} & 67.2 & 66.7 & 69.4 \\ \midrule
BERT-L \small{(SP-WSD)}           & 59.3 & 56.8 & 67.4 & 78.4 & 59.4 & 60.0 & 61.1 \\
XLNet-L \small{(SP-WSD)}          & 73.9 & 71.6 & 75.6 & 81.3 & 75.8 & \textbf{78.3} & 76.0 \\
RoBERTa-L \small{(SP-WSD)}        & 63.8 & 59.1 & 71.3 & 66.6 & 68.7 & 73.3 & 66.7 \\
ALBERT-XXL \small{(SP-WSD)}       & 65.9 & 63.7 & 69.4 & 74.9 & 66.3 & 75.0 & 69.5 \\ \midrule
LMMS-SP\textsubscript{BERT-L}     & 64.1 & 62.3 & 67.1 & 82.6 & 63.5 & 51.7 & 68.3 \\
LMMS-SP\textsubscript{XLNet-L}    & \textbf{75.9} & \textbf{73.7} & \textbf{75.8} & 81.5 & \textbf{78.0} & 75.0 & \textbf{79.4} \\
LMMS-SP\textsubscript{RoBERTa-L}  & 67.4 & 63.4 & 73.9 & 70.8 & 71.1 & 75.0 & 68.1 \\
LMMS-SP\textsubscript{ALBERT-XXL} & 69.9 & 68.8 & 72.4 & 76.4 & 69.9 & 66.6 & 70.9 \\ \bottomrule
\end{tabular}
}
\caption{Results on SCWS (Spearman correlation scores, $\rho \times 100$), considering the entire set of pairs (ALL) as well as results for subsets pairing particular Parts-of-Speech (with $n$ denoting the number of instances for each subset), similarly to \cite{colla-etal-2020-lesslex}.}
\label{tab:scws_comparison}
\end{table}

\end{appendices}

\bibliography{mybibfile,anthology}

\end{document}